\documentclass{cshonours}
\usepackage{graphics}           
\usepackage{xcolor}             
\usepackage[draft]{graphicx}    
\usepackage{float}              
\usepackage{esvect}             
\usepackage{textcomp}           
\usepackage{amsmath}            
\usepackage{amssymb}            
\usepackage{tabularx}           
\usepackage{bm}                 
\usepackage{rotating}           
\usepackage{caption}            
\usepackage{array}
\usepackage{verbatim}           
\usepackage{changepage}         
\usepackage{bibentry}           
\usepackage{pdfpages}           
\makeatletter\let\saved@bibitem\@bibitem\makeatother
\usepackage{hyperref}
\makeatletter\let\@bibitem\saved@bibitem\makeatother

\newcommand{\todo}[1]{{\textcolor{red}{#1}}}
\newcommand{\etc}{etc. }
\newcommand{\ie}{i.e. }
\newcommand{\eg}{e.g. }
\newcommand{\etal}{et al. }
\newcommand{\etals}{et al.'s }
\newcommand{\degree}{$^{o}$ }
\newcommand{\vo}{\vec{o}\@ifnextchar{^}{\,}{}}             
\newenvironment{conditions}
  {\par\vspace{\abovedisplayskip}\noindent\begin{tabular}{>{$}l<{$} @{${}={}$} l}}
  {\end{tabular}\par\vspace{\belowdisplayskip}}
\newcommand{\loss}{\mathcal{L}}

\newcolumntype{P}[1]{>{\centering\arraybackslash}p{#1}}     

\title{Real World Robotic Exploration using Deep Neural Networks Trained in Photorealistic Reconstructed Environments}
\author{
    \begin{tabular}{ l l }
        Author     & Isaac Ronald Ward              \\
        Supervisor & Professor Mohammed Bennamoun   
    \end{tabular}                                   \\
    \vspace{8cm}                                    \\
    \begin{tabular}{ l l }
        Word count & $13560$                        \\
        Page count & $54$
    \end{tabular} \\
     (excluding appendix) 
}
\date{2019}

\begin{document}
\nobibliography*    
\maketitle

\categories{A.2, I.7.2}

\begin{abstract}

In this work, an existing deep neural network approach for determining a robot's pose from visual information (RGB images) is modified, improving its localization performance without impacting its ease of training. Explicitly, the network's loss function is extended in a manner which intuitively combines the positional and rotational error in order to increase robustness to perceptual aliasing. An improvement in the localization accuracy for indoor scenes is observed: with decreases of up to $9.64$\% and $2.99$\% in the median positional and rotational error respectively, when compared to the unmodified network.

Additionally, photogrammetry data is used to produce a pose-labelled dataset which allows the above model to be trained on a local environment, resulting in localization accuracies of $0.11$m \& $0.89$\degree. This trained model forms the basis of a navigation algorithm, which is tested in real-time on a TurtleBot (a wheeled robotic device). As such, this work introduces a full pipeline for creating a robust navigational algorithm for any given real world indoor scene; the only requirement being a collection of images from the scene, which can be captured in as little as $330$ seconds of video. 

\end{abstract}

\begin{acknowledgements}
This work acknowledges the machine learning and robotics research team at the University of Western Australia. Namely, Professor Mohammed Bennamoun for his guidance and supervision, Mohammad Amir Asim Khan Jalwana for his technical advice and suggestions, and Uzair Nadeem for generously sharing his valuable 3D reconstruction data.

\noindent Special thanks as well to Madeleine Louise McKenzie, for allowing her FujiFilm X-T20 and 23mm prime autofocus lens to be used for the majority of the photogrammetry experiments completed in this work.

\end{acknowledgements}

\begin{externalcontributions}
The research into deep neural networks, computer vision, and imaging completed during this research project has contributed to external works which have been published or submitted for consideration. Naturally the body of this dissertation thus overlaps with some of the material in these publications/works. For completeness, said works which are relevant to this dissertation have been made available where possible, and they can be requested directly in the case where it was not possible to make them publicly accessible:

\noindent \cite{ward2019survey} \bibentry{ward2019survey}                   \\
\noindent \cite{ward2019augloss} \bibentry{ward2019augloss}               \\
\noindent \cite{allaert2019opticalflow} \bibentry{allaert2019opticalflow} \\
\end{externalcontributions}

\tableofcontents

\chapter{Introduction}
\label{ch:intro}

Machine learning has developed extraordinarily in the recent years, primarily due to breakthroughs in data collection and hardware acceleration, as well as the refinement of algorithms and architectures. As a result, machine learning techniques have facilitated the development of groundbreaking and destabilizing systems in a variety of fields, from Agriculture \cite{kamilaris2018agriculture} to Zoology \cite{guilford2009zoology}. A particular branch of machine learning is deep learning, and a particular deep learning architecture --- deep neural networks --- are of particular interest to this work. Deep neural networks have a multitude of attributes which encourage their application to problems in numerous domains; efficiency, generalization, ease-of-use, \etc But perhaps one of the most transformative is how the application of deep neural networks allows a \textit{data-driven} approach to be used when solving regression problems (instead of a knowledge driven approach).

The problem of concern in this work is that of \textbf{robotic navigation}, specifically within indoor contexts. Techniques which allow unmanned ground vehicles (UGVs), flying drones and other simple robotic devices to navigate semi-autonomously in their environments do currently exist, but improvements to such pipelines are required, especially with respect to robustness and generalization. Indeed, current techniques --- which include Simultaneous Localization And Mapping (SLAM), Monte Carlo based methods and so forth --- are \textit{not} robust to changes in illumination or clutter, they have high memory demands, require cumbersome-to-gather prior information regarding the environment or have other undesirable requirements/limitations. 

A key element in any robotic navigation pipeline, which deep learning techniques can improve, is \textbf{localization}. That is, figuring out \textit{where} one is relative to a given origin. A deep learning approach is a natural fit for such a problem; neural networks allow the mapping between an input set and an output to be learned, provided one has enough examples to learn from. In the case of robotic navigation, a device's on-board sensor array may include inertial measurement units (IMUs), odometers, cameras, rangefinders, \etc As such, this information can be used to \textit{regress} the device's position and rotation (herein referred to as its pose). Such techniques supersede the performance of Gobal Positioning System (GPS) devices, and provide a richer set of information with respect to a device's position and rotation. In this work, pose regression systems are a key point of study, and they are applied in the context of an indoor robotic navigation task. 

This idea is not a new one. It was proved as early as 2015 that such a pose regression system can in fact be developed \cite{kendall2015posenet}, but numerous avenues for improvement do exist (see Chapter~\ref{ch:litrev:s:imloc}). For example, accuracy and efficiency could always be improved, but a more interesting investigation is to emphasize \textit{indoor} performance and generalization, or other attributes that could contribute to more reliable robotics. Improvements such as these are motivated in Chapter~\ref{ch:litrev}, and are discussed in greater depth in Chapter~\ref{ch:obj}. 

In summary, the contributions provided by this work are threefold:
\begin{enumerate}
    \item The introduction of a pose-labelled dataset of an office scene inside the University of Western Australia's Computer Science \& Software Engineering building (see Section~\ref{ch:data:s:new}). 
    \item The proposal of a modification to pose regression networks, where the implemented proposal outperforms other systems in its class (especially in indoor environments). A quantitative comparison is provided.
    \item A demonstration of the proposed modified network being used with a robotic, TurtleBot platform. The modified pose regression network forms the basis of the robot's navigation functionality.
\end{enumerate}

\chapter{Literature review}
\label{ch:litrev}

\vspace{-2cm}

This chapter aims to provide a brief yet informative survey into the technologies and algorithms which are central to this work. Namely, the topics of interest relate to navigation, photogrammetry and pose determination. A simplified overview this work can be considered as the sum of these three technologies: \textbf{photogrammetry} is used to create the data necessary to train a deep learning based \textbf{pose determination} algorithm, which forms the basis of a robotic \textbf{navigation} algorithm. 

Since this review of the literature is necessarily brief, the reader is encouraged to read the cited publications should a more comprehensive understanding be required. 

\section{Robotic navigation pipelines}
\label{ch:litrev:s:nav}

The term `robotic navigation' is complex and typically refers to a \textit{set} of technologies which together allow for a device to navigate in some semi-autonomous manner. Again, though definitions vary wildly, here robotic navigation will be considered as a simple pipeline of \textbf{localization} and \textbf{pathfinding}. Image-based localization algorithms will be discussed in more depth in Section~\ref{ch:litrev:s:imloc}, as that particular branch of localization algorithms are of importance to this work. This section will focus primarily on navigation algorithms --- particular those of use in the context of indoor navigation for wheeled robots.

\subsection{Navigation algorithms}

As stated in the introductory chapter, a multitude of navigation algorithms do currently exist --- especially for simple robots. Some key algorithms (not necessarily the most performant) are discussed here, and their strengths and weaknesses are catalogued, forming a basis for the research outlined in this work.

\label{ch:litrev:s:nav:b:dead}
\textbf{Dead reckoning} is a simple map-based localization method where a device's kinematic information (odometry) is the only input used to estimate it's location. Accumulating error compounds in these estimates and the overall error quickly exceeds operational requirements. This naive method is not appropriate for indoor navigation applications, since the system cannot correct the errors which are introduced from the device's odometers. Clearly a more robust, adaptive method is required, especially if the \textit{kidnapped robot problem} is to be addressed:

\textit{The kidnapped robot problem.}
\newline
Consider a robot that has moved from its known location to an arbitrary location in its environment. If the device's sensors are disabled for the duration of the movement, it will have no data with which to approximate its new pose with respect to its original known location. The so called `kidnapped robot problem' asks how a device can \textit{re-localize} itself with any certainty from this new location, and simulates how a device might recover from a catastrophic failure. This problem is closely related to the `wake-up robot problem', whereby a device is tasked to localize itself when moved to an unknown location and then turned on.

\textbf{Monte-Carlo methods}. Monte Carlo Localization (MCL) and its variants are map-based methods, which use particle filters over a state space to take a probabilistic approach to dealing with compounding errors from sensory input\footnote{Please consult the work outlined in Fox \etals \cite{fox1999mcl} for \textbf{explicit} operational details of the MCL algorithm.}. These algorithms form a family of solutions to the \textit{kidnapped robot problem}. The operation of MCL is conceptually simple:
\begin{enumerate}
    \item A population of `particles', each which represents a pose, are randomly distributed over the legal locations on a map. See Fig.~\ref{fig:floormap} for an example of an occupancy map.
    
    \item The device's odometer registers its movement (with some error, as discussed earlier in Section~\ref{ch:litrev:s:nav:b:dead}). These translations are applied to each particle --- along with a random component which approximates the error in the device's odometry --- thus updating the pose represented by each particle. More explicitly, if a device is at an unknown position, then the algorithm's particle distribution over the map is completely uniform and is represented by a set of particles with positions $p_{i}$. If the device then moves such that the odometer returns an approximate change in its position of $\Delta x$, then each particle's position is updated such that $p_{i}=p_{i}+\Delta x+\epsilon$ where $\epsilon$ is a random value that approximates the error of the odometer.
    
    \item After each movement, the device's sensors are probed and a filter is applied to the set of particles $p_{i}$. If the sensor observations could not occur at $p_{i}$ then the $i^{th}$ particle is removed and re-initialized at a more suitable location in pose space (the modal pose, a random legal pose or the pose barycentre).
    
    \item After a number of sensor observations, the particles converge around a single pose, which must be the current pose; the device has been localized.
\end{enumerate}

MCL and its successor - Adaptive Monte Carlo Localization (AMCL) - are considered the `gold standard' for robotic localization as they provide a robust, context-independent method for robotic localization which greatly improves upon dead reckoning approaches. These algorithms do not require expensive sensors; the sensor observations used to inform the particle filter can be simple time-of-flight depth sensors and otherwise only an odometer is required.

Though otherwise robust and efficient, MCL algorithms typically fail to adapt to changing environments. Indoor domestic environments introduce moving obstacles, lighting changes and are not static, so neither of these solutions will operate robustly in indoor domestic contexts. 

\textbf{Deep network methods} use deep neural networks in order to generate paths, actions or other key elements in the navigation pipeline. One such example is outlined in \cite{zhu2016targetdrl}, where a neural network is trained on synthetic data to directly regress a robot's next movement, entirely from an image input. The robot renders a virtual image at its current location in a virtual 3D environemnt (or captures an image with a mounted camera when operating in the real world), and one of the four cardinal directions: forward, backward, left or right, is regressed from the network. This direction is the estimated direction that the robot needs to move in order to reach its target (specified only from an image). The device completes this movement and then repeats until it reaches its target. In this way, a deep neural network is the basis for the navigation pipeline. The pipeline produces significantly lower average trajectory lengths when compared to random walks and purpose-built reinforcement learning approaches: $210.7$ compared to $2744.3$ and $723.5$ respectively. Other similar methods exist \cite{levine2015visuomotor, mnih2015human}, demonstrating that there is potential for using deep neural networks in a robot's navigational pipeline. 

\textbf{Path planning methods}. In the case of this work, it is perhaps more prudent to consider path planning methods, rather than localization algorithms, as the intent here is to have the localization aspect provided by a pose regression network. Hence, in order to create the full navigation pipeline, this localization module needs only to be paired with a path planning algorithm. Path planning algorithms are simple for cases where a starting pose, a goal pose and an occupancy map are provided. Since 3D reconstructions and architectural maps can be converted to scaled grid-based occupancy maps quite easily, some appropriate algorithms would be Dijkstra's shortest path algorithm \cite{dijkstra1959path}, rapidly exploring random trees (RRTs) \cite{bry2011rrt, lavalle2000rrt} or some other heuristic-based path planning algorithms \cite{zhu2016targetdrl}.

\section{Photogrammetry}
\label{ch:litrev:s:photo}

Deep learning approaches, specifically neural networks, are data-hungry algorithms \cite{khan2018cnn}. In order to train such a model to regress a device's pose from what it sees, numerous labelled examples are required. More explicitly, images where the pose of the camera at the time the image was taken need to be collected, in order to allow the relationship between image and pose to be estimated. For a direct recount of how such data was collected for the experiments in this work, see Chapter~\ref{ch:data:s:new:ss:coll}. This section will instead focus on a more general discussion of the techniques which can be used, specifically focusing on one of the most prolific in the literature: \textbf{photogrammetry}.

Photogrammetry, or the use of photographs in recovering exact measurements, has been used increasingly for complex 2D to 3D reconstructions in the recent years (though it has been applied in topography, modelling, image analysis and even artistic endeavours. Though its application is broad, the focal point for this chapter is its use in 3D reconstruction from images --- a process which is accomplished using two distinct steps: Structure-from-Motion and Multi-View Stereo.

\subsection{Structure-from-Motion}

\begin{figure}[h]
    \centering
    \includegraphics[width=0.95\linewidth]{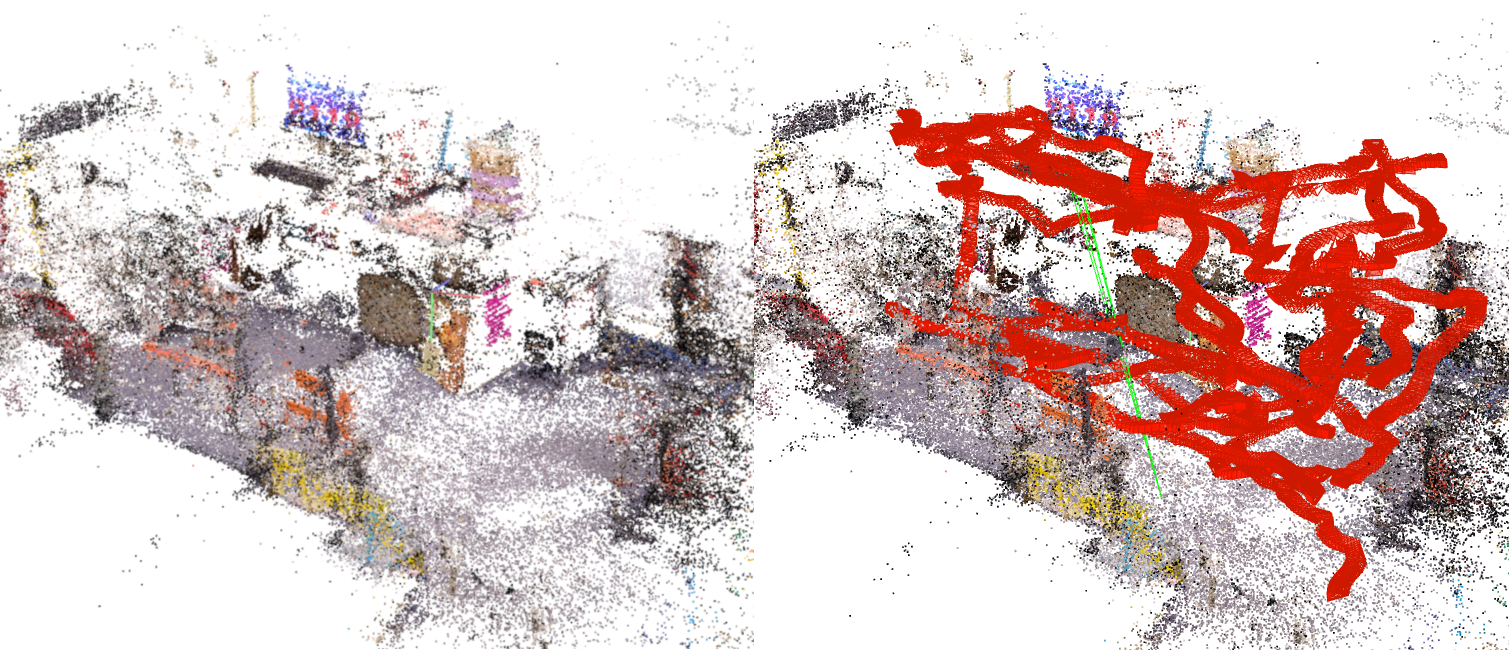}
    
    \caption{The sparse point cloud generated from a SfM reconstruction using 3228 images of an indoor office room (left), and the same point cloud with the camera poses throughout the image-capturing sequences, visualised using red frustrums (right). The green lines illustrate the relationship between a reconstructed point and the images used in its reconstruction.  }
    \label{fig:roomvidsparse}
\end{figure}

Structure-from-Motion (SfM) refers to the process by which a camera's pose and calibration parameters are estimated from a collection of images. These images can be taken from video sequences, can have different calibration parameters (\eg varying focal lengths) and can be ordered or un-ordered. This process creates \textit{structural} information from the image set (see Figure~\ref{fig:roomvidsparse}), and operates in the following steps:

\begin{enumerate}
    \item For a set of $N$ images $I_i$, a set of SIFT features $F_i$ are computed for each image \cite{lowe1999sift} (see Section~\ref{ch:litrev:s:imloc:ss:other:b:sift} for an introduction to SIFT features). Typically, successful reconstructions will have thousands of features computed per image, \ie  $\sum_{i=1}^{N} F_{i} \cdot \frac{1}{N} \approx 4000$. 
    \item Since it is expected that each image in $I_i$ will have significant visual overlap with at least one other image in $I_i$, the feature sets $F_i$ are exhaustively matched.
    \item As each image $I_i$ is geometrically verified, it can be incrementally registered to the scene if its features correspond to the set of already registered images. 
    \item As images are registered, reconstruction of the 3D scene via iterative triangulation; a 3D point can be estimated given at least two known views and two corresponding feature points. Bundle adjustment and outlier filtering is completed in each iteration after the triangulation process, in order to refine results \cite{triggs2000bundleadjustment}.
    \item Once all images that will be registered, are registered, the scene is considered sparsely reconstructed.
    \item At this point, each image $I_i$ has been labelled with a pose $p_i$ and has its camera calibration matrix $K_i$ estimated. A point cloud of the scene featured in $I_i$ has been generated also. This process is often described as adding 3D \textit{structure} to $I_i$.
\end{enumerate}

A \textbf{point cloud} can be described simply as an unordered list of positions in 3D space (typically with a corresponding colour). Point clouds and other 3D environmental representations natively encode the distance information between features --- information which is required for robust localization algorithms. The \textit{resolution} of a point cloud is defined by the distance between any given point and its neighbouring points; if more points cover a region of space then the resolution of the cloud is greater. If the resolution of the point cloud is sufficiently large it is considered ‘dense’, otherwise it is considered ‘sparse’.

\subsection{Multi-View Stereo}

Multi-View Stereo (MVS) is a process which takes pose, camera calibration and visual information from each image in a set, and uses it to create a 3D \textit{dense} point cloud. MVS fuses the depth and normal maps of \textit{multiple} images in a pairwise (stereo) manner, allowing 3D points to be projected from the pixels in an image. 

\begin{figure}[h]
    \centering
    \includegraphics[width=0.95\linewidth]{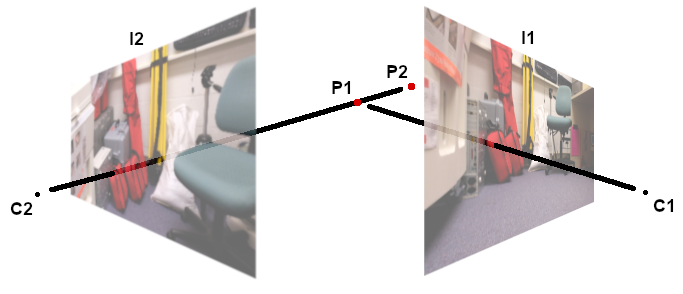}
    
    \caption{A visualisation of a simple MVS process deciding whether to reconstruct a point. Here, the camera $C_1$ has captured the image $I_1$, and has used its depth information to project the pixel (over the red bag in $I_1$) into 3D space at $P_1$, and similar for $P_2$. If $P_1$ and $P_2$ were to lay in close enough positions, then it would be likely that the point belongs in the scene. Using ray tracing ensures that the check for matching pixels only needs to occur along the ray rather than over the entire image, reducing the search space from 2D to 1D \cite{schoenberger2016mvs}. }
    \label{fig:epipolar}
\end{figure}

Specifically, a pixel in an image with a known depth generates a ray between the pixel's projected 3D position and the camera's position (note that the image's depth information and pose is determined in the SfM process). This pixel \textit{likely} corresponds to a point in the actual scene if the ray passes through a corresponding (similar) pixel in a second image projected into 3D (see Figure~\ref{fig:epipolar}). Naturally this process can be repeated for all pixels, thus producing 3D coordinates for each pixel (creating a dense point cloud). In practice, patches are often used rather than pixels, and a variety of \textit{photo-consistency} measures, which estimate the likelihood of correspondence, can be used.

\begin{figure}
    \centering
    \includegraphics[width=0.95\linewidth]{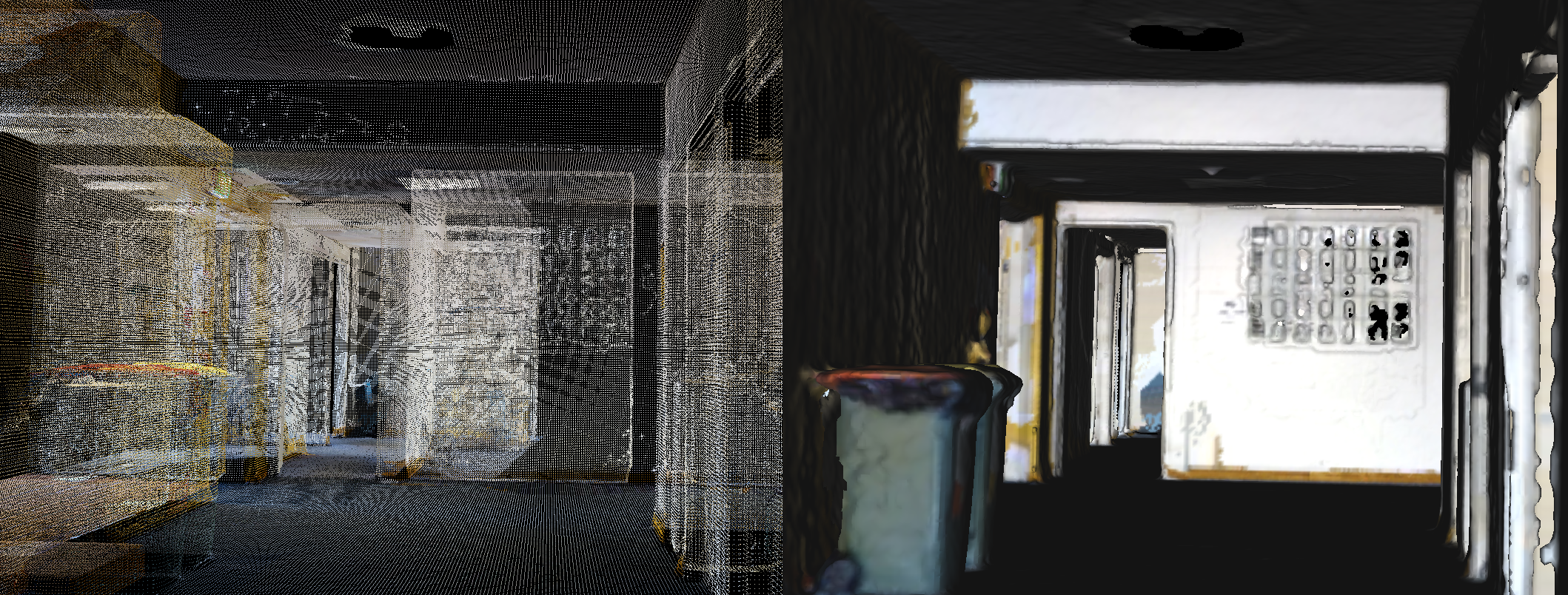}
    
    \caption{Point cloud generated from a LiDAR scanner (left) and the mesh produced by Poisson meshing said point cloud (right). Both images are taken from the same viewpoint, and the continuity of the 3D mesh can be seen when compared to the point cloud, which has spaces between the points.}
    \label{fig:pointcloudmesh}
\end{figure}

Dense point clouds are incredibly valuable in computer vision tasks firstly due to the computational effort required to produce them in an SfM, MVS process and secondly since they can be meshed in order to render synthetic images. Meshes do not have ‘gaps’ between points (unlike point clouds) and thus provide richer, more continuous colour information. The continuity of surfaces found in meshes allow back-tracing methods to be used to render synthetic images from meshes. Large, labelled datasets can be produced as the viewpoint the images are rendered from is known (with respect to the mesh's origin). See Figure~\ref{fig:pointcloudmesh} for a visualization of the differences in visual continuity between point clouds and meshes.

An example pipeline which is referred to extensively in this work is \textit{COLMAP}, which is described as ``a general-purpose SfM and MVS pipeline'' \cite{schoenberger2016sfm, schoenberger2016mvs}. Indeed, the \textit{COLMAP} pipeline uses a patch matching process during MVS, as discussed above.

\section{Image-based localization}
\label{ch:litrev:s:imloc}

To repeat, an object's pose is defined as the combination of its \textbf{position} and its \textbf{rotation}. In the case of a camera, a common task is using the image captured at a given pose to recover said pose. Methods which accomplish this task are broadly referred to as \textit{image-based} localization methods, since they operate on images.

A brief survey of innovative, novel and relevant techniques for determining the pose of a camera is presented in this following chapter. Attention is paid to the processing steps and characteristics of each pipeline, in order to explain the motivation for the pose regression study accomplished in this work.

\subsection{Learning-based approaches}

In Chapter~\ref{ch:intro}, it was stated that deep learning approaches allow complex problems to be re-framed as data-driven tasks. Image-based localization is no exception to this statement. In fact, the pose labelled images produced from the pipelines discussed in Section~\ref{ch:litrev:s:photo} can be used to train deep neural networks or other machine learning architectures. Indeed, the following systems are capable of being trained in such a manner. The majority of the systems here are deep neural networks which take visual image input(s) and regress values for the pose.

\textbf{PoseNet (and derivatives)}. In 2015, Kendall et al. \cite{kendall2015posenet} introduced a 6 degree of freedom re-localization system in the form of a convolutional neural network (CNN) with affine regressors at the output. The system regresses the camera's pose from a single RGB image, and as such does not require hand-crafted feature extraction. The CNN (based on GoogLeNet) performs poorly when trained to regress translation and rotation separately, so it is trained to instead extract 6-dimensional pose labels in one forward pass of the network. Large amounts of training data would be required to train a CNN to be capable of pose regression `from scratch'; which is why their network was pre-trained on the `ImageNet' and `Places' datasets. The results demonstrate that much like in the case of classifier neural nets, regression tasks benefit greatly from transfer learning.


The predictions from PoseNet also demonstrate that it is possible to regress a camera's pose from purely RGB images (without depth information). The original PoseNet produces median pose error of $1.92$m and $2.70$\degree on 7 Scene's `Chess' sequence (more extensive results presented in Table~\ref{tab:benchresults1} and Table~\ref{tab:benchresults2}). Impressively, PoseNet can be ran on a modern laptop computer; the CNN weights file is only $50$MB in size and a single forward pass takes $5$ms --- this is considerably less time and memory required for feature-based image matching localization methods.

The approach notably illustrates that CNN methods generalize well to changes in lighting and motion blur, a property not found in SIFT based methods. This is due to the CNN's ability to extract \textit{high level, complex, robust} features from input images. As a corollory, the authors observe that the network actually \textit{leverages} textureless spaces --- they are just as important as feature rich regions when determining pose. These generalization properties are echoed in \cite{brachmann2017lessmore} and \cite{walch2016posenetlstm}, which suggest that CNN machine learning techniques are well suited to indoor robotic localization problems, as real world indoor environments typically bring about changes in lighting and occlusion.

PoseNet has been built upon to created a number of expanded architectures, which are relavent to this work. In particular, PoseNet had its loss function replaced with a loss function that emphasized the geometric difference between the regressed and ground-truth pose in \cite{kendall2017posenetgeo}. This pipeline (although slower to train) considerably outperformed its predecessor (See Table~\ref{tab:benchresults1}). The augmentation demonstrates that a network's loss function is key in training said network to accurately regress pose --- a notion that is pivotal in this work. 

\textit{Similar} work was done by replacing the later stages of the network with Long Short-Term Memory (LSTMs) networks in \cite{walch2016posenetlstm}. LSTMs store data sequences and identify the most pertinent correlations within the data during training. They achieve this by accumulating `relevant' information automatically in a hidden state and have found success in language processing tasks, and other tasks with long data sequences. In this case, the feature vector output by PoseNet's final fully connected layer is of length $2048$ and the LSTM architectures reduce this dimensionality (in a structured way) drastically to four $128$ length highly-correlated feature vectors. Each of these four feature vectors are submitted to a final set of fully connected layers to complete the pose regression task. 

This novel network achieves $0.24$m and $5.77$\degree median localization errors on \textit{7Scenes Chess} sequence and thus significantly outperforms the state-of-the-art CNN based machine learning localization methods. In particular, it outperforms revised PoseNet results by $29$\% and $5.4$\% in positional and rotational median error respectively. 

\textbf{LessMore}. The work of Brachmann et al. in \cite{brachmann2017lessmore} and \cite{brachmann2016dsac} contribute to the design of a CNN-based method for localization which greatly exceeds state-of-the-art methods in performance. The `Less-More' network (as it will be referred) learns to regress the scene coordinates of input images using RGB images alone --- a 3D model of the scene is not necessary (but is optional) during training. The pipeline is trained end-to-end by using pairs of pose labelled RGB images in three distinct training steps. 


\textbf{Firstly}, a training process which attempts to optimize the scene coordinate predictions of pixels is used to coarsely assign the correct range of scene coordinates to different spatial zones of the 3D scene. In this step a 3D model can be provided which prevents the system from heuristically assuming that all points are equidistant from the camera plane. \textbf{Secondly}, the re-projection error of the system is optimized by minimizing the calculated re-projection with respect to the re-projection calculated from the ground truth pose. \textbf{Finally} the error in the predicted pose with respect to the ground truth label is calculated; this obviously requires \textit{every component in the pipeline to be differentiable}, so differentiable RANSAC or `DSAC' \cite{brachmann2016dsac} is used as an optimization component towards the end of the architecture.

The system's architecture is as such:
\begin{enumerate}
    \item A CNN is used to learn the correspondences between 2D pixels in an input image and 3D points in the underlying scene, i.e. to learn to \textit{regress} scene coordinates for each pixel in a given RGB image.
    \item Using solutions to the `perspective-n-point' (PnP) problem, only four pixels with corresponding scene coordinates in an image are required to compute camera pose. A population of pose hypotheses are thus generated for each input image via the random selection of groups of four pixels.
    \item Each hypothesis is scored using a separate CNN which regresses a score based on the input of local re-projection errors. \textit{Local} re-projection errors are considered so as to not provide information about the global image structure. Hypotheses are selected with a probability derived from their score values. Probabilistic selection is essential to facilitating end-to-end training (as each component in the pipeline must be differentiable).
    \item Hypotheses are refined by determining the number of pixels which would be considered inliers for each high scoring hypothesis. Pose hypotheses are optimized to maximize the inlier count.
\end{enumerate}

Even without a supplied 3D-model,  the system outperforms PoseNet, Active Search and DSAC in almost all cases of the \textit{7Scenes} dataset. In the \textit{Chess} sequence a median localization error of only $0.02$m and $0.7$\degree (which decreases to $0.02$m and $0.5$\degree when a 3D-model is supplied for training) is observed. Unlike other CNN-based methods, this system competes with Active Search methods in performance and works towards bridging the gap in performance between CNN-based methods and feature-based matching methods. Furthermore, this system has all the benefits of CNN methods discussed previously, such as generalizing well to unseen views. A forward pass of the pipeline takes $200$ms for each image, which is considerably longer than a single forward pass of PoseNet, and limits the network's real-time applicability in navigation pipelines on resource constrained hardware.

\subsection{Other algorithms}

Some of the most successful image-based localization pipelines do not use learning architectures, but instead employ traditional methods: feature based matching, comparison methods, \etc  The success of traditional methods in comparison to deep-learning based methods has been noted in many works \cite{khan2018cnn}, however such traditional algorithms tend generalize poorly, and in the case of image-based localization, they are not robust to texture-less surfaces, changes in illumination, or alterations to the scene --- making them a poor fit for localization in complex, domestic environments.

None-the-less, their performance against common benchmarks is often unparalleled. Such pipelines should be studied closely if the gap between traditional and deep-learning based techniques is to be closed, but before these pipelines are considered, it is wise to first briefly touch on the most ubiquitous feature used in these algorithms.

\label{ch:litrev:s:imloc:ss:other:b:sift}
The \textbf{Scale-Invariant Feature Transform} (SIFT) \cite{lowe1999sift} is used prolifically in computer vision tasks and appears in a multitude of pose estimation pipelines. Although an entire survey paper could be dedicated to SIFT and SIFT-based localization approaches \cite{zheng2018siftsurvey}, the topic is only briefly touched on here. In essence, SIFT features are a brand of local image features, and as the name suggests, they are invariant to scaling, translation and rotation, whilst being highly distinctive. SIFT features are calculated over any given image $I$ as follows:
\begin{enumerate}
    \item The image's \textit{scale space} is computed. This is done by convolving $I$ with a Guassian blur filter $G$ as per Equation~\ref{eq:guassianblur}. $G$'s $\sigma$ value is linearly increased in order to create a set of progressively blurred images, at which point $I$ is scaled to half size ($\frac{1}{4}$ area) and another set of blurred images is produced using the same process. The blurred image sets at each scale are referred to as octaves: $4$ octaves and $5$ blur levels ($\sigma$ values) per octave are used in \cite{lowe1999sift}.
    
    \begin{equation}
        \label{eq:guassianblur}
        G(x, y, \sigma) = \frac{1}{2 \pi {\sigma}^{2}} e^\frac{-(x^2 + y^2)}{2 \sigma^2} 
    \end{equation}
    where:
    \begin{conditions}
        (x, y) & the location of each pixel  \\
        \sigma & the scale parameter (quantifies the amount of blur)
    \end{conditions}
    
    \item The Difference of Guassians (DoG) are calculated via simple subtraction of each blurred image with the next blurred image and so on (within each octave). The DoG's are a computationally inexpensive, scale invariant approximation of the Laplacian of Guassians (LoG) which can be used to find key points in an image.
    \item The key points lie at the extrema of the DoGs. The extrema are located at a pixel level via a $26$ neighbour search (using adjacent scales within the octave), so the least and greatest scales are skipped as they do not have adjacent scales. The sub-pixel location is then approximated using a Taylor expansion around the extrema.
    \item Low contrast and edge key points are filtered out by considering the intensity of the gradient's magnitude at the key point's location.
    \item The key points must now be oriented to introduce rotation invariance. The gradient directions and magnitudes around each key point are calculated. The $360$\degree of possible rotations is split into bins, and the magnitude of the gradient is added to the appropriate bin depending on its direction. Bins with values greater than $80\%$ of the maximum bin's value (including the maximum bin) are determined to be key points with orientations equal to the orientation of the bin.
    \item Finally, key points are used to create a feature vector encoding, thus creating a set of SIFT features.
\end{enumerate}

\begin{table*}[h]
    \centering
    
    \begin{tabularx}{\linewidth}    
        { *{4}{P{4.4cm}} }
        
        \includegraphics[width=4.4cm]{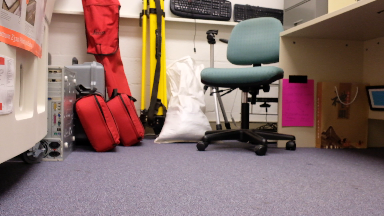} & 
        \includegraphics[width=4.4cm]{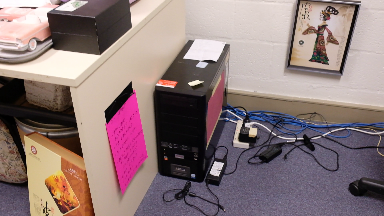} &
        \includegraphics[width=4.4cm]{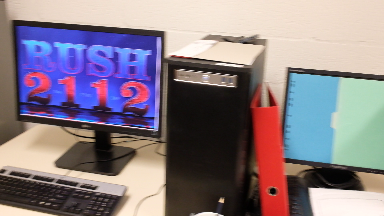} \\
        
        \includegraphics[width=4.4cm]{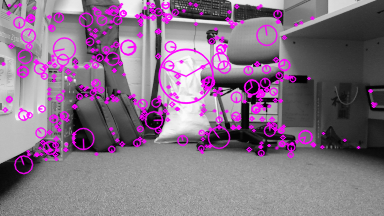} & 
        \includegraphics[width=4.4cm]{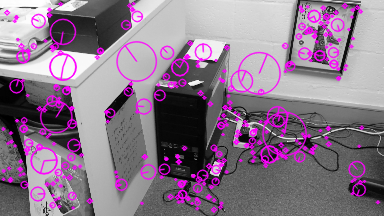} &
        \includegraphics[width=4.4cm]{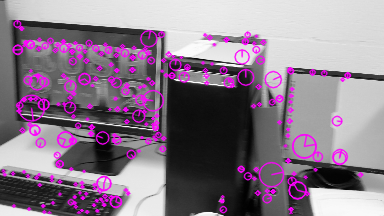} \\
        
    \end{tabularx}
    \caption{ Sample images from the data collected in this work (top row) with their SIFT features computed and displayed (bottom row). The SIFT features have their magnitude and orientation proportional to the size and rotation of the circles. The SIFT features are displayed on a grayscale version of the image, and in pink for visibility. Image best viewed in colour.}
    \label{tab:siftsamples}
    
\end{table*}

A visualization of the SIFT features calculated over an image used in this work's photogrammetry experiments is provided in Table~\ref{tab:siftsamples}. With SIFT summarized, the branch of image localization methods which employ them can now be discussed.

\textbf{Feature-based matching approaches}. SIFT based techniques belong to a broader class of image retrieval techniques, which match the features found in a query image to a relevant location in a pose-labelled image database, thus determining the query image's pose. Specifically, a query image $I_{q}$ (taken at some pose within a scene of interest) has its features (SIFT or otherwise) $F_{q}$ calculated. A database of $N$ images $I_{i}$ from the same scene are labelled with poses $p_{i}$, and have their features $F_{i}$ pre-calculated. $F_{q}$ can be compared to $F_{i}$ for $i = 1, 2, ..., N$ using some measure of feature similarity. If the images are from the same scene and the database adequately covers the environment, then it is likely that there will be an $i = j$ such that $F_{j}$ is the \textit{most similar} to $F_{q}$. Since images with similar features are likely to be `looking at' the same content (due to the distinctive nature of SIFT features), the pose $p_{j}$ is likely to be similar to the unknown pose of $p_{q}$. $p{j}$ can then be returned or refined, localizing $I_{q}$. This simple procedure forms the basis for many feature-based matching techniques.

Sattler \etal introduced one such pipeline in \cite{sattler2011fast2d3d}, in this work, the query images are registered to a 3D scene model, rather than a separate image database. The method establishes correspondences between 2D local features (as above) and \textit{3D} points in a model, thus illustrating how 3D information produced by SfM processes can increase localization accuracy to be comparable to GPS --- even in outdoor scenes. 

This notion is echoed in \cite{sattler2017necessary}, where Image-based algorithms such as DenseVLAD (which uses the DenseVLAD descriptor --- based on the handcrafted RootSIFT descriptor) and NetVLAD (which uses CNN based feature extraction to create VLAD descriptors) are compared to Camera Pose Voting (CPV), a structure-based algorithm which produces state-of-the-art results on the Dubrovnik dataset. It is found that pure image-based methods achieve the lowest accuracy (highest quantile errors) whilst offering efficiency advantages. It is concluded that 3D models are crucial for \textit{accurate} visual localization methods, and that methods which integrate structure-based \textit{and} image-query methods produce results that draw from the strengths of each method. Hence, there is motivation for the SfM processes required for the work proposed in this dissertation.

\textbf{Active Search} is a \textit{specific} example of an algorithm which performs pose estimation using traditional methods \cite{sattler2017efficient}. The algorithm matches 2D features in a query image to 3D features (\ie point descriptors) in a model, and determines sets of points which could be visible in a query image by using co-visibility data --- generated in an SfM 3D reconstruction process --- to filter matches. Again, 3D information proves useful to the pose estimation task; the rate of successful localization using this schema was found to be $91.04\%$ compared to $85.47\%$ for standard query matching methods. Due to its performance, the active search algorithm is used as a benchmark for RGB only, traditional, visual, localization algorithms in Section~\ref{ch:resdis}.

\chapter{Objectives}
\label{ch:obj}

\vspace{-2cm}

As explored in the review, current pose estimation techniques exhibit a range of problems: a lack of robustness to changes in illumination and clutter, poor generalization, and high resource demands to name but a few. Hence, there are multiple avenues for improvement --- especially if the problem domain is constrained to indoor robotic navigation applications.

The objectives of this work thus align closely with areas in the literature which could be further explored \& extended, or could be further improved. They are as follows:
\begin{enumerate}
    \item To use photogrammetry to facilitate the training of a deep pose regression neural network. An accessible environment should be used so that local testing can take place --- obviously it is not possible to test a robotic platform in the real-world environments that feature in existing datasets (\ie  Cambridge Landmarks). Also, current datasets do not give full coverage of enclosed environments --- typically focusing on surveying a scene only in one major direction. For example, in the \textit{7Scenes Chess} scene, the wall `behind' the camera is never imaged, so the final reconstruction includes only the environment viewed in \textbf{roughly} one direction --- this is not useful for a robotic navigation pipeline, as the robot may view any part of the environment, from any location in practice. The Computer Science \& Software Engineering (CSSE) building offices at the University of Western Australia (UWA) are to be used as the basis for these photogrammetry experiments.
    
    \item To improve upon the accuracy of current pose regression networks without impacting the ease-of-training. Robustness to perceptual aliasing \cite{zaval2010aliasing} should also be improved. Current systems consider the error in position and rotation separately when regressing poses, when in fact both quantities have mutual effects on each other \textbf{and} a camera's pose \cite{ward2019augloss}. Such an observation should be considered when formulating a network's loss function; as this phenomenon will impact the weights which are determined throughout the training process (see Section~\ref{ch:methods:s:exp:ss:pose} for more details).
    
    \item To use the pose regression network as the basis for a navigation algorithm. Current navigation algorithms require highly accurate map data or extensive sensor arrays, and image-based techniques are still prone to clutter, illumination, specularity, symmetry and texture-less regions. In indoor, domestic environments, these challenging scenarios may manifest in the form of mess, day-night cycles, shiny surfaces, hallways and plain walls respectively. The image-based navigation algorithm should thus incorporate the generalization qualities of a neural network (hence the pose regression network), whilst still maintaining the reliability of modern navigation methods.
\end{enumerate}

\begin{figure}[H]
    \centering
    \includegraphics[width=0.9\linewidth]{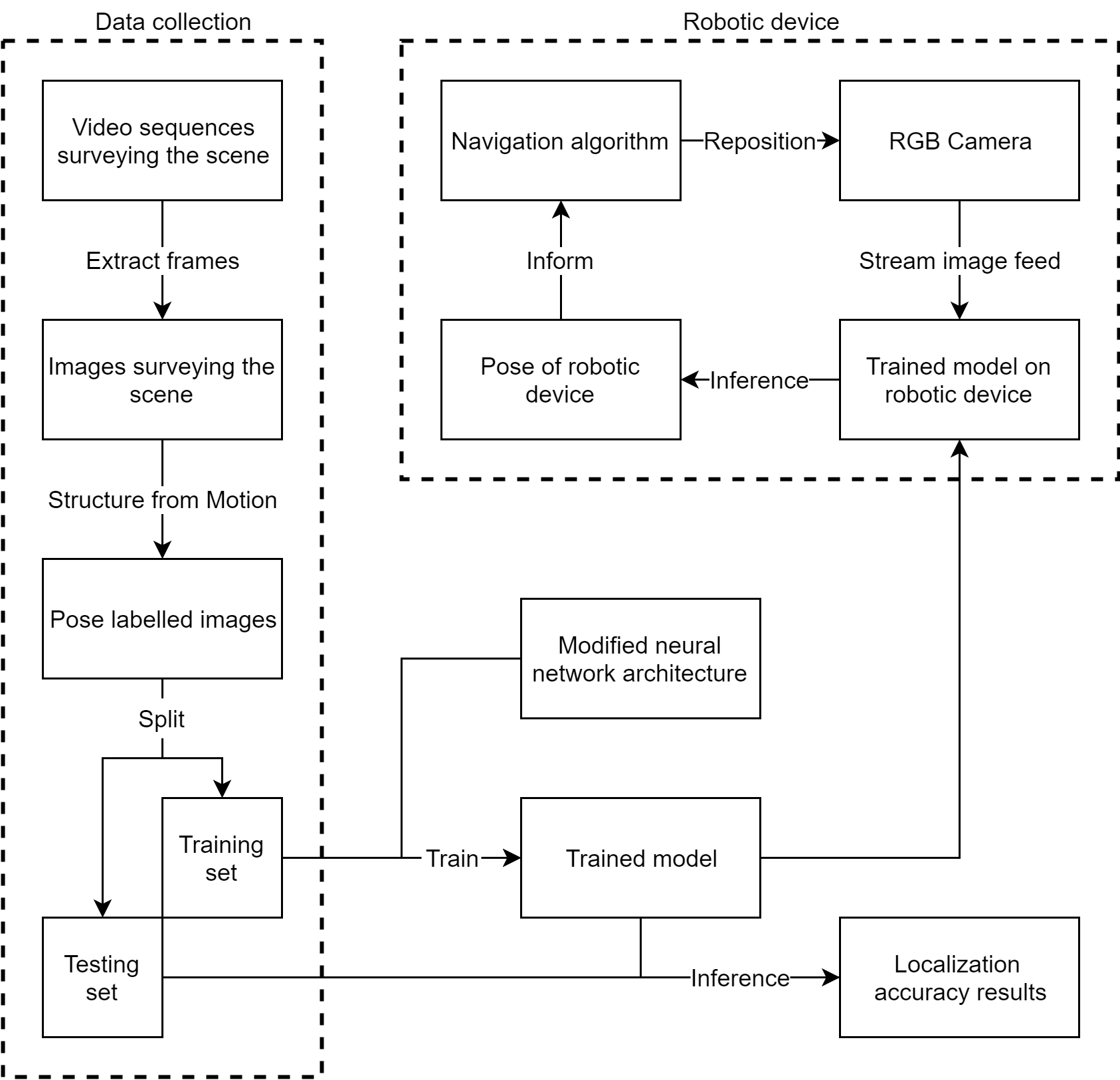}

    \caption{ The entire data collection, model training, and robotic navigation pipeline. Note that the only input required are the video sequences surveying the scene --- in practice these captured sequences need to total only $320$ seconds. }
    \label{fig:fullpipeline}
\end{figure}

These three key objectives form the basis for a pipeline which allows a robotic navigation algorithm to be trained directly from a set of images which adequately survey any given scene, as in Figure~\ref{fig:fullpipeline}. 

\chapter{Datasets}
\label{ch:data}

In order to adequately train and test any proposed pose regression algorithm, two considerations must be made in order to appropriately select data:  
\begin{enumerate}
    \item How does the proposed algorithm perform with respect common benchmarks? Answering this is necessary in order to provide a fair, consistent and scientific comparison between existing algorithms and the proposed.
    \item How does the proposed algorithm perform in the context it was designed for? In this case, how well does the proposed algorithm perform in real-world, indoor contexts (for the purpose of robotic navigation). A key objective of this work is to answer this question.
\end{enumerate}

\section{Existing datasets}
\label{ch:data:s:exist}

\begin{sidewaystable}
    \centering
    
    \begin{tabularx}{\linewidth}    
        { *{7}{P{2.7cm}} }
        
        \includegraphics[width=2.7cm]{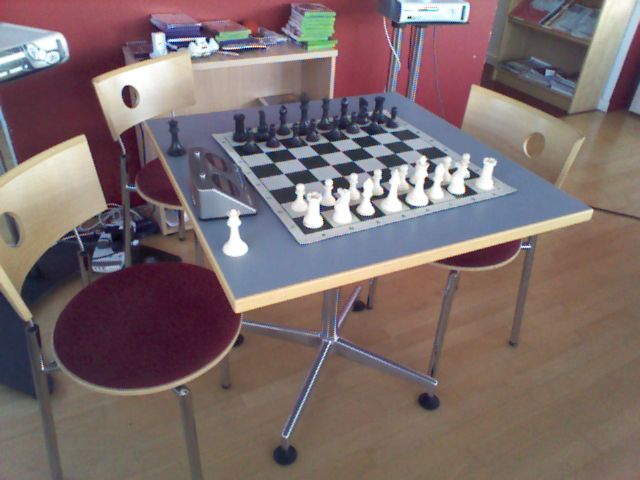} & 
        \includegraphics[width=2.7cm]{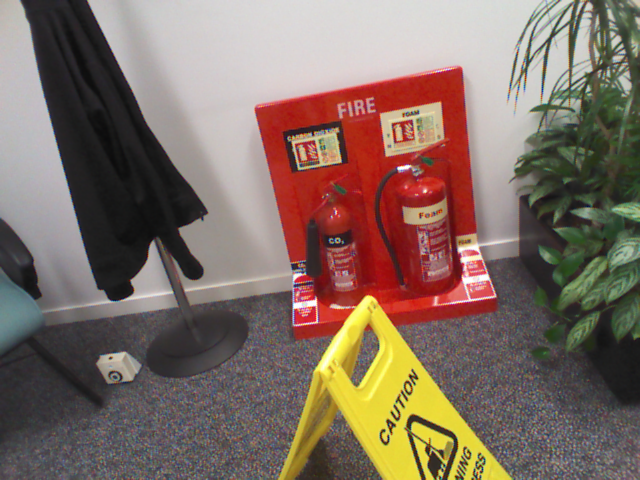} &
        \includegraphics[width=2.7cm]{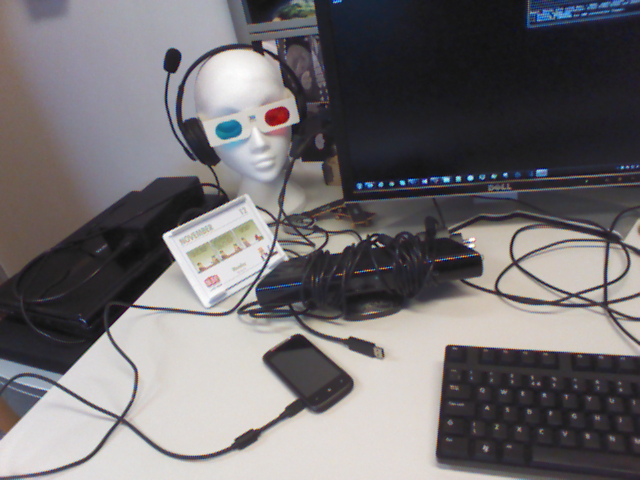} &
        \includegraphics[width=2.7cm]{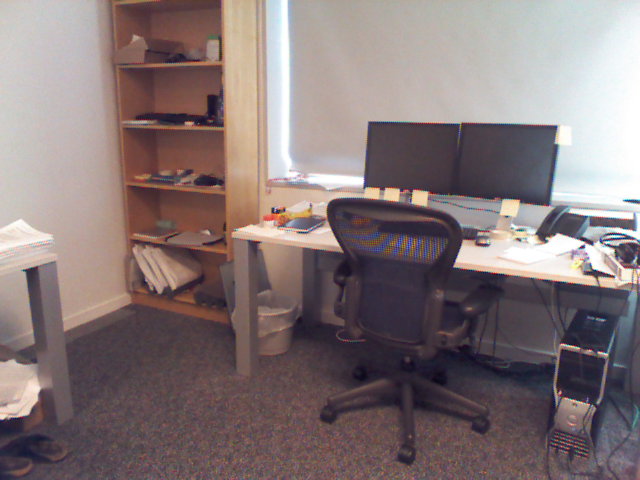} &
        \includegraphics[width=2.7cm]{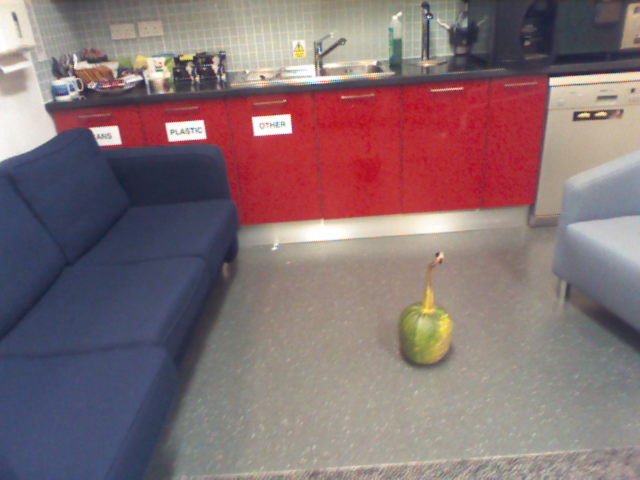} &
        \includegraphics[width=2.7cm]{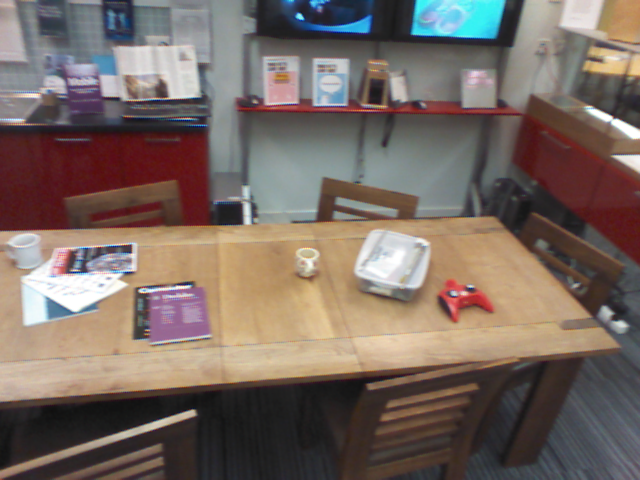} &
        \includegraphics[width=2.7cm]{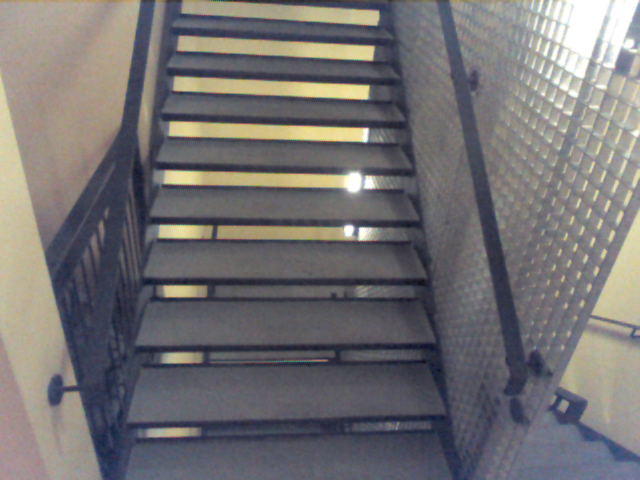} \\
        
        Chess & Fire & Heads & Office & Pumpkin & Red Kitchen & Stairs \\
        
    \end{tabularx}
    \captionsetup{justification=centering}
    \caption{ Sample images from each of the 7 scenes in the \textit{7Scenes} dataset. }
    \vspace{5mm}
    
    \begin{tabularx}{\linewidth}    
        { *{6}{P{3.22cm}} }

        \includegraphics[width=3.22cm]{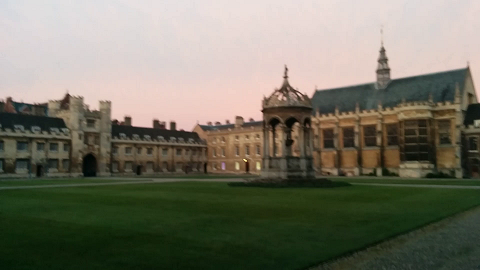} & 
        \includegraphics[width=3.22cm]{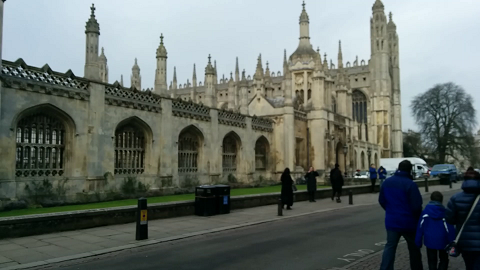} &
        \includegraphics[width=3.22cm]{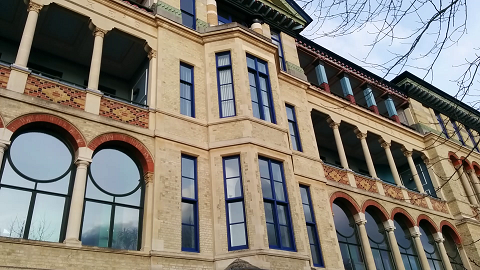} &
        \includegraphics[width=3.22cm]{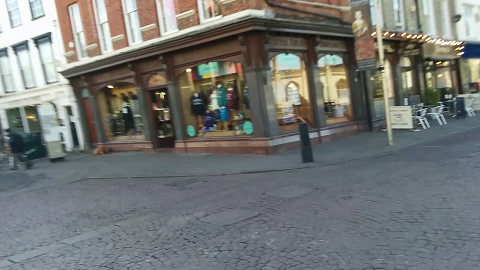} &
        \includegraphics[width=3.22cm]{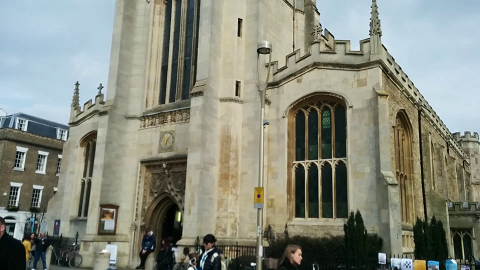} &
        \includegraphics[width=3.22cm]{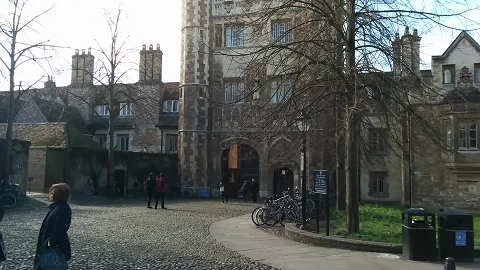} \\
        
        Great Court & Kings College & Old Hospital & Shop Facade & St Mary's Church & Street \\
        
    \end{tabularx}
    \captionsetup{justification=centering}
    \caption{ Sample images from each of the 6 scenes in the \textit{Cambridge Landmarks} dataset.}
    \vspace{5mm}
    
    \begin{tabularx}{\linewidth}    
        { *{5}{P{3.96cm}} }
        
        \includegraphics[width=3.96cm]{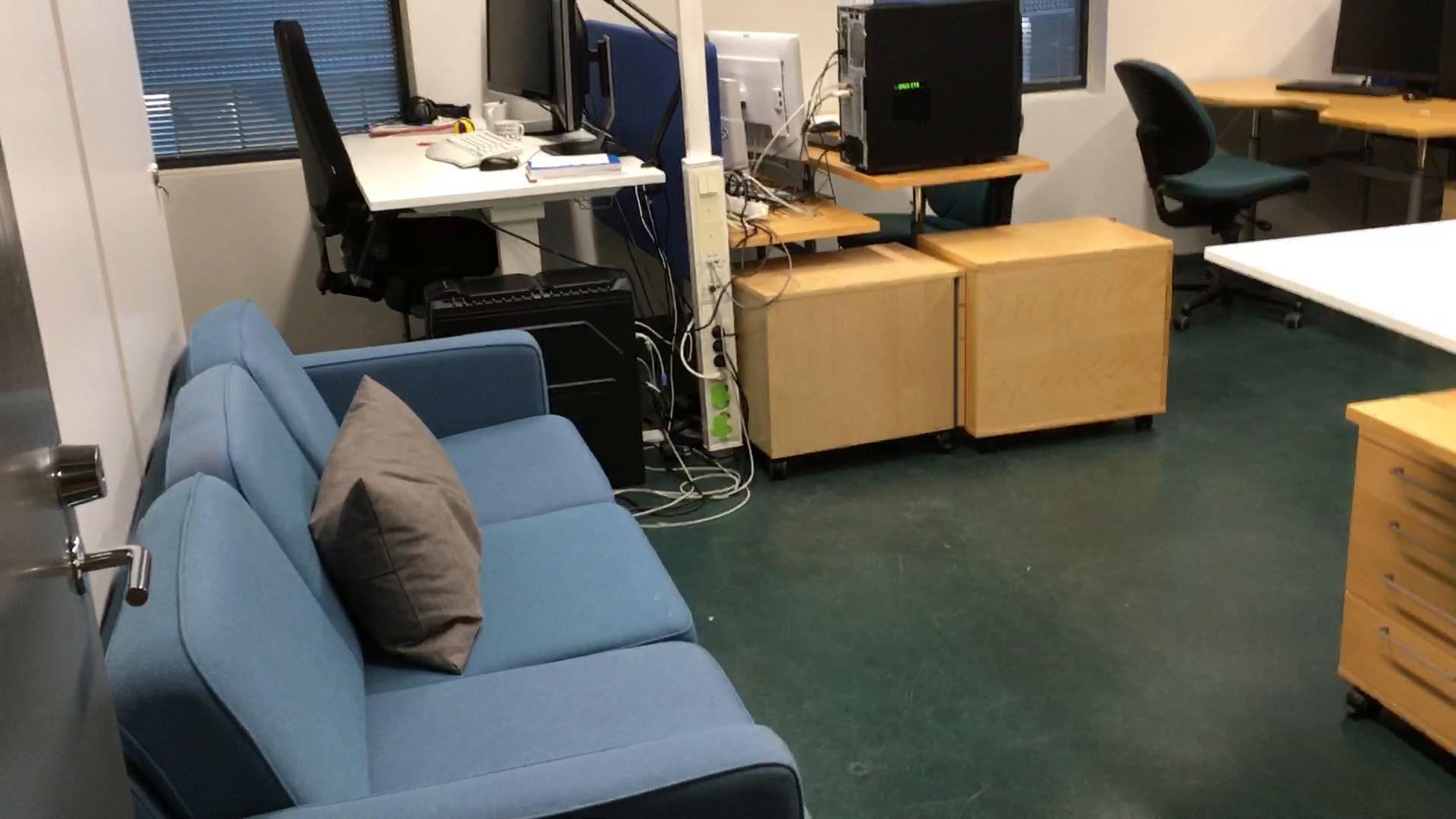} & 
        \includegraphics[width=3.96cm]{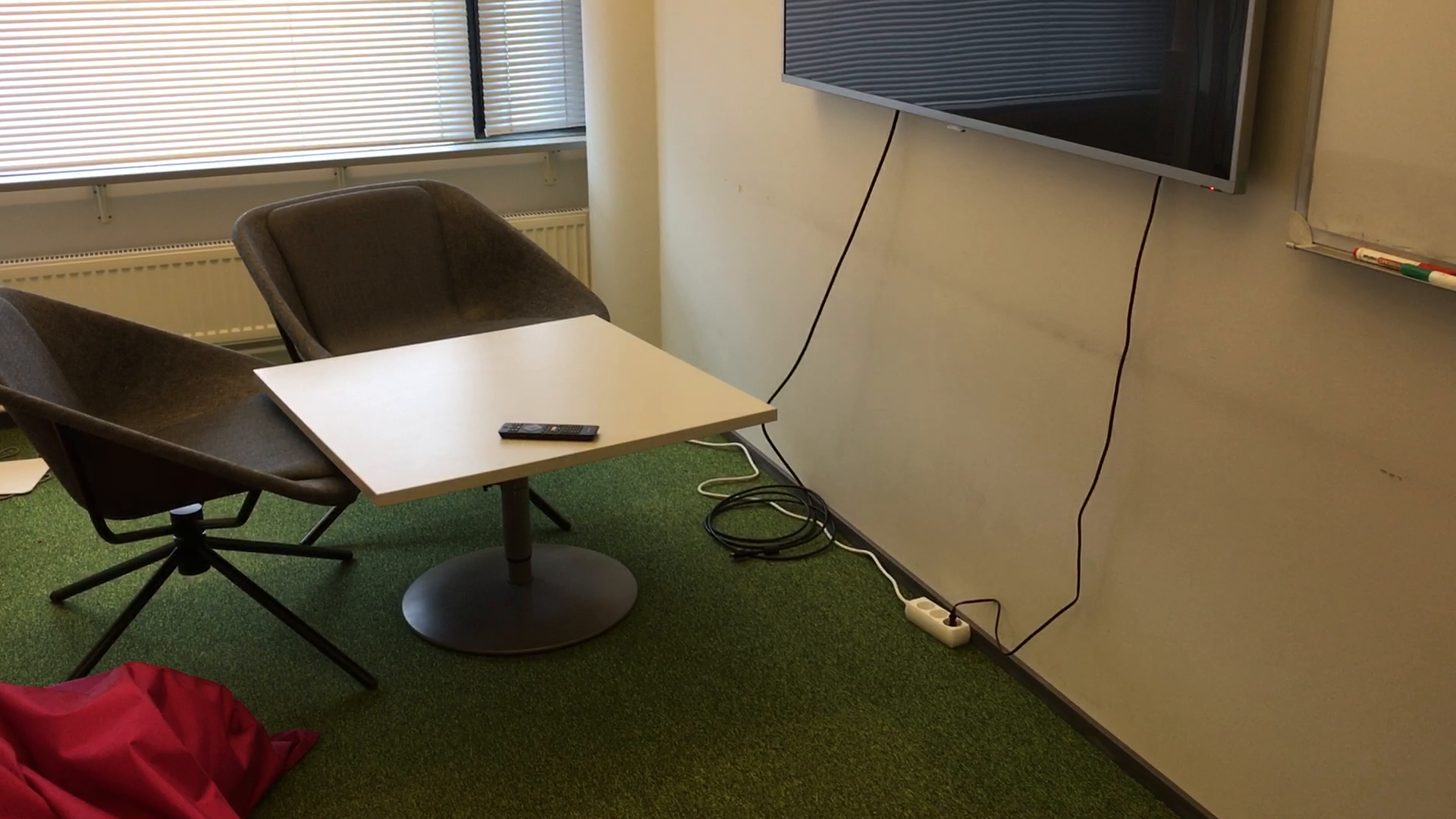} & 
        \includegraphics[width=3.96cm]{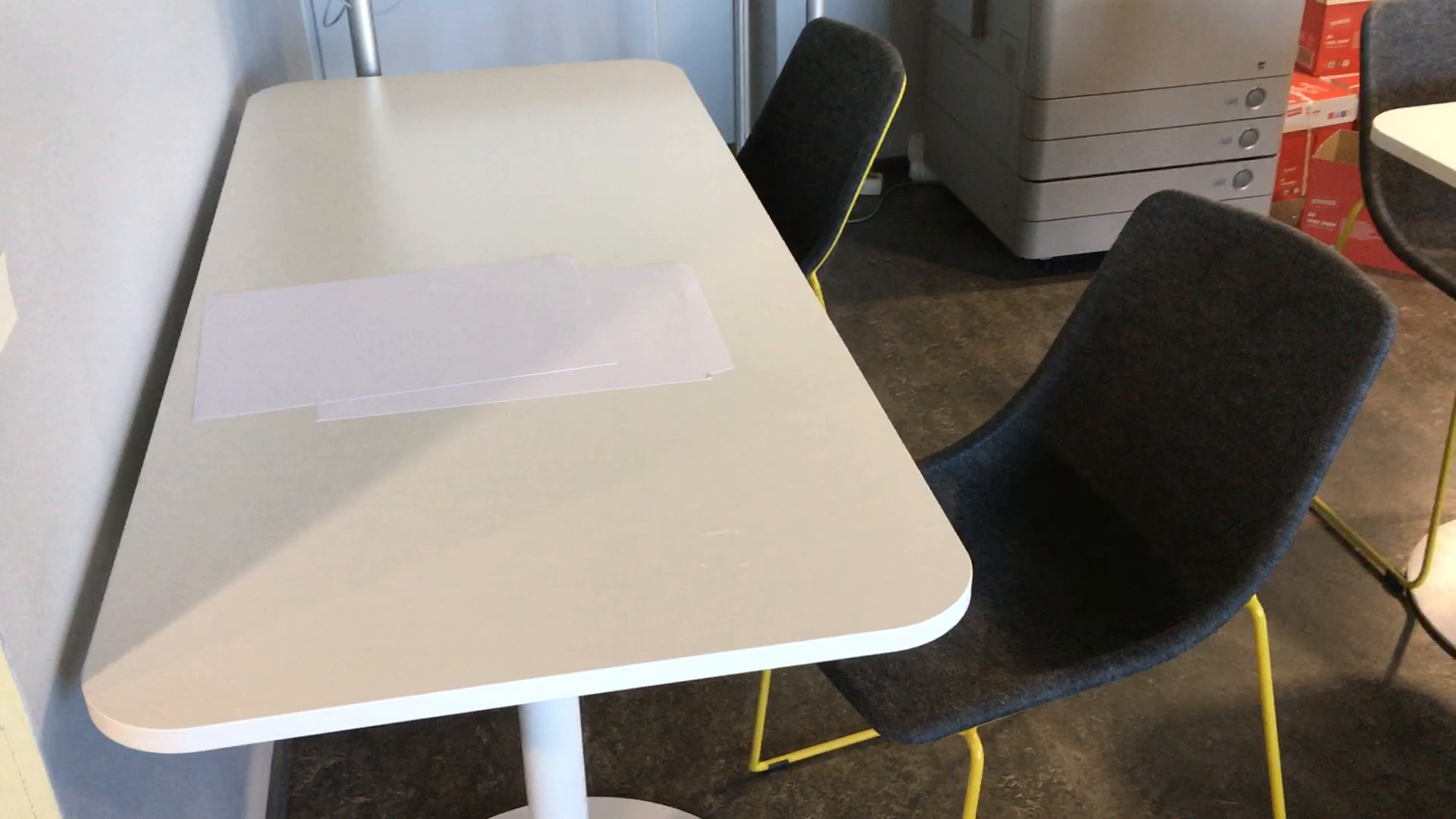} & 
        \includegraphics[width=3.96cm]{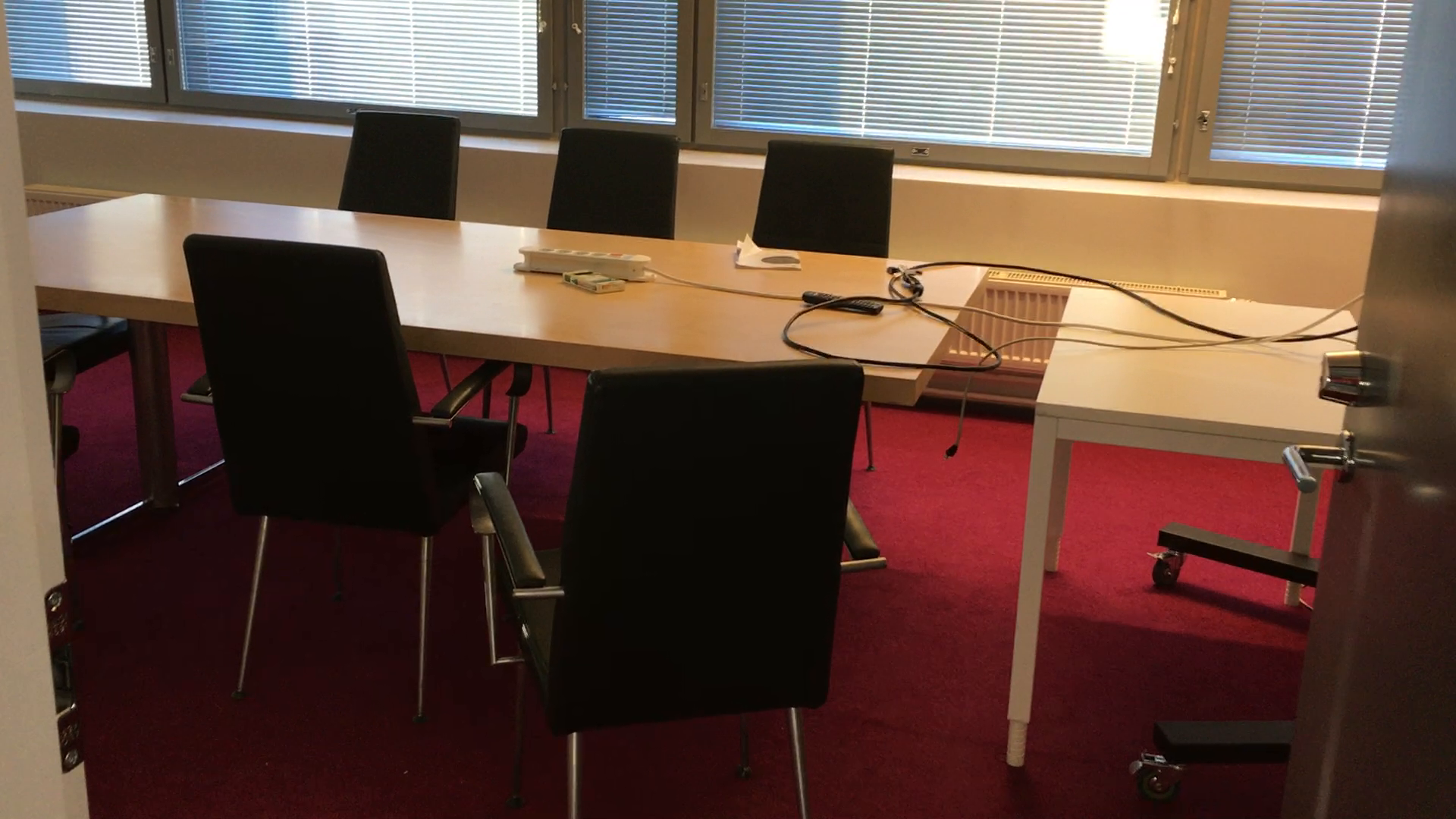} & 
        \includegraphics[width=3.96cm]{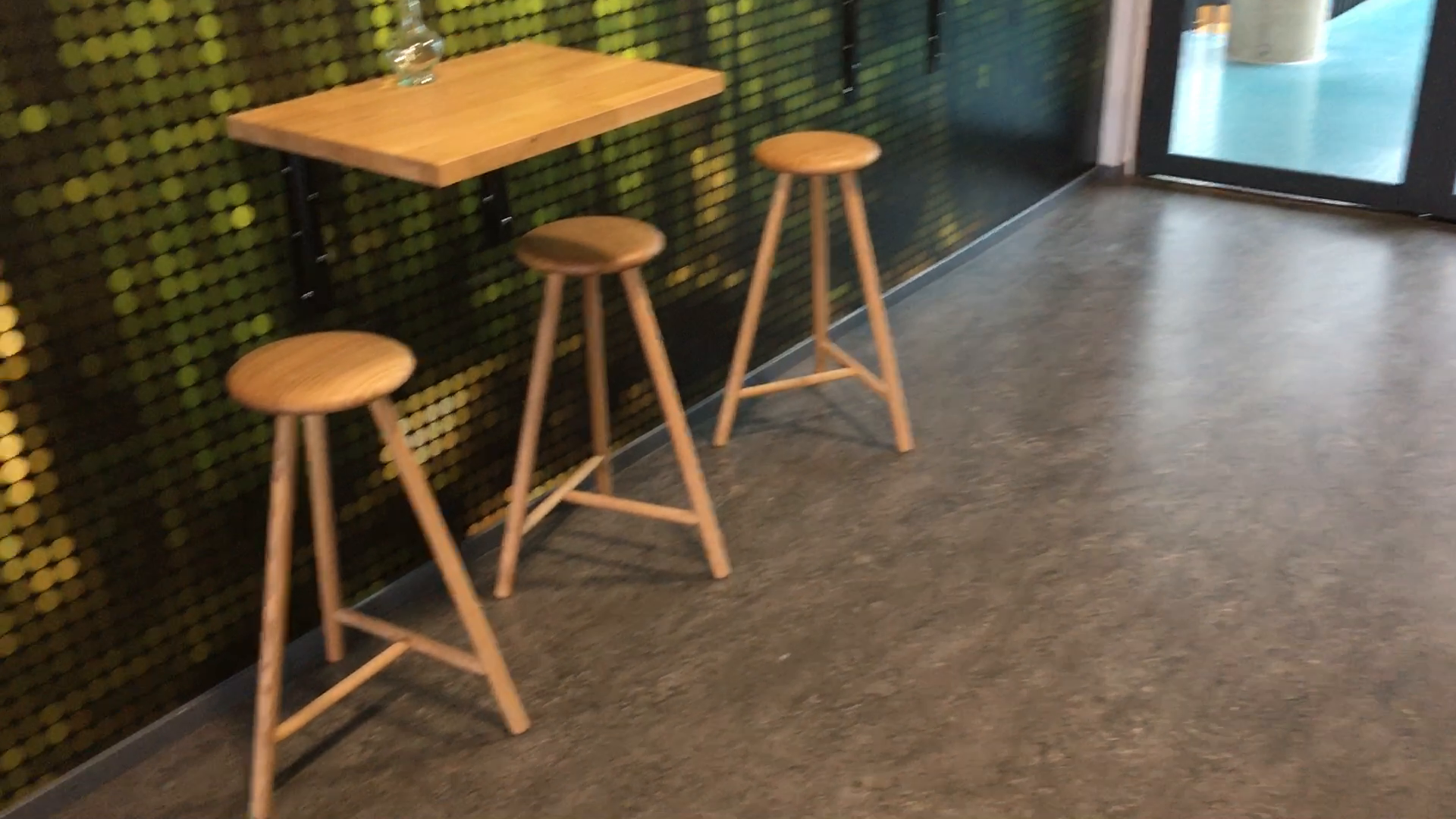} \\
        
        Office & Meeting & Kitchen & Conference & Coffee Room \\
        
    \end{tabularx}
    \captionsetup{justification=centering}
    \caption{ Sample images from each of the 5 scenes in the \textit{University} dataset. }
    \vspace{5mm}
    
    \begin{tabularx}{\linewidth}    
        { *{6}{P{3.22cm}} }

        \includegraphics[width=3.22cm]{img/photogrammetry/sift/1-resize.png} & 
        \includegraphics[width=3.22cm]{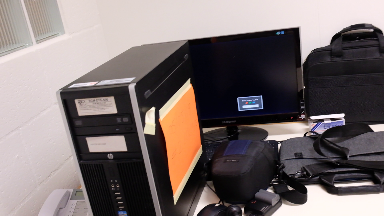} & 
        \includegraphics[width=3.22cm]{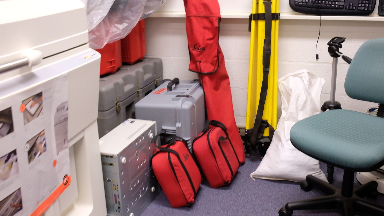} & 
        \includegraphics[width=3.22cm]{img/photogrammetry/sift/4-resize.png} & 
        \includegraphics[width=3.22cm]{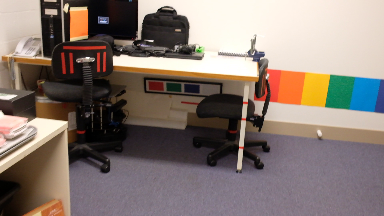} & 
        \includegraphics[width=3.22cm]{img/photogrammetry/sift/6-resize.png} \\
        
    \end{tabularx}
    \captionsetup{justification=centering}
    \caption{ $6$ sample images from the \textit{UWA-lab} scene. }
    \vspace{5mm}
    
    \label{tab:datasetsamples}
\end{sidewaystable}

In choosing the \textit{existing} datasets, consideration is payed mainly to how prolific their use is in the literature --- more common datasets are more likely to be considered in the testing of competing systems. The following datasets are thus chosen as they allow for a broader comparison of the proposed algorithm(s) against existing algorithms in a similar class.

\textbf{7Scenes} \cite{shotton2013score}. $7$ indoor locations in a domestic office context. The dataset features large training and testing sets (in the thousands). The camera path moves continuously through the scene in order to gather images in distinct sequences. Images include motion blur, featureless spaces and specular reflections (see Table~\ref{tab:hardframes}), making this a challenging dataset, and one that has been used prolifically in the camera relocalization literature. The ground truths poses are gathered with KinectFusion, and the RGB-D frames have a resolution of $640\times480$px. 

\textbf{Cambridge Landmarks} \cite{kendall2015posenet, kendall2015posenetuncertain, kendall2017posenetgeo}. $6$ outdoor locations in and around Cambridge, The United Kingdom. The larger spatial extent and restricted data size make this a challenging dataset to learn to regress pose from --- methods akin to the one presented in this work typically only deliver positional accuracy in the scale of metres. In the context of \textit{domestic} robotics and other indoor applications, the results against this benchmark are less important, but the dataset does provide a common point of comparison, and also features large expanses of texture-less surfaces. Ground truth poses are generated by a SfM process, so some comparison can be drawn between this dataset and the one(s) created in this work.

\textbf{University} \cite{laskar2017camera}. $5$ indoor scenes in a university context. Ground truth poses are gathered using odometry estimates and ``manually generated location constraints in a pose-graph optimization framework'' \cite{laskar2017camera}. The dataset, similarly to \textit{7Scenes}, includes challenging frames with high degrees of perceptual aliasing, where multiple frames (with different poses) give rise to similar images \cite{zaval2010aliasing}. Although the scenes are registered to a common coordinate system in the \textit{University} dataset and thus a network \textit{could} be trained on the full dataset, the models created in this work are trained and tested \textit{scene-wise} for the purpose of consistency.

The quantitative sizes of these datasets are explicitly outlined in Table~\ref{tab:datasetsizes}, and samples can be viewed in Table~\ref{tab:datasetsamples}.

\section{Proposed dataset}
\label{ch:data:s:new}

A new dataset is generated in order to facilitate local testing of the proposed algorithm; scenes are based in the CSSE building offices at the UWA. The design of this dataset considers the context of indoor robotic navigation, in that the environment considered is an indoor office context. In order to match the datasets described above, and to provide ample instances to train and test on, approximately $3288$ images are taken of the lab environment. As discussed in Section~\ref{ch:litrev:s:photo}, the data is labelled through a photogrammetry process --- this is more desirable than using KinectFusion (which requires a hand-held depth camera) or using odometry from a mechanical arm (which requires expensive hardware), as it is a `hands-off' approach that adds to the pipeline's ease-of-use (see Figure~\ref{fig:fullpipeline}).

\subsection{Data collection}
\label{ch:data:s:new:ss:coll}

For a full list of the photogrammetry experiments attempted in this work, see Section~\ref{ch:methods:s:exp:ss:photo}, as the focus in this section is on the scene which was considered the most successfully reconstructed (this will herein be referred to as the UWA-lab scene).

Video sequences are taken using a FujiFilm X-T20 with a 23mm prime autofocus lens in order to ensure a fixed calibration matrix between sequences. The camera's auto-focus is used during the capture of video sequences to ensure that images detect fine detail. Images are taken at a resolution of $1920\times1080$px, and a high aperture (f$11$) is used to ensure a large and well focused depth of field. The high resolution and level of focus in these images improve reconstruction in areas which only have very fine details which would be smoothed out with motion blur: the spackled texture of plain white walls, the matte finishes on metal and so forth. 

The $15$ video sequences for the UWA-lab scene are recorded at $50$ frames-per-second (to reduce motion blur), and total $\sim 330$ seconds of captured video, or $16440$ individual frames/images. In order to increase the baseline between images and reduce the size of the dataset to be comparable with existing datasets, the $5^{th}$ frame from each video sequence is extracted --- hence the $3288$ images mentioned above (see Table~\ref{tab:datasetsamples}).

Note that the UWA-lab scene was designed to be comparable in size to the \textit{7Scenes} and \textit{University} datasets on average. The number of images in the scene is also limited for two reasons: to limit the amount of time a user would have to spend manually recording video and to ensure the dataset is sufficiently challenging. Sample images gathered can be viewed in Table~\ref{tab:datasetsamples}.


These images are used as the input to \textit{COLMAP}'s SfM, MVS pipeline \cite{schoenberger2016sfm}, thus producing pose labels via the method described in Section~\ref{ch:litrev:s:photo}. These labels are considered ground truth for the training processes which occur in later experiments. The resulting dense reconstruction can be viewed in Table~\ref{tab:denseresults}. Further details regarding the choices made during data collection and dataset design can be viewed in Chapter~\ref{ch:methods}.

\begin{table}
    \centering
    \begin{tabular}{ l | c | c c }
        
        \hline\hline
        
                    & Extents   &       &       \\ 
        Scene       & (metres)  & \# Train & \# Test  \\ 
        
        \hline\hline
        
        Chess   & $3\times2\times1$     & 4000  & 2000 \\ 
        Fire    & $2.5\times1\times1$   & 2000  & 2000 \\ 
        Heads   & $2\times0.5\times1$   & 1000  & 1000 \\ 
        Office (7Scenes) & $2.5\times2\times1.5$ & 6000  & 4000 \\ 
        Pumpkin & $2.5\times2\times1$   & 4000  & 2000 \\  
        Red Kitchen & $4\times3\times1.5$   & 7000  & 5000 \\ 
        Stairs  & $2.5\times2\times1.5$ & 2000  & 1000 \\ 
        \hline
        Average & $2.7\times1.9\times1.2$ & 3714  & 2429 \\ 
        
        \hline\hline
        
        Office (University) & $7\times4.5$   & 2196 & 1099 \\ 
        Meeting      & $6.5\times2$   & 1701 & 945  \\ 
        Kitchen      & $6\times7$     & 2076 & 990 \\ 
        Conference   & $5.5\times7.5$ & 1838 & 949 \\ 
        Coffee Room  & $6\times9$     & 2071 & 959 \\ 
        \hline
        Average      & $6.2\times6$ & 1976 & 988 \\ 
        
        \hline\hline
        
        Great Court  & $95\times80$   & 1532 & 760  \\ 
        King's College & $140\times40$  & 1220 & 343  \\ 
        Old Hospital & $50\times40$   & 895  & 182  \\ 
        Shop Facade  & $35\times25$   & 231  & 103  \\ 
        St Mary's Church & $80\times60$   & 1487 & 530  \\ 
        Street       & $500\times100$ & 3015 & 2923 \\ 
        \hline
        Average      & $150\times58$ & 1397 & 807  \\ 
        
        \hline\hline
        
        UWA-lab     & $3.3\times2.7\times4.6$   & 2288 & 1000 \\ 
        
        \hline\hline
        
    \end{tabular}
    
    \caption{ The size metrics of the \textit{7Scenes}, \textit{University} and \textit{Cambridge Landmarks} datasets. The \textit{UWA-lab} scene's size metrics are also provided. ($x \times y \times z$) dimensions are given where possible, otherwise ($x \times z$) dimensions are given (where the $y$ axis is the axis perpendicular to the ground). }
    \label{tab:datasetsizes}
    
\end{table}

\chapter{Methodology}
\label{ch:methods}

This section provides an insight into the how the research in this dissertation was conducted: the background, motivation and justification of the experimental methods used are detailed, and the individual experiments are themselves outlined. 

The performance criteria for the proposed systems are also covered in this section, in order to make clear why certain choices were made regarding the design of algorithms, and the choices made in performance trade-offs\footnote{The explicit hardware and software requirements are enumerated in the Appendix~\ref{ap:reqs}}.


\section{Performance metrics}
\label{ch:methods:s:perf}

The results procured by any experiment need to be discussed with respect to some quantifiable performance criteria for a proper scientific investigation. The performance metrics chosen to evaluate the \textbf{pose regression system} are detailed here:

\begin{itemize}
    \item \textbf{Accuracy}: the system should be able to regress a camera's pose with a level of positional and rotational accuracy that is competitive with other pipelines in the same class. Accuracy is reported using per-scene median positional and rotational error, and by viewing the distribution of these errors.

    \item \textbf{Robustness}: the system should be robust to perceptual aliasing, motion blur and other challenges posed by the datasets. The system should also be robust to noise and lower quality input data, as these are properties of the scans acquired by the Kinect camera.

    \item \textbf{Resource intensiveness}: evaluation should occur in real-time ($\sim 30$ frames per second), and the system should be suitable in hardware limited real-time applications, \eg, when operating on a small robotic device or notebook laptop.

    \item \textbf{Ease-of-training}: the baseline network here is the default PoseNet \cite{kendall2015posenet}, which can be trained with only RGB images, in $\sim 10$ hours on the GPU hardware used here (see Table~\ref{tab:hardware} for explicit hardware details). The network developed in this work should be similarly easy to train, and not encumbered by modifications to its architecture or operation.

\end{itemize}

If the pose regression system has these attributes, it should then be fit for handling the \textit{localization} required by the robotic navigation pipeline, which can use the poses predicted over a period of time to inform its pathing, avoidance and so forth. The only performance attribute which needs to be formally considered when evaluating the navigation module is its \textbf{resource intensiveness}, as defined above, as the other attributes will be inherited from the pose regression network.

Since a key idea here is to be able to create a \textit{usable} robotic navigation system \textit{automatically} from only a set of images surveying any given scene, the \textbf{ease-of-use} of the entire pipeline (photogrammetry\textrightarrow pose regression training\textrightarrow navigation algorithm as per Figure~\ref{fig:fullpipeline}) will be considered also, referring to the amount of manual input required, the computing time and the pipeline's likelihood of failure.

\section{Experimental protocols}

A full list of all the relevant experiments conducted throughout this project is provided here. The background, motivation of the methods are detailed, as well as a justification of the chosen procedures. The results are presented and discussed in Chapter~\ref{ch:resdis}. 

\subsection{Initial investigations} 
\label{ch:methods:s:exp:ss:init}

The research conducted in this work primarily relates to pose regression networks and how they can be applied to a robotics context, but the work was initially catalyzed by a variety of initial investigations. These exact experiments are outlined in this section. \textbf{The results of these initial investigations are also presented and discussed in this section}, firstly to provide the motivation for the more extensive experiments (outlined in Section~\ref{ch:methods:s:exp:ss:pose} and Section~\ref{ch:methods:s:exp:ss:rnav}), and secondly to reserve Chapter~\ref{ch:resdis} for only the most systematic, important and extensive results. 

These short, proof-of-concept investigations --- which tested the viability of certain approaches --- were numerous, so only the most relevant and informative experiments are outlined here. Generally these experiments failed, but they illustrated possible improvements which were then pursued during the more extensive, important experiments. 

\subsubsection{\textbf{Experiment 0.1}: LiDAR data rather than photogrammetry}

Here, the viability of using Light Detection and Ranging (LiDAR) scans of the CSSE building interior at UWA are investigated. Rather than attaching a pose label to an image (as in a SfM process), ground truth labels are generated using synthetic renderings. The LiDAR data is gathered using a Minolta Vivid Series 3D laser scanner (this scan data was gathered prior to the commencement of this project). This produces a point cloud which is meshed (Poisson meshing) to produce a continuous surface. The formal difference between a point cloud and a mesh is outlined in Section~\ref{ch:litrev:s:photo}, and some example LiDAR data can be seen in Figure~\ref{fig:pointcloudmesh}.

At this point, a virtual camera is placed in the virtual space and a continuous path for the camera is designed which places the camera at known poses. A synthetic rendering of the mesh from this known pose is captured, thus providing labelled, synthetic images. 

This experiment was designed to test the viability of synthetic renderings in regressing pose, and indeed there has been some success using these types of methods in the literature \cite{zhu2016targetdrl, tongloy2017async}. These images are used as training and testing data against the LessMore network (see Section~\ref{ch:litrev:s:imloc} for operation details), producing median position and rotation errors of $1.26$m and $21.7$\degree respectively.

These results are far below the reported accuracy of the LessMore network on the \textit{7Scenes} dataset (less than $0.05$m, $5$\degree median errors are expected), so the synthetic renderings approach was not pursued any further. Meshing artifacts, and the relative sparsity of the point cloud likely contributed to this poor localization accuracy. This experiment resulted in the pursuing of photogrammetry based techniques, informing the first key objective of this document -- that is, to use photogrammetric processes rather than LiDAR based techniques when gathering data (outlined in Chapter~\ref{ch:obj}).

\subsubsection{\textbf{Experiment 0.2}: Photogrammetry (at the CSSE building scale)}

As a direct corollary of the results of the previous experiment, a photogrammetry approach was pursued to construct the labelled dataset. An SfM process allows labels to be attached to real-world, non-synthetic images (which are higher in quality and lower in meshing artifacts) and produces a denser point cloud (though a dense point cloud is not strictly necessary for training a pose regression network). 

\begin{table*}[h]
    \centering
    
    \begin{tabularx}{\linewidth}    
        { *{4}{P{6.7cm}} }
        
        \includegraphics[width=6.7cm]{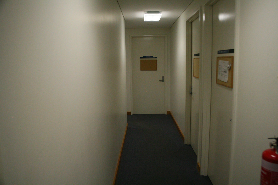} & 
        \includegraphics[width=6.7cm]{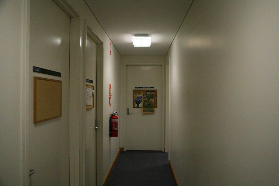} \\
        
        \includegraphics[width=6.7cm]{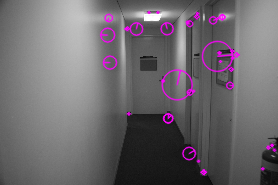} & 
        \includegraphics[width=6.7cm]{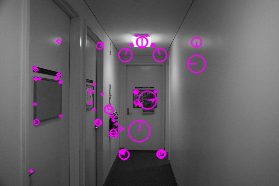} \\
        
    \end{tabularx}
    \caption{ An example of the visual similarity found in hallways. Additionally, the effect of large texture-less regions and low contrast can be observed in the relatively low population of SIFT features (displayed on a grayscale version of each image, and in pink for visibility). Image best viewed in colour.}
    \label{tab:hallsym}
    
\end{table*}

Initially the entire West Wing of the first floor of the CSSE building at the UWA was surveyed using a DSLR camera, but the reconstruction was unsuccessful. This was due to the visual symmetry in the buildings hallways (see Table~\ref{tab:hallsym}), the texture-less walls and in general, the scale of the building --- the reconstruction was unable to register images throughout the hallway, resulting in only a subset of the West Wing being reconstructed. Separating the reconstructions into three smaller rejoins (which could then be joined manually) still failed, primarily due to the visual symmetry found in hallways. These failed sparse reconstructions for the West Wing can be viewed in Table~\ref{tab:failedwestwing}. 

This outcome guided the photogrammetry experiments outlined in Chapter~\ref{ch:data} \& Section~\ref{ch:resdis:s:photo}, causing them to be constrained to smaller environments, which were visually richer in texture, and promoted more successful reconstructions. These experiments ensured that the first key objective in this document --- the use of photogrammetric techniques --- was satisfied (said objective is outlined in Chapter~\ref{ch:obj}).

\begin{table*}[h]
    \centering
    
    \begin{tabularx}{\linewidth}    
        { *{3}{P{4.5cm}} }
        
        \includegraphics[width=4.5cm]{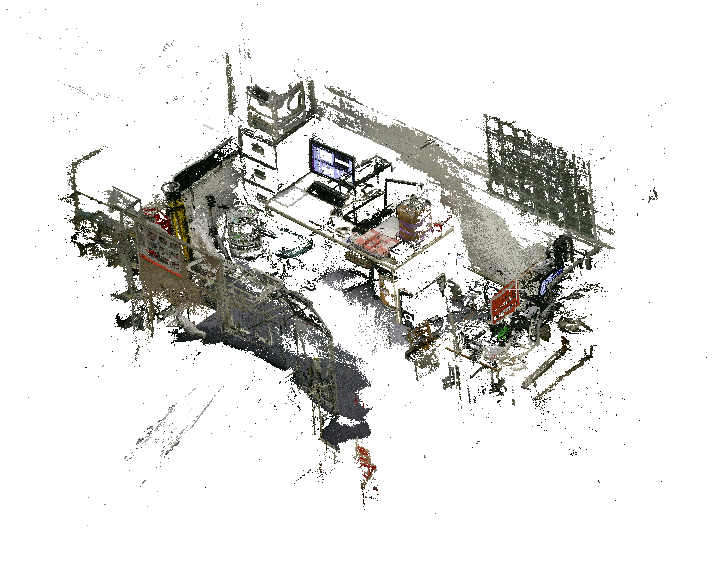} &
        \includegraphics[width=4.5cm]{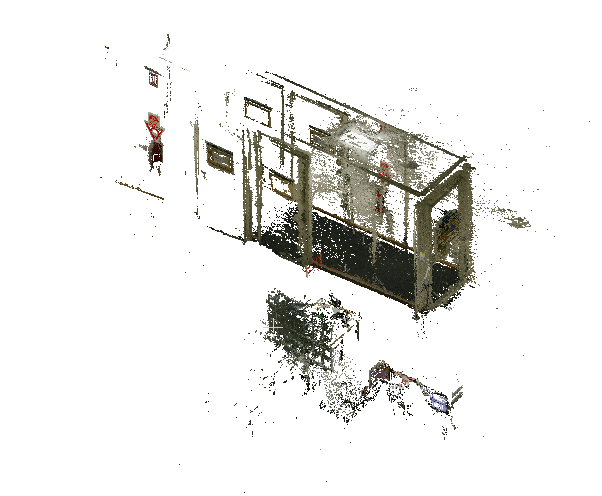} &
        \includegraphics[width=4.5cm]{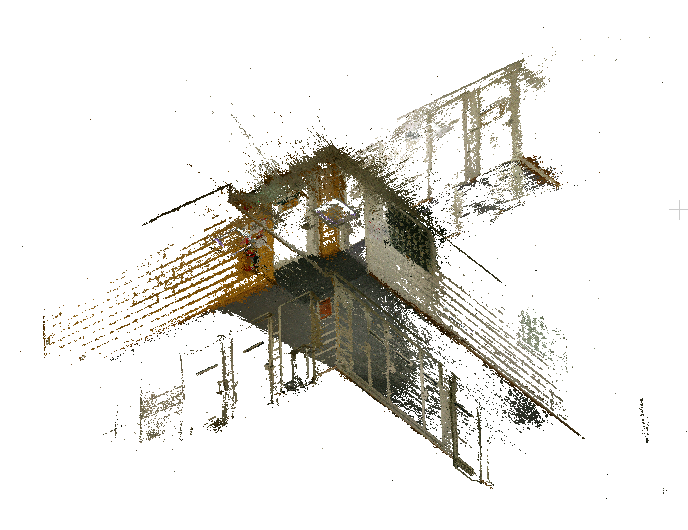} \\
        
        \includegraphics[width=4.5cm]{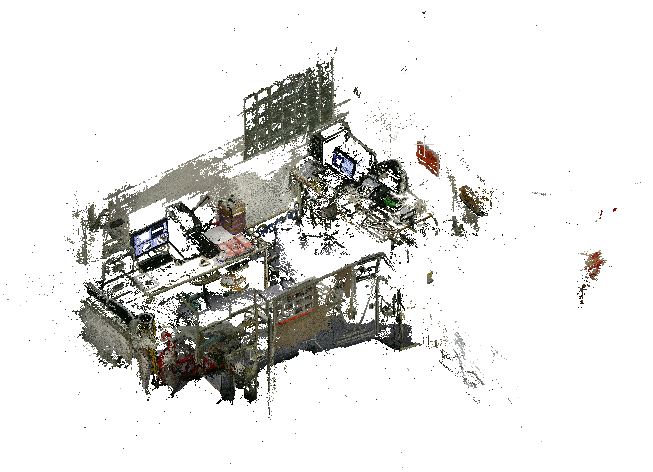} &
        \includegraphics[width=4.5cm]{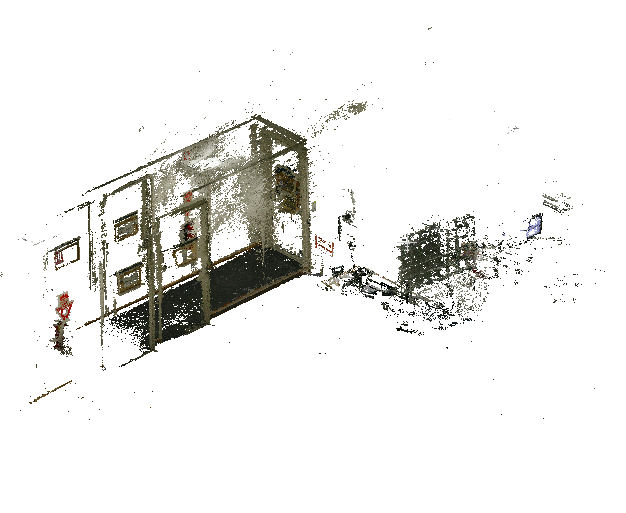} &
        \includegraphics[width=4.5cm]{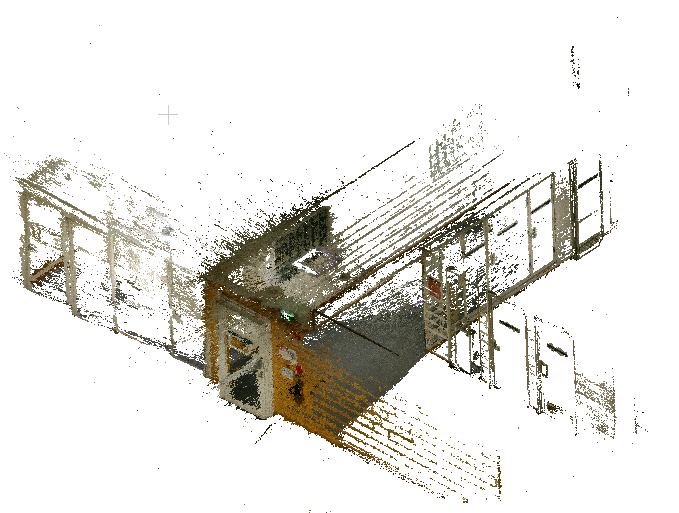} \\
        
        The lab room & The symmetrical hallways & The lobby area
        
    \end{tabularx}
    \caption{ Visualizations of the sparse reconstructions of the West Wing (when separated into three separate reconstructions) from different viewpoints. Note the overall sparsity and incompleteness --- especially in the hallway area. }
    \label{tab:failedwestwing}
    
\end{table*}

\subsubsection{\textbf{Experiment 0.3}: Photogrammetry (at the room scale)}
\label{ch:methods:s:exp:init:roomscale}

The scale considered here is on the order of a typical office room (see Table~\ref{tab:datasetsizes} for exact dimensions), so the UWA CSSE 3D printing lab is chosen to reconstruct the scene. The motivation behind this location choice and reduced scale is to avoid the issues encountered in the \textbf{Experiment 0.2} regarding visual symmetry \& scale, and to allow for local testing of the robotic navigation system in later experiments. Also, this constrained size is similar to the extents of the scenes in the \textit{7Scenes} and \textit{University} datasets (which were the datasets that the proposed network ultimately performed most accurately in).

Images in this particular section were not captured in the same manner as the final set of images used were (outlined in Section~\ref{ch:data:s:new:ss:coll}). Instead, $3042$ images were captured individually, at different positions and orientations within the scene, such that the scene was adequately covered by the image set. This reconstruction failed even though the scene size was considerably reduced. 

It is likely that this failure was due to the lack of matched image pairs over the image set: although this was greater than in the hallway scene (see Figure~\ref{tab:hallsym}), it was still not enough to reconstruct the scene via SfM. This initial experiment thus provided \textbf{two key avenues for improvements which were pursued during the more extensive experiments}:

\begin{enumerate}
    \item The number of matching image pairs needs to be increased. This can be achieved by hanging posters and framed images on walls, placing colourful and textured items around the scene, and removing large texture-less spaces (such as blank whiteboards). Surfaces which reflect specularly can also be covered. As demonstrated in Table~\ref{tab:sfmresults}, it is the number of matched pairs which promotes reconstruction, not the number of features per image.
    
    \item The data collection method needs to be sped up. Individually capturing $3042$ images took $\sim 1.5$ hours, so video sequences should be used to decrease the amount of time a user would have to spend capturing images (thus improving the full pipeline's ease-of-use).
\end{enumerate}

\subsubsection{\textbf{Experiment 0.4}: Adaptive Monte Carlo Localization as a baseline}

\begin{figure}
    \centering
    \includegraphics[width=0.95\linewidth]{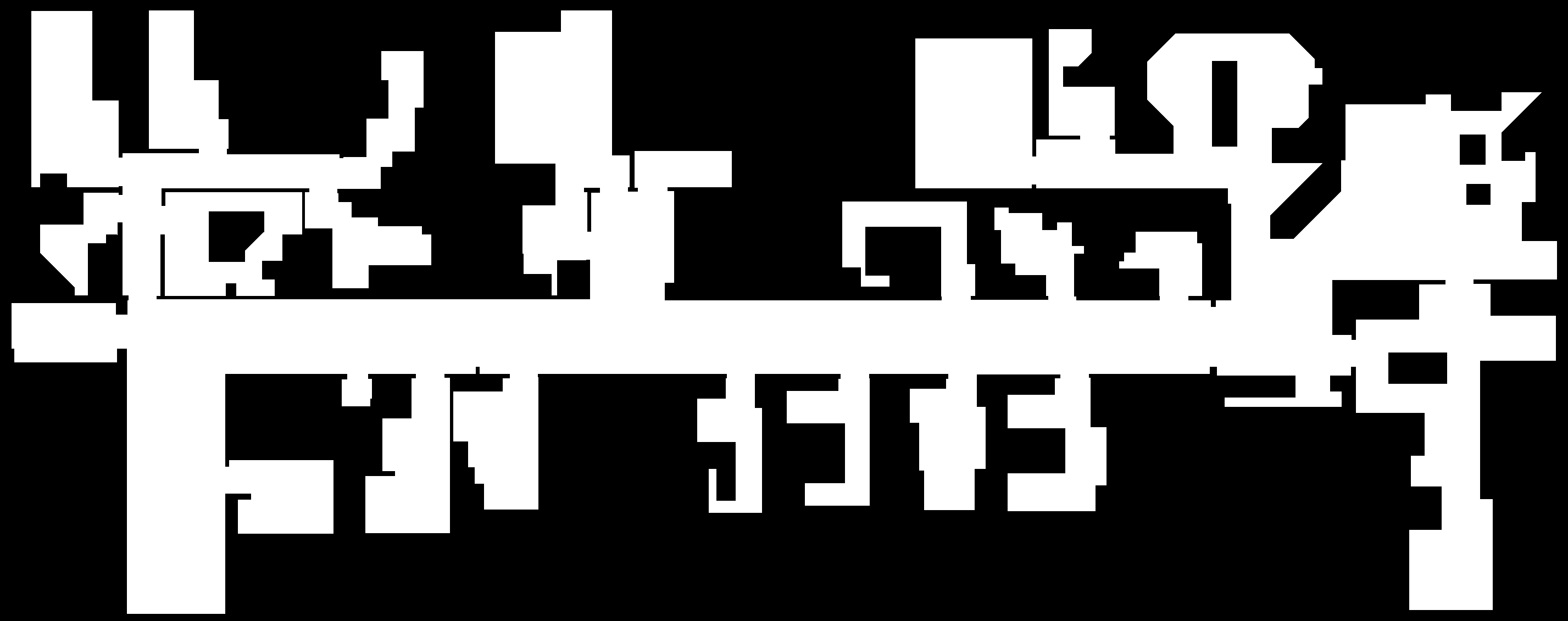}
    
    \caption{A 2D occupancy map of the CSSE building at UWA from the `top down' perspective (often referred to as a `floor map'). The value of each pixel indicates if it is filled by an obstacle (with black representing an occupied pixel and white representing an unoccupied pixel). The resolution of this map is $62.197$ pixels per metre.}
    \label{fig:floormap}
\end{figure}

Adaptive Monte Carlo Localization (AMCL) extends Monte Carlo Localization (MCL) (see Section~\ref{ch:litrev:s:nav} for details) by maintaining the random distribution of particles throughout state space (See Section~\ref{ch:litrev:s:nav} for more detail). The Robot Operating System (ROS) software library set allows for an AMCL module to be ran on the TurtleBot given that a 2D map is provided. This map is generated from a top-down view of the LiDAR scan from \textbf{Experiment 1.1}, which can be seen in Figure~\ref{fig:floormap}.

The AMCL module utilises depth scans --- not RGB image data --- from the Kinect camera to filter the particles in state space, yet irrecoverably fails when attempting to localize the robot in the building's symmetrical hallways. This shows that not only the visual, but also the \textit{structural} symmetry of the hallways makes them a challenging environment for localization, since the AMCL algorithm only considers the structural information (using a depth sensor). 

Seeing as the structure of the hallways cannot be easily altered, this experiment also suggests that visual features (posters, paintings and so forth) need to be added to the walls in order to improve feature extraction (which benefits the SfM experiments and the pose regression networks). It also suggests that depth \& odometry based particle feature approaches are insufficient for navigating highly symmetrical environments --- visual, or other information is required. This reinforced the need for an architecture which accepts visual inputs --- neural networks. Additionally, removing the need for odometry and RGB-D sensors in place of only RGB sensors improves the entire pipeline's ease-of-use and accessibility (see Figure~\ref{fig:fullpipeline} for a flow diagram outlining this full pipeline). 

Ultimately the shortcomings observed during this experiment encouraged the design of the second and third key objectives of this document (outlined in Chapter~\ref{ch:obj}), that is; to use improved neural networks which take visual input, and to use said networks to drive a performant robotic navigation system. Related experiments being outlined and presented in Section~\ref{ch:methods:s:exp:ss:pose:sss:poseregbenchmark} and Section~\ref{ch:resdis:s:poseregbenchmark} respectively.

\subsection{Photogrammetry experiments}
\label{ch:methods:s:exp:ss:photo}

Photogrammetry was chosen to produce pose-labelled image data for its ease-of-use: there is no requirement for non-standard sensors (RGB-D cameras), so the method is accessible for those who may want to repeat this work --- the only requirement is a computer which can run COLMAP (see Appendices~\ref{ap:reqs}). The data collection process for the final dataset which was used in this work is outlined in Section~\ref{ch:data:s:new:ss:coll}, and was motivated by the results of the initial investigations (see Section~\ref{ch:methods:s:exp:init:roomscale}).

\subsubsection{\textbf{Experiment 1.1}: Photogrammetry (at the room scale) with increased visual richness}
\label{ch:methods:s:exp:ss:photo:sss:roomscalehighfeat}

The choice of the UWA CSSE 3D printing lab was motivated and justified in Section~\ref{ch:methods:s:exp:init:roomscale}, but put shortly: the reconstruction was unsuccessful due to visually sparse regions in the supplied images. As stated in Section~\ref{ch:litrev:s:photo}, SIFT features are calculated over input images and these features inform the exhaustive matching process during the SfM procedure --- if the feature count is too low, or image pairs are not easily matched (due to excessive similarity), the reconstruction will inevitably fail.

Hence, a variety of objects/markers/landmarks are placed in the 3D printing lab to ensure that a greater count of SIFT features will be calculated. Once these items were in place, data was collected using video sequences rather than individual images: this improves upon the pipeline's ease-of-use, as only $\sim 330$ seconds of video needed to be gathered as an input to the SfM process, rather than taking $\sim 3000$ individual photographs. The results of this reconstruction are detailed in Section~\ref{ch:resdis:s:photo}, and the details of this data collection process are provided in Section~\ref{ch:data:s:new:ss:coll}.

\subsection{Pose regression experiments}
\label{ch:methods:s:exp:ss:pose}

The existing research into pose regression networks typically involves designing new architectures for regressing pose from RGB images (or other inputs). Interestingly however, some of the more recent works have obtained competitive results by redesigning aspects of these pose regression networks \cite{kendall2015posenetuncertain, walch2016posenetlstm} --- in particular the loss function \cite{kendall2017posenetgeo}.

A network's loss function facilitates its training; as batches of training data are input to the deep neural network, its parameters are updated according to how successful its predictions are. Each change to these parameters essentially \textit{re-configures} the network to respond differently to input data. The loss function provides a numerical performance metric by considering the predicted values against the expected values --- provided by an SfM process in this case. The differentiable property of the loss function also facilitates the backpropogation process, which allows a neural network's to be trained in a manner which guarantees an improvement in performance \cite{khan2018cnn}. Once a large set of training data is fed to the network, it it presumed to reach a set of parameters which provide high performance for similar input data.

Naturally, experimentation with this loss function greatly alters the final parameter configuration of a neural network --- indeed, initial forays into this area demonstrate that misguided, non-intuitive loss function formulations drastically decrease performance.

\subsubsection{\textbf{Experiment 2.1}: Loss function formulation}
\label{ch:methods:s:exp:ss:pose:sss:lossfunctionform}

A common technique used when regressing a camera's pose from an image is to formulate the loss as a linear combination of positional and rotational error (using tuned hyperparameters as coefficients). In this work it is observed that changes to rotation and position mutually affect the captured image, and in order to improve performance, a network's loss function should include a term which combines error in both position and rotation. To that end a geometric loss term is designed which considers the similarity between the predicted and ground truth poses using both position and rotation, and it is used to augment the existing image localization network PoseNet \cite{kendall2015posenet}.

There are many intuitive, geometric mathematical formulations which combine these errors; and naturally each requires testing (listed with results in Section~\ref{ch:resdis:s:lossfunctionform}). To test these, I modify a TensorFlow implementation of PoseNet by changing its loss function to the formulation being tested, and then train the network on the \textit{7Scenes Chess} scene, after which point it is tested on the scene's test set. The spatial extents (see Table~\ref{tab:datasetsizes}), indoor context, and challenges (see Table~\ref{tab:hardframes}) put forth by the Chess scene deem it a highly general and thus suitable scene for benchmarking loss formulations. A Tesla K40c graphiCSSE card is leveraged when training to perform the required machine learning computations, allowing the network to be trained in $\sim 10$ hours. 

The median positional and rotational localization errors are catalogued to give an idea of which formulations will give the best localization accuracy in general, further informing the design of future formulations --- this methodology allows for the space of loss formulations to be adequately explored using quantitative comparisons. 

\subsubsection{\textbf{Experiment 2.2}: Hyper-parameter tuning}
\label{ch:methods:s:exp:ss:pose:sss:hyperparamtuning}

With the formulation decided on, a grid search for the hyper-parameters which minimize median localization errors is performed. These experiments are highly systematic: the pose regression network is automatically retrained and tested for a set of structured hyperparameter combinations over the \textit{7Scenes Chess} scene. The three hyperparameters are:

\begin{enumerate}
    \item The positional error's weighting.
    \item The rotational error's weighting.
    \item The proposed loss term's weighting(s).
\end{enumerate}

See Section~\ref{ch:resdis:s:hyperparamtune} for the results of these hyperparameter selections, and a discussion of the effect they have on the final trained model.

\subsubsection{\textbf{Experiment 2.3}: Pose regression bench-marking}
\label{ch:methods:s:exp:ss:pose:sss:poseregbenchmark}

At this point the proposed loss term's formulation and hyperparameters are now \textit{exactly} decided upon, and wide scale benchmarking against the existing datasets discussed in Section~\ref{ch:data:s:exist} is performed. These results provide the most definitive and extensive reports of the proposed loss term devised in this work: the performance (median localization error) is catalogued for $19$ scenes across $4$ datasets and compared directly to localization algorithms found in the literature, with a focus placed on its comparison against algorithms with similar processing steps.

The methodology here is simple: for each scene in each dataset, the required algorithms are trained and tested (per-scene) as required. The test results are compared in Section~\ref{ch:resdis:s:poseregbenchmark}.

\subsection{Robotic navigation experiments}
\label{ch:methods:s:exp:ss:rnav}

The pose regression experiments outlined above allow for the development and testing of some powerful neural networks, but without application their practicality is brought into question. The following robotic navigation experiments are designed to demonstrate that the generalization capability of a pose regression network can be inherited by the navigation pipeline, if it is built around the pose regression network.

\begin{figure}[h]
    \centering
    \includegraphics[width=0.6\linewidth]{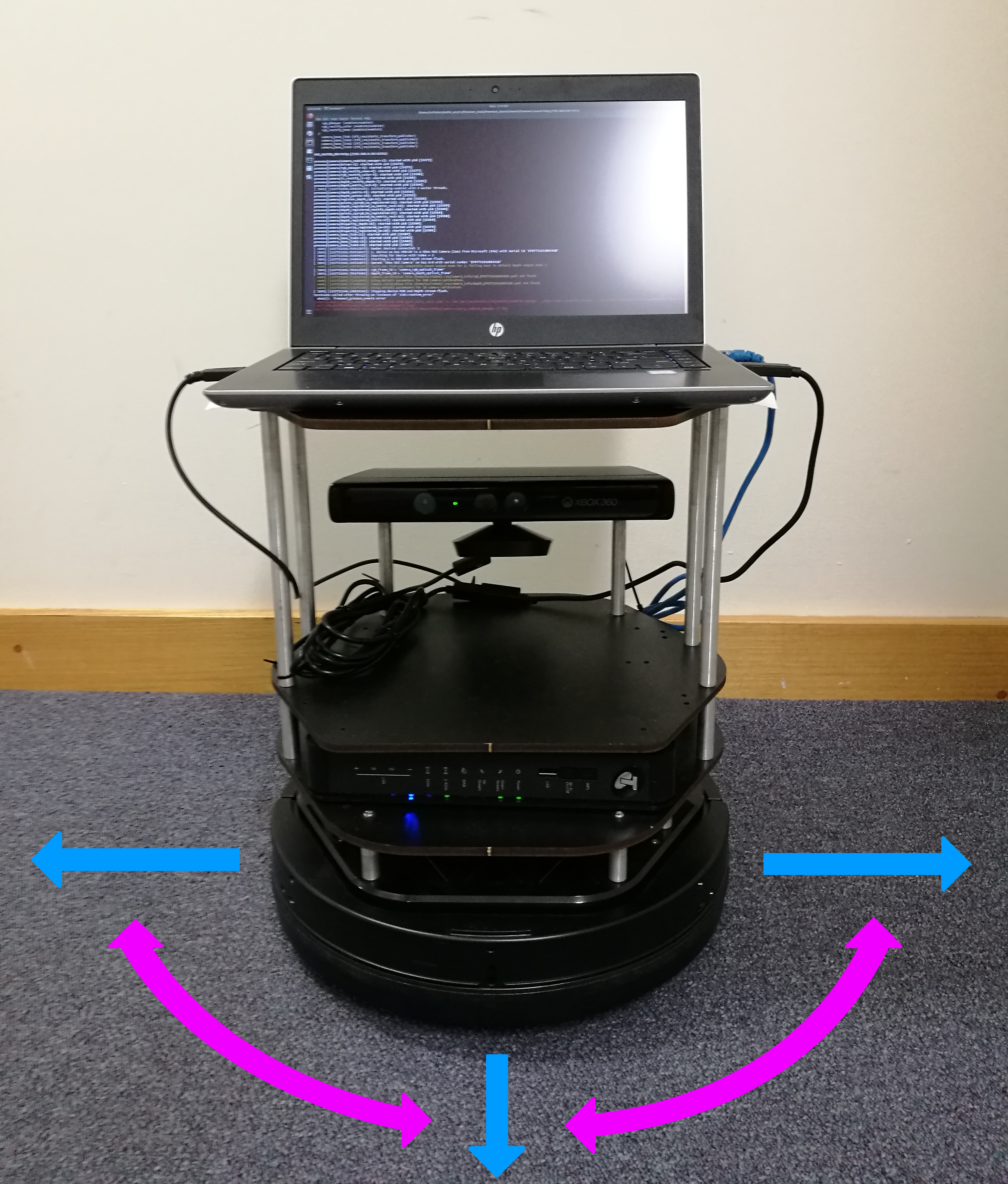}

    \caption{See that the TurtleBot's construction constrains it to have only 3 degrees of freedom. The device can only rotate around its yaw axis, and move in the $(x,z)$ plane. The camera is also always fixed in the direction of movement, as it cannot move separately with respect to the device. Note that the y-axis is defined to be pointing upwards, perpendicular to the ground plane in all the coordinate systems used in this work. }
    \label{fig:turtlebotdof}
\end{figure}

Note that the design of the TurtleBot only allows for 3 degrees of freedom: yaw axial rotation, and movement in the $(x,z)$ plane. The proposed network regresses 6 degrees of freedom: rotation as a quaternion, and position in 3D space. Only the three relevant components are considered in these tests (see Figure~\ref{fig:turtlebotdof} for a visual explanation).

\subsubsection{\textbf{Experiment 3.1}: Simple paths}
\label{ch:methods:s:exp:ss:rnav:sss:simplepaths}

The positional and rotational regression power of the network (trained on the UWA-lab scene) is tested here. The position and rotation metrics are considered separately in these tests. These pathing tests are created by tele-operating the TurtleBot, and moving it along a simple path whilst capturing video from the Kinect camera in real time (examples of these kinds of paths can be seen rendered in Figure~\ref{fig:navpaths}). The resulting set of images creates the test set. 

Note that the set of poses predicted by the regression network is difficult to compare \textit{directly} to the ground truth, since the ground truth can only be obtained by using rough measures of position and rotation when gathering the data (\ie measuring the TurtleBot's poses manually/by hand in the real world and considering odometry measurements, which can be innacurate as stated in Section~\ref{ch:litrev:s:nav:b:dead}). 

Therefore, since the poses of each image produced by the TurtleBot cannot be easily produced, and to ensure that this set of experiments is more than just a simple demonstration, the \textit{distribution} of poses is considered:

\begin{itemize}
    \item For the first type of simple path, the Turtlebot is moved in a \textbf{straight line} from one location in the room to another. In this case the regressed rotation is expected to be constant, as the Kinect \textit{always} points in the direction that the TurtleBot is facing, which again, is constant. The $(x,z)$ positions should move in a straight line, as per the TurtleBot. Hence, for a quantitative comparison, the standard deviation in the regressed yaw values should be low (since a perfect system would regress all values to be the same, as the rotation is constant), and the regressed $(x,z)$ positions should fit to a straight line, so the sum of squared residuals with respect to a linear model can be considered as a performance metric.
    
    \item For the second type of simple path, the TurtleBot's position remains constant as it completes a $360$\degree \textbf{rotation}, whilst capturing images at precisely $30$ frames-per-second (as reported by the ROS image capturing code). Conversely to the first type of path, here it is expected that the position remain constant and that the rotation's yaw component be uniformly distributed in the range [$0$, $360$]\degree. To produce quantitative results, the standard deviation in the distribution of the regressed position values should be low, and it needs to be tested if equal numbers of the regressed yaw samples fall into various, equally sized bins in the range [$0$, $360$].
\end{itemize}

In both cases, the camera captures many image samples as the device moves along a given path, and thus the coverage of the scene is high --- the starting and ending positions of the straight line paths are chosen such that at least $70$\% of the spatial extent of the scene in any given direction is traversed, and for in place rotations, the scene is sampled in every direction.

\subsubsection{\textbf{Experiment 3.2}: Compound paths}
\label{ch:methods:s:exp:ss:rnav:sss:advancedpaths}

\begin{figure}
    \centering
    \includegraphics[width=0.95\linewidth]{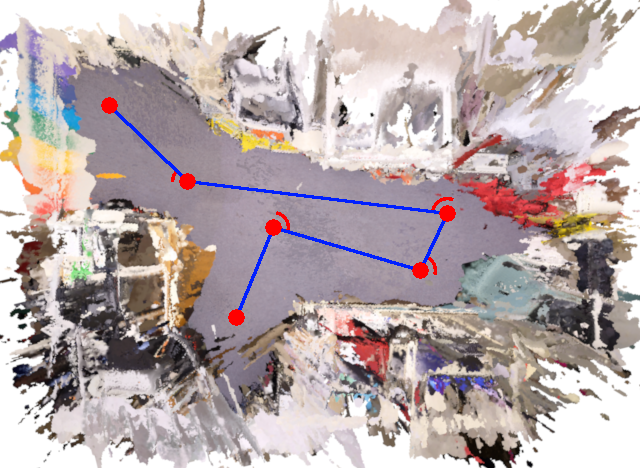}

    \caption{A compound TurtleBot path with multiple turns (red) and straight line movements (blue). See that the path can be described as a sequence of `simple paths', which are outlined in Section~\ref{ch:methods:s:exp:ss:rnav:sss:simplepaths}. This type of movement is identical to the type of movement found in Roombas and other simple UGVs. }
    \label{fig:navpaths}
\end{figure}

The goal here is to test the positional and rotational accuracy in tandem. The paths here are thus simply an extension of the paths outlined above, but the position \textit{and} rotation of the TurtleBot (and thus the Kinect camera) are \textit{both} changed over the course of the path. See Figure~\ref{fig:navpaths} for a visualization. It is not possible to compare the regressed pose distribution to the actual distribution as in Section~\ref{ch:methods:s:exp:ss:rnav:sss:simplepaths} above, since the movement here is less constrained. However, the experiment still demonstrates the regression power of the network, and allows comments to be made regarding its robustness, failure cases and efficiency. For a discussion of the results procured during this experiment, see Section~\ref{ch:resdis:s:advancedpaths}

\chapter{Results and discussion}
\label{ch:resdis}

In this section, the results of the experiments outlined in Section~\ref{ch:methods:s:exp:ss:pose}, Section~\ref{ch:methods:s:exp:ss:photo} and Section~\ref{ch:methods:s:exp:ss:rnav} are discussed with respect to a set of relevant performance metrics outlined in Section~\ref{ch:methods:s:perf}\footnote{The results of any brief, initial investigations were discussed in Section~\ref{ch:methods:s:exp:ss:init} so as to reserve this section for only the most important results}.

\section{Experiment 1.1: Photogrammetry (at the room scale) with increased visual richness}
\label{ch:resdis:s:photo}

Quantitative results regarding this photogrammetry experiment as compared to previous photogrammetry experiments are provided here (see Table~\ref{tab:sfmresults}), as are some rendered samples of the final reconstruction (see Table~\ref{tab:denseresults}).

\begin{table}[h]
    \centering
    \begin{tabular}{ c | l | c | c | c | c | c }
        
        \hline\hline
        
             & Scene       &  Reg.   & Avg. & Matched  & Avg. feat. & Recon.   \\
        ID   & description &  img.   & SIFT & pairs    & matches    & points \\
        
        \hline\hline
        
        0 & WW, full              &  $998$ & $3149$ & $1.81\times10^{4}$ & \bm{$63.06$} & $1.83\times10^{6}$ \\
        
        \hline 
        
        1a & WW, lobby             & $202$ & $1925$ & $2.03\times10^{4}$ & $18.91$ & $1.64\times10^{6}$ \\
        1b & WW, hallway           & $201$ & $1459$ & $2.01\times10^{4}$ & $12.60$ & $2.86\times10^{6}$ \\
        1c & WW, lab               & $283$ & $4019$ & $3.99\times10^{4}$ & $16.82$ & $5.18\times10^{6}$ \\
        
        \hline
        
        2 & UWA-lab (plain) &  $3042$ & \bm{$4524$} & $5.01\times10^{6}$ & $11.27$ & $3.31\times10^{7}$ \\
        
        \hline
        
        3 & UWA-lab (rich) &  \bm{$3228$} & $1339$ & \bm{$5.21\times10^{6}$} & $8.85$ & \bm{$4.51\times10^{7}$} \\
        
        
        
        \hline\hline
        
    \end{tabular}

    \caption{ A collection of statistics from the individual photogrammetry experiments conducted in and around the West Wing of UWA's CSSE building (including the final and most successful experiment). Note that the average number of SIFT features is reported \textit{per-image}, and the average number of feature matches is reported \textit{per-image pair}, with the total number of matched image pairs also provided. Results are extracted directly from SQL databases produced by COLMAP. Abbreviations: img., images; reg., registered; avg., average; feat., feature; recon., reconstructed; WW, West Wing. }
    \label{tab:sfmresults}
\end{table}

As per the initial investigations, reducing the scale of the surveyed scene from the entire West Wing, to smaller, more manageable sub-scenes, generally increased the success of the reconstruction (determined by the number of reconstructed 3D points, and a visual inspection).

See that the UWA-lab (ID $3$) photogrammetry experiment (which is used for all the pose regression and navigation experiments in this work) has the greatest number of matched image pairs, and the greatest number of produced 3D points (despite being similar to the experiment with ID $2$). The average number of SIFT features calculated per image is much lower in the experiment with ID $3$ than in the experiment with ID $2$, due to lower resolution, blurrier video sequences being used in the former, compared to individual $4K$ images being used in the latter. However, the results illustrate that it is the number of matched image pairs that is more important during reconstruction. More broadly, this confirms that the landmark placement visually enriched the scene, thus allowing for a more successful reconstruction.

\begin{table}[h]
    \centering
    \begin{tabular}{ l | c | c }
    
        \hline\hline
    
        Description of scene & Pose regression network & Median errors \\
    
        \hline\hline
        
        UWA-lab (plain) & PoseNet (default)        & $1.27$m, $7.87$\degree \\
        UWA-lab (plain) & Proposed                 & $1.14$m, $7.90$\degree \\
        
        \hline
        
        UWA-lab (rich) & PoseNet (default)   & $0.15$m, $1.17$\degree \\
        UWA-lab (rich) & Proposed            & $0.11$m, $0.89$\degree \\
    
        \hline\hline
        
    \end{tabular}

    \caption{ The localization accuracy results of the proposed and default PoseNet pose regression networks when trained on the UWA-lab scene with and without visually rich landmarks being placed before data collection. See that a visually rich scene not only improves photogrammetric reconstruction results (see Table~\ref{tab:sfmresults}) but also \textbf{considerably} increases localization accuracy (due more accurately pose-labelled images).  }
    \label{tab:featurerichcomp}
    
\end{table}

The proposed and default PoseNet pose regression networks were tested on the datasets produced by experiments with ID $2$ and $3$ to assess whether the visual richness would improve localization accuracy, and indeed it does (quite considerably --- see Table~\ref{tab:featurerichcomp}). It is hypothesized that this is due to the more accurate pose labels, and an image set which has its features more readily extracted from CNN architectures (as opposed to hand-crafted features).

\begin{table}[h]
    \centering
    \begin{tabular}{ *{3}{P{4.5cm}} }
    
        
        \includegraphics[width=4.5cm]{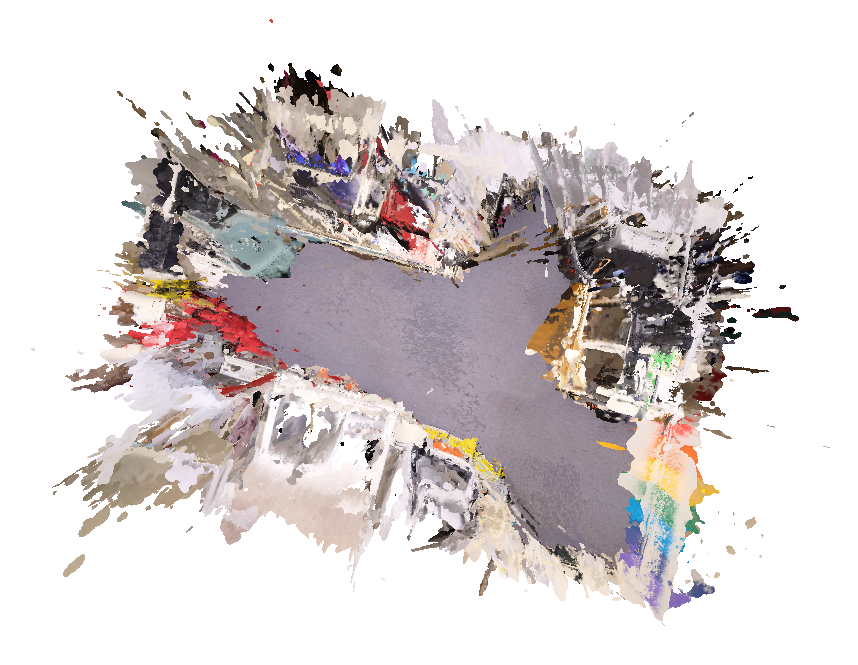} & 
        \includegraphics[width=4.5cm]{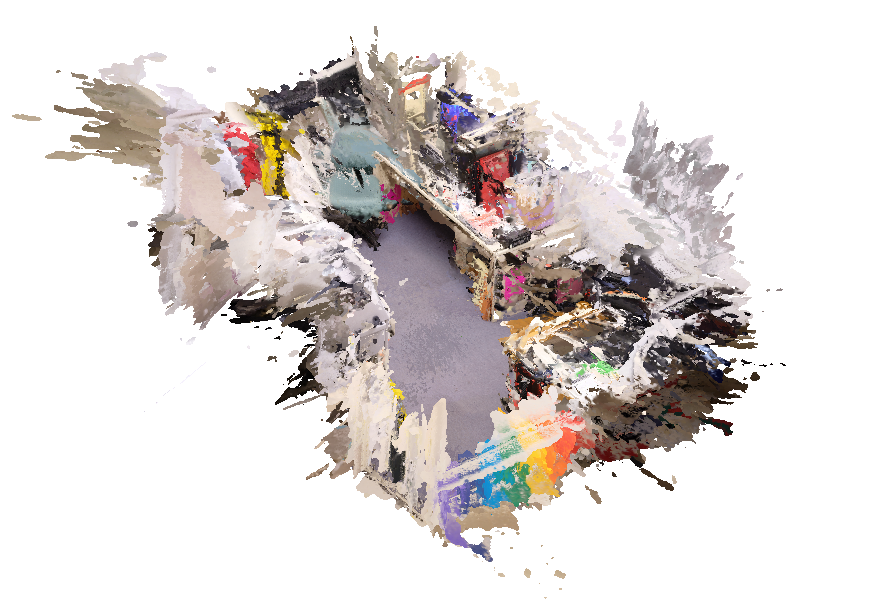} & 
        \includegraphics[width=4.5cm]{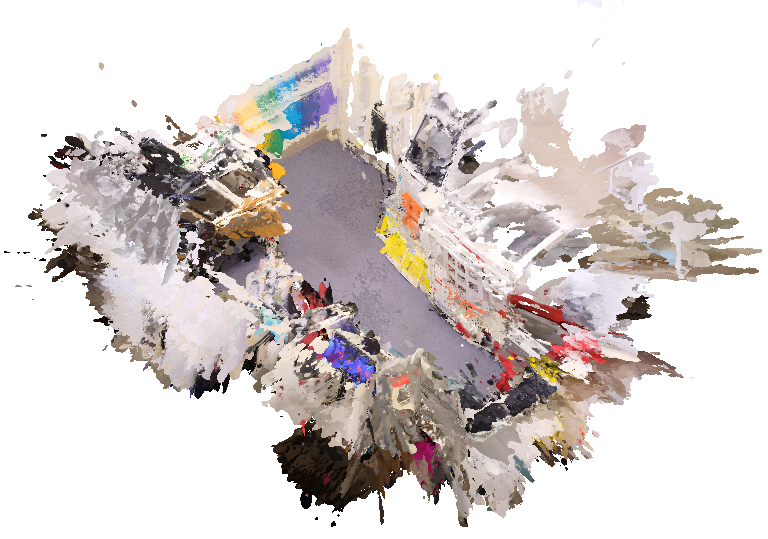} 
        
    \end{tabular}

    \caption{ Rendered samples of the final reconstruction's dense point cloud ($>4.50\times10^{7}$ points) from different angles. }
    \label{tab:denseresults}
    
\end{table}

\section{Experiment 2.1: Loss function formulation}
\label{ch:resdis:s:lossfunctionform}

The list of tested loss function formulations, and the median errors produced against the \textit{7Scenes Chess} scene when using each formulation (appended to the PoseNet architecture), are outlined in Table~\ref{tab:lossforms}.

\begin{table}
    \centering
    \begin{tabular}{ c | l | c }
        
        \hline\hline
        
        ID & Loss function formulation & Median errors \\
        
        \hline\hline
        
        0 & $\loss_{default} = \omega_{i} \cdot \| \hat{\vv{\textbf{p}}}_{i} - \vv{\textbf{p}} \| + \beta_{i} \cdot \| \hat{\vv{\textbf{q}}}_{i} - \vv{\textbf{q}} \| $ & $0.37$m, $7.45$\degree \\
        1 & $\loss = \loss_{default} + 512 \cdot (1 - \cos{\theta})$ & $0.82$m, $8.01$\degree \\
        2 & $\loss = \loss_{default} + \| \vv{\textbf{d}} \|^2 \cdot (1 - \cos{\theta})$ & $0.30$m, $8.49$\degree \\
        3 & $\loss = \loss_{default} + \| \vv{\textbf{d}} \| + \| \vv{\textbf{d}} \|^2 \cdot (1 - \cos{\theta}) $ & $0.30$m, $7.38$\degree \\
        4 & $\loss = \loss_{default} + \| \vv{\textbf{d}} \| + \| \vv{\textbf{d}} \|^2 \cdot (1 - \cos{\theta}) + 512 \cdot (1 - \cos{\theta}) $ & $1.17$m, $8.37$\degree \\
        5 & $\loss = \beta_{i} \cdot \| \hat{\vv{\textbf{q}}}_{i} - \vv{\textbf{q}} \| + \| \vv{\textbf{d}} \|^2 \cdot (1 - \cos{\theta}) + \| \vv{\textbf{d}} \|$ & \bm{$0.28$}m, $7.50$\degree \\
        \textbf{6} & $\loss = 1.5 \cdot \beta_{i} \cdot \| \hat{\vv{\textbf{q}}}_{i} - \vv{\textbf{q}} \| + \| \vv{\textbf{d}} \|^2 \cdot (1 - \cos{\theta}) + \| \vv{\textbf{d}} \|$ & \bm{$0.30$}m, \bm{$7.04$}\degree \\
        6$*$ & $\loss = 1.5 \cdot \beta_{i} \cdot \| \hat{\vv{\textbf{q}}}_{i} - \vv{\textbf{q}} \| + \| \vv{\textbf{d}} \|^2 \cdot (1 - \cos{\theta}) + \| \vv{\textbf{d}} \|$ & $0.33$m, $7.25$\degree \\
        7 & $\loss = 1.2 \cdot \beta_{i} \cdot \| \hat{\vv{\textbf{q}}}_{i} - \vv{\textbf{q}} \| + \| \vv{\textbf{d}} \|^2 \cdot (1 - \cos{\theta}) + \| \vv{\textbf{d}} \|$ & $0.29$m, $7.71$\degree \\
        8 & $\loss = 1.8 \cdot \beta_{i} \cdot \| \hat{\vv{\textbf{q}}}_{i} - \vv{\textbf{q}} \| + \| \vv{\textbf{d}} \|^2 \cdot (1 - \cos{\theta}) + \| \vv{\textbf{d}} \|$ & $0.29$m, $7.70$\degree \\
        9 & $\loss = \| \vv{\textbf{d}} \|^2 \cdot (1 - \cos{\alpha}) + \| \vv{\textbf{d}} \|^2 \cdot (1 - \cos{\theta}) + \| \vv{\textbf{d}} \|$ & $0.37$m, $9.26$\degree \\
        
        \hline\hline
        
    \end{tabular}

    \caption[.]{ A number of loss function formulations which were tested, and the errors they produced when used to inform the PoseNet architecture on the \textit{7Scenes Chess} scene. Note the actual number of tests was $33$, but only the most relevant are displayed here. All were trained with a batch size of $75$, and $30000$ iterations (apart from ID $6*$, which was trained for $45000$ iterations to test overfitting). See Figure~\ref{fig:geolossdiag} for a visual explanation of some of the terms.}
    \label{tab:lossforms}
\end{table}

The following symbols are used as shorthand to reduce excessive repetition in Table~\ref{tab:lossforms}: 
\begin{itemize}
    \item $\hat{\vv{\textbf{p}}}_{i}$, the regressed position from the $i^{th}$ affine regressor ($i = 1, 2, 3$, with $i = 3$ typically being the most accurate). 
    \item $\vv{\textbf{p}}$, the ground truth position. 
    \item $\hat{\vv{\textbf{q}}}_{i}$, the regressed rotation (as a quaternion) from the $i^{th}$ affine regressor.
    \item $\vv{\textbf{q}}$, the ground truth rotation (as a quaternion). 
    \item $\omega_{i}$, the hyperparameter which weights the importance of the positional error from the $i^{th}$ affine regressor. In practice $\omega_{1}=\omega_{2}=0.3$ and $\omega_{3}=1$. 
    \item  $\beta_{i}$, the hyperparameter which weights the importance of the rotational error from the $i^{th}$ affine regressor. In practice $\beta_{1}=\beta_{2}=150$ and $\beta_{3}=500$. 
    \item $\vv{\textbf{d}}$, the displacement between $\vv{\textbf{p}}$ and $\hat{\vv{\textbf{p}}}_{3}$. 
    \item $\theta$, the angle between $\vv{\textbf{p}}$ and $\vv{\textbf{d}}$. 
    \item $\alpha$, the angle between $\vv{\textbf{p}}$ and $\hat{\vv{\textbf{p}}}_{3}$. 
\end{itemize}

Obviously in order to regress pose successfully, positional and rotational error terms must be considered in the loss function, if either error is not present, then the network will not reduce said error when its layer weights are updated. A number of entries in Table~\ref{tab:lossforms} demonstrate this (note that even unsuccessful loss functions are included for completeness). 

The default (ID $0$) loss function formulation simply uses a hyperparameter weighting of the positional and rotational errors. Clearly this \textit{does} represent both quantities of interest, but it is not the most performant loss function in Table~\ref{tab:lossforms}, as it considers the coupled quantities entirely separately.

Of more interest perhaps is the geometric loss term introduced in this work (ID $6$). It combines positional and rotational errors more intuitively, and in a manner which better considers the intricacies of the pose regression task. The motivation behind the first part of the geometric loss term's design is to ensure that the predicted pose is indeed always `looking' at the correct scene content. The transformation from world coordinates $[\textbf{W}]$ to image coordinates $[\textbf{I}]$ using a calibration matrix $[\textbf{K}]$, a translation represented by the 3-vector $\vv{\textbf{t}}$ and a rotation defined by a rotation matrix $[\textbf{R}]$ as per Equation~\eqref{eq:worldtoimage} is considered initially.

\begin{equation}
\label{eq:worldtoimage}
    [\textbf{I}] = [\textbf{K}][\textbf{R} \ \text{\textbrokenbar} \vv{\textbf{t}}][\textbf{W}]
\end{equation}

Camera coordinates [$\textbf{C}$] can instead be converted to, as in Equation~\eqref{eq:worldtocamera}, if the camera calibration matrix is unavailable.

\begin{equation}
\label{eq:worldtocamera}
    [\textbf{C}] = [\textbf{R} \ \text{\textbrokenbar} \vv{\textbf{t}}][\textbf{W}]
\end{equation}

Transforming from world to camera coordinates can thus be accomplished easily, and from there the poses at each camera position can be extracted. Herein lies the problem; in the camera's coordinate system, the centre of the camera lies at the origin. This extends to any camera coordinate system calculated in this way, meaning that the quantities of interest cannot be directly compared, as they do not share a common coordinate system. 

The world coordinate system is however fixed for all cameras, and multiple `cameras' (or a single camera in different poses) can be compared in this coordinates system once transformed (Equation~\eqref{eq:cameratoworld}).

\begin{equation}
\label{eq:cameratoworld}
    [\textbf{W}] = [\textbf{R} \ \text{\textbrokenbar} \vv{\textbf{t}}]^{-1}[\textbf{C}]
\end{equation}

More explicitly, since the camera is by convention always facing in the negative z-direction, a vector representing its direction in world space can be calculated as per Equation~\eqref{eq:ztopose}. This vector is calculated with respect to world coordinates and can thus be used for comparison across different camera positions.

\begin{equation}
\label{eq:ztopose}
    p = [\textbf{R} \ \text{\textbrokenbar} \vv{\textbf{t}}]^{-1} \ [0 \ \ 0 \ \ {-1}]^{T}
\end{equation}

This method (Equation~\eqref{eq:ztopose}) can be used to calculate a pose vector given values of $[\textbf{R}]$ and $\vv{\textbf{t}}$. Hence, pose vectors for the ground truth and for predictions can be calculated provided that the translation and rotation of the desired pose is known. Maximizing the similarity between such vectors ensures that one is facing in the correct direction, and that the prediction is aligned with the ground truth.

\begin{figure}
    \centering
    \includegraphics[width=0.95\linewidth]{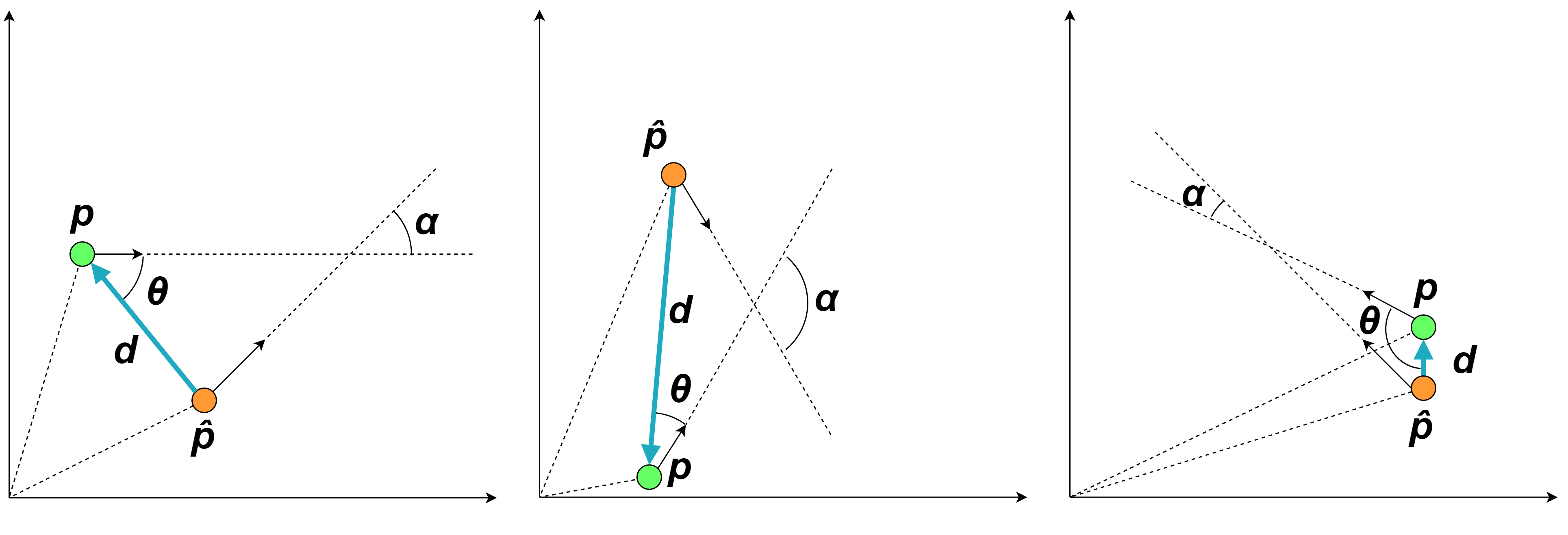}

    \caption[.]{ A visual explanation of the terms considered in the construction of the geometric loss term. The displacement vector $\vv{\textbf{d}}$ is the displacement between the position of the ground truth pose $\vv{\textbf{p}}$ and the position of the predicted pose $\hat{\vv{\textbf{p}}}$. The magnitude of $\vv{\textbf{d}}$ is analogous to the positional error. The angle $\theta$ quantifies how in-line (colinear) $\hat{\vv{\textbf{p}}}$ and $\vv{\textbf{p}}$ are --- when $\theta$ is small, the position $\hat{\vv{\textbf{p}}}$ must lie in the direction $\vv{\textbf{p}}$ is facing, and vice versa. The angle $\alpha$ is analogous to the rotational error --- when small, the predicted and ground truth poses must be facing in the correct direction. Three examples are provided. }
    \label{fig:geolossdiag}
\end{figure}

A loss term can then be formulated which combines both translational and rotational error by considering the difference between the predicted and ground truth position vectors, which is termed $\vv{\textbf{d}}$, see Figure~\ref{fig:geolossdiag}. The cosine similarity between $\vv{\textbf{d}}$ and the ground truth position vector --- rather than the difference between the ground truth position vector and the predicted position vector --- is used as a measure of similarity.

A loss term designed in this way ensures that the network, at a high level, considers both position and rotation when regressing pose from an image. 

\section{Experiment 2.2: Hyper-parameter tuning}
\label{ch:resdis:s:hyperparamtune}

Additional experiments were conducted concerning the effect of hyperparameter choices on the performance of the network. A grid search was used to narrow down the hyperparameter choices for the formulation in Equation~\eqref{eq:lossterm} (see the loss function with ID 6 in Table~\ref{tab:lossforms}). 

\begin{equation}
\label{eq:lossterm}
    \loss = 1.5 \cdot \beta_{i} \cdot \| \hat{\vv{\textbf{q}}}_{i} - \vv{\textbf{q}} \| + \| \vv{\textbf{d}} \|^2 \cdot (1 - \cos{\theta}) + \| \vv{\textbf{d}} \|
\end{equation}

Of particular interest however is the effect of finely changing the hyperparameter coefficient for the rotational error. Alterations from $1.5$ to $1.2$ and from $1.5$ to $1.8$ decreased median rotational accuracy markedly in each case (decrease of $2.3$\degree). In fact, these changes caused median \textit{positional} accuracy to be increased by $21\%$ and $23\%$ respectively. This suggests that the choice of $1.5$ desirably maximizes the median rotational accuracy. 

In general, other hyperparameter tuning operations greatly reduced localization accuracy, so the hyperparameters which weight $\| \vv{\textbf{d}} \|^2 \cdot (1 - \cos{\theta})$ and $\| \vv{\textbf{d}} \|$ are kept at $1.0$.

In short, the final loss term used to train the proposed model (Equation~\eqref{eq:lossterm}) is the result of an exploration in the space of possible geometric loss terms, and the decisions behind the term's design were experimentally informed.

\section{Experiment 2.3: Pose regression bench-marking}
\label{ch:resdis:s:poseregbenchmark}

The proposed network's performance with respect to the criteria outlined in Section~\ref{ch:methods:s:perf} is detailed here.

\textbf{Performance criteria --- Accuracy}

The median positional and rotational error is reported in Table~\ref{tab:benchresults1} for a variety of pipelines of varying classes: traditional SIFT based methods, augmented architecture networks, and augmented loss function networks (see Section~\ref{ch:litrev:s:imloc} for more information regarding these pipelines). 

In order to demonstrate the consistency and generalization of the proposed network, all scenes in all datasets are trained using the same setup. Each model is trained per-scene over $300,000$ iterations with a batch size of $75$ on a Tesla K40c, which takes $\sim 10$ hours to complete.

\begin{sidewaystable}
    \begin{tabularx}{\linewidth}    
        { l | c | c c | c c c }
        
        \hline\hline
              &  Active                           & Geometric                               & LSTM                            & Bayesian                                & Default                               & Proposed           \\ 
        Scene &  Search \cite{sattler2017efficient} & PoseNet \cite{kendall2017posenetgeo}   & PoseNet \cite{walch2016posenetlstm} & PoseNet \cite{kendall2015posenetuncertain}   & PoseNet \cite{kendall2015posenet}   & model \\ 
        
        \hline\hline
        
        Chess               & $0.04$m, $1.96$\degree & $\bm{0.13}$m, $\bm{4.48}$\degree & $0.24$m, $5.77$\degree & $0.37$m, $7.24$\degree & $0.32$m, $8.12$\degree & $\bm{0.31}$m, $\bm{7.04}$\degree \\ 
        Fire                & $0.03$m, $1.53$\degree & $\bm{0.27}$m, $\bm{11.3}$\degree & $0.34$m, $11.9$\degree & $\bm{0.43}$m, $13.7$\degree & $0.47$m, $14.4$\degree & $0.49$m, $\bm{13.3}$\degree \\ 
        Heads               & $0.02$m, $1.45$\degree & $\bm{0.17}$m, $\bm{13.0}$\degree & $0.21$m, $13.7$\degree & $0.31$m, $\bm{12.0}$\degree & $0.29$m, $\bm{12.0}$\degree & $\bm{0.24}$m, $15.7$\degree \\ 
        Office (7Scenes)    & $0.09$m, $3.61$\degree & $\bm{0.19}$m, $\bm{5.55}$\degree & $0.30$m, $8.08$\degree & $0.48$m, $8.04$\degree & $0.48$m, $\bm{7.68}$\degree & $\bm{0.40}$m, $10.0$\degree \\ 
        Pumpkin             & $0.08$m, $3.10$\degree & $\bm{0.26}$m, $\bm{4.75}$\degree & $0.33$m, $7.00$\degree & $0.61$m, $\bm{7.07}$\degree & $\bm{0.47}$m, $8.42$\degree & $0.49$m, $9.50$\degree \\  
        Red Kitchen         & $0.07$m, $3.37$\degree & $\bm{0.23}$m, $\bm{5.35}$\degree & $0.37$m, $8.83$\degree & $0.58$m, $\bm{7.54}$\degree & $0.58$m, $11.3$\degree & $\bm{0.53}$m, $7.98$\degree \\ 
        Stairs              & $0.03$m, $2.22$\degree & $\bm{0.35}$m, $\bm{12.4}$\degree & $0.40$m, $13.7$\degree & $\bm{0.48}$m, $\bm{13.1}$\degree & $0.56$m, $15.4$\degree & $\bm{0.48}$m, $14.7$\degree \\ 
        \hline
        Average             & $0.05$m, $2.46$\degree & $\bm{0.23}$m, $\bm{8.12}$\degree & $0.31$m, $9.85$\degree & $0.47$m, $\bm{9.81}$\degree & $0.45$m, $11.0$\degree & $\bm{0.42}$m, $11.2$\degree \\ 
        
        \hline\hline
        
        Great Court         & ---                    & $\bm{6.83}$m, $\bm{3.47}$\degree & ---                    & ---                    & ---                    & --- \\ 
        Street              & $0.85$m, $0.83$\degree & $\bm{20.3}$m, $\bm{25.5}$\degree & ---                    & ---                    & $\bm{3.67}$m, $\bm{6.50}$\degree & --- \\ 
        King's College      & $0.42$m, $0.55$\degree & $\bm{0.88}$m, $\bm{1.04}$\degree & $0.99$m, $3.65$\degree & $\bm{1.74}$m, $4.06$\degree & $1.92$m, $5.40$\degree & $2.28$m, $\bm{4.05}$\degree \\ 
        Old Hospital        & $0.44$m, $1.01$\degree & $3.20$m, $\bm{3.29}$\degree & $\bm{1.51}$m, $4.29$\degree & $2.57$m, $\bm{5.14}$\degree & $\bm{2.31}$m, $5.38$\degree & $3.90$m, $8.75$\degree \\ 
        Shop Facade         & $0.12$m, $0.40$\degree & $\bm{0.88}$m, $\bm{3.78}$\degree & $1.18$m, $7.44$\degree & $\bm{1.25}$m, $\bm{7.54}$\degree & $1.46$m, $8.08$\degree & $2.48$m, $10.2$\degree \\ 
        St Mary's Church    & $0.19$m, $0.54$\degree & $1.57$m, $\bm{3.32}$\degree & $\bm{1.52}$m, $6.68$\degree & $\bm{2.11}$m, $8.38$\degree & $2.65$m, $8.48$\degree & $3.02$m, $\bm{7.79}$\degree \\ 
        \hline
        Average$^{1}$       & $0.29$m, $0.63$\degree & $1.63$m, $\bm{2.86}$\degree & $\bm{1.30}$m, $5.52$\degree & $\bm{1.92}$m, $\bm{6.28}$\degree & $2.09$m, $6.84$\degree & $2.92$m, $7.70$\degree \\ 
        
        \hline\hline
        
    \end{tabularx}
    \\ 
    
    \caption{ The results of various camera localization methods for the \textit{7Scenes} and \textit{Cambridge Landmarks} datasets. Similar classes of systems are grouped (left to right: SIFT-based, augmented architecture networks and other networks). The lowest errors of each group are emboldened. Note that the proposed model is competitive with systems in the same class in \textit{indoor} datasets with respect to median positional error. 
    \newline
    $^{1}$ Average calculated using only the scenes: \textit{King's College}, \textit{Old Hospital}, \textit{Shop Facade} \& \textit{St Mary's Church} as full dataset performance is not available for all pipelines.
    \newline
    $^{2}$ For cells where results could not be procured (due to implementation availability or otherwise), a `---' is used. The inclusion of such results would not greatly effect any conclusion on the performance of the systems, particularly with respect to the \textit{Street} scene, as its large spatial extents only reduce the consistency of the Cambridge Landmarks dataset.
    }
    \label{tab:benchresults1}
\end{sidewaystable}

\begin{table}
    \centering
    \begin{tabular}{ l | c c  }
        
        \hline\hline
              & Default                               &  Proposed    \\ 
        Scene & PoseNet \cite{kendall2015posenet}   &  model \\ 
        
        \hline\hline
        
        Office (University) & $1.05$m, $16.2$\degree & $\bm{0.91}$m, $\bm{11.0}$\degree \\ 
        Meeting             & $1.78$m, $10.1$\degree & $\bm{1.30}$m, $\bm{9.58}$\degree \\ 
        Kitchen             & $\bm{1.19}$m, $\bm{12.5}$\degree  & $1.25$m, $15.5$\degree \\ 
        Conference          & $2.88$m, $\bm{13.3}$\degree       & $\bm{2.83}$m, $15.8$\degree \\ 
        Coffee Room         & $1.41$m, $14.9$\degree       & $\bm{1.21}$m, $\bm{13.3}$\degree \\ 
        \hline
        Average             & $1.66$m, $13.4$\degree & $\bm{1.50}$m, $\bm{13.0}$\degree \\ 
        
        \hline\hline
        
    \end{tabular}

    \caption{ A study on the direct effects of using the proposed model, instead of the default PoseNet. The lowest errors of each group are emboldened. Note that the contribution outlined in this work consistently outperforms the default PoseNet in both median positional and median rotational error throughout the \textit{University} dataset. 
    }
    \label{tab:benchresults2}
    
\end{table}

It is observed that the proposed model outperforms the default version of PoseNet in some scenes --- particularly the \textit{Stairs} scene. In the \textit{Stairs} scene, repetitious structures, \eg staircases, make localization harder, yet the proposed model is robust to such challenges. The network is outperformed in others scenes; namely outdoor datasets with large spatial extents, but in general, performance is improved for the indoor datasets \textit{7Scenes} and \textit{University} (see Table~\ref{tab:benchresults2}).

Note that the proposed model remains consistent and performant when compared to pose determination pipelines of a similar class (that is, some variation of PoseNet with a slightly augmented architecture), at least when considering median errors. Although the median errors are a useful metric for comparing performances across many scenes, a breakdown of the error distributions can be observed by viewing cumulative error histograms, which are provided in Table~\ref{tab:cumhists}.

\begin{table*}
    \vspace{-2.5cm}
    \centering
    
    \begin{tabularx}{\linewidth}    
        { *{2}{P{6.7cm}} }
        
        \includegraphics[width=6.7cm]{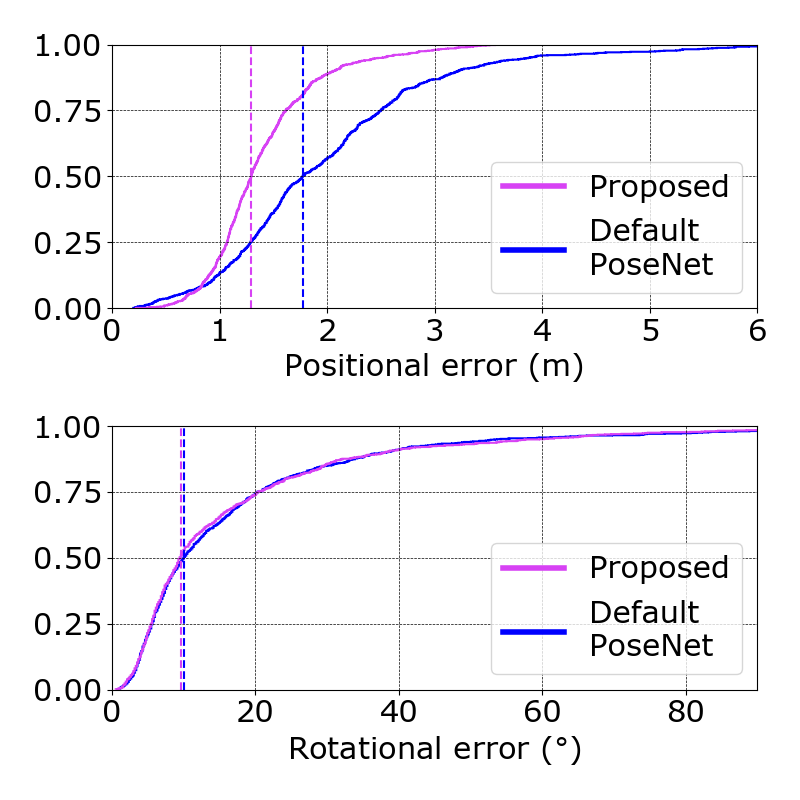} &
        \includegraphics[width=6.7cm]{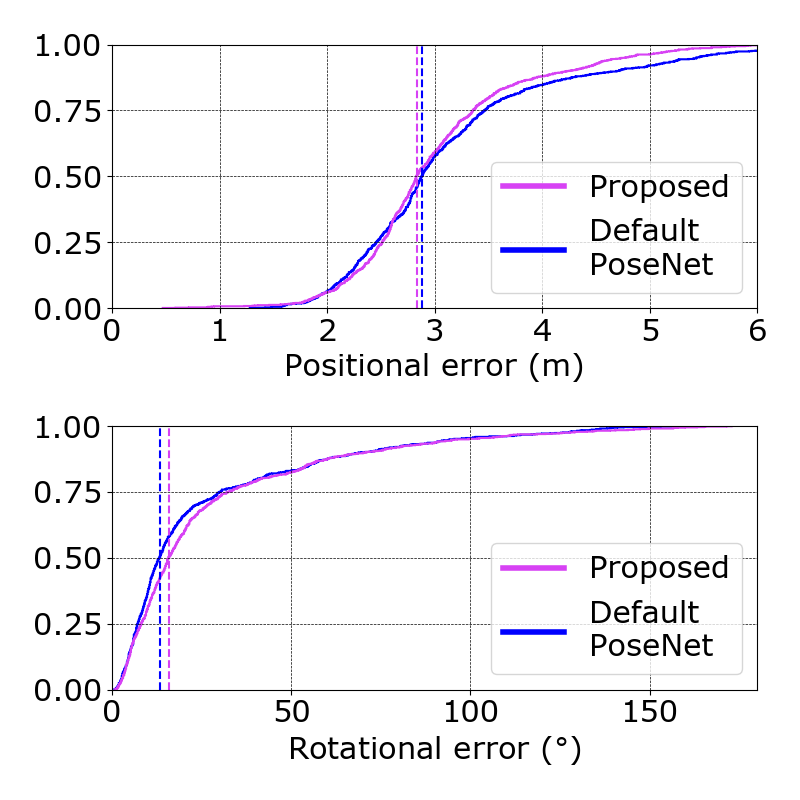} \\
        
        \textit{Meeting} & 
        \textit{Conference} \\
        
        \includegraphics[width=6.7cm]{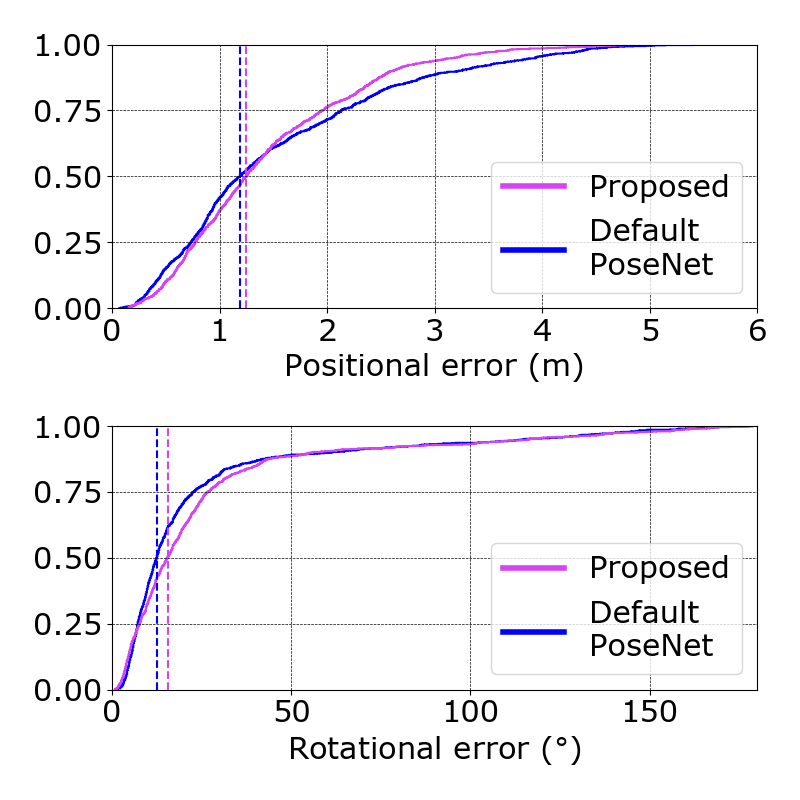} &
        \includegraphics[width=6.7cm]{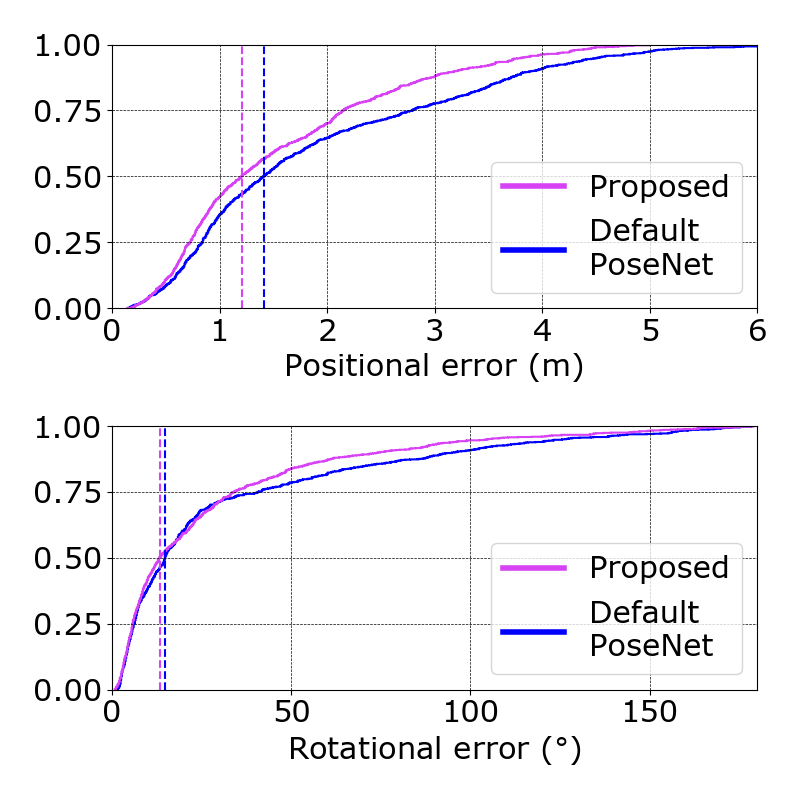} \\
    
        \textit{Kitchen} & 
        \textit{Coffee Room} \\
        
        \includegraphics[width=6.7cm]{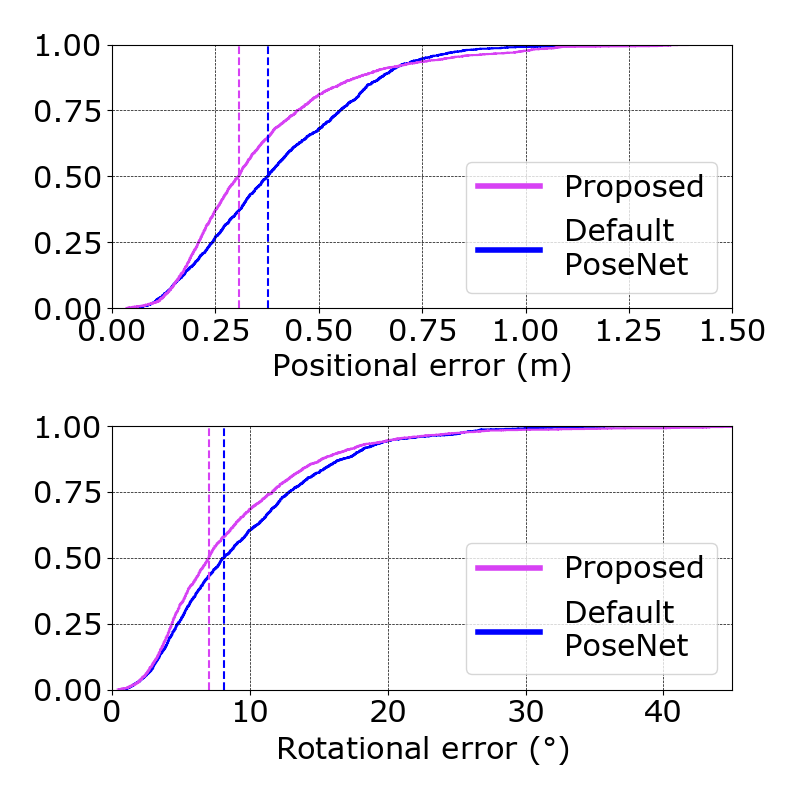} &
        \includegraphics[width=6.7cm]{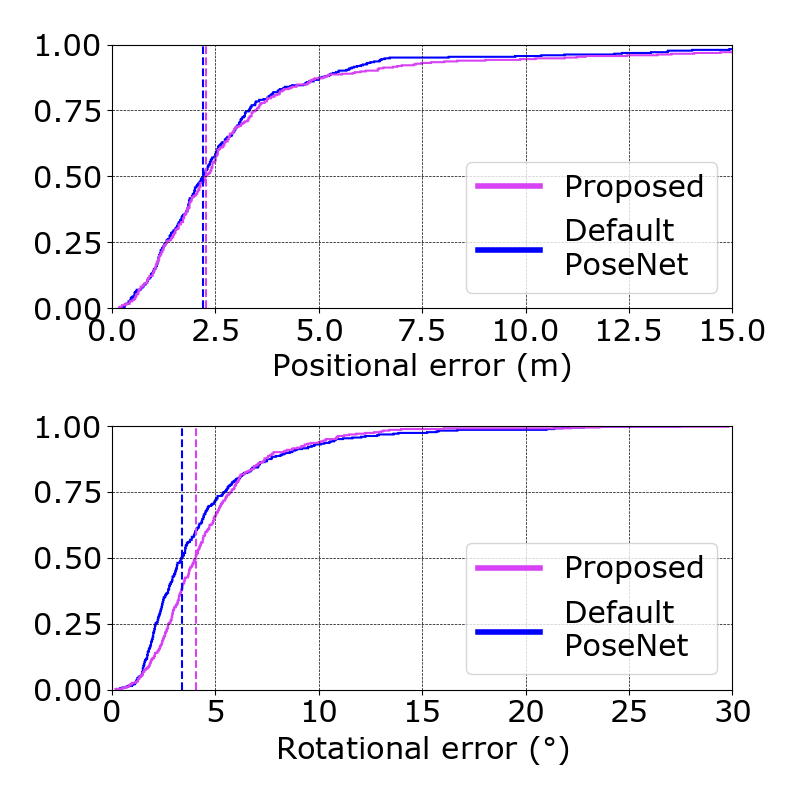} \\
        
        \textit{Chess} & 
        \textit{Kings College} \\
        
    \end{tabularx}
    \caption{Cumulative histograms of positional and rotational errors for a number of key scenes, with median values plotted as a dotted line. Images best viewed in colour.}
    \label{tab:cumhists}
\end{table*}

In Table~\ref{tab:cumhists}, the proposed model's positional error distribution is strictly less than (shifted to the left of) the default PoseNet's positional error distribution for the indoor scenes (except \textit{Conference}, where performance is comparable). Additionally, the maximum error of the proposed model is lower in the scenes \textit{Meeting, Coffee Room} and \textit{Kitchen}, meaning that the proposed model is robust to some of the most difficult frames offered by the \textit{University} dataset. The distribution of the positional errors and rotational errors are compared for the default PoseNet and for the proposed model. Median values (provided in Table~\ref{tab:benchresults1} and Table~\ref{tab:benchresults2}) are plotted for reference.

The proposed model's errors are strictly less than the default PoseNet's throughout the majority of the \textit{Chess} and \textit{Coffee Room} distributions. The default PoseNet outperforms our proposed model with respect to rotational accuracy in the [$10$, $30$]\degree range in the \textit{Coffee Room} scene.

\textbf{Performance criteria --- Robustness}

The robustness of the proposed model to challenging test frames --- that is, images with motion blur, repeated structures or demonstrating perceptual aliasing \cite{li2018fullframe} --- can be determined via the cumulative histograms in Table~\ref{tab:cumhists}. For the purpose of visualization some difficult testing images from the \textit{7Scenes} dataset are displayed in Table~\ref{tab:hardframes}. 

The hardest frames in the test set by definition produce the greatest errors. Consider the positional error for the \textit{Meeting} scene: the proposed model reaches a cumulative value of $1.0$ on the y-axis before the default PoseNet does, meaning that the \textit{hardest} frames in the test set have their position regressed more accurately (with less relative error). This analysis extends to each of the cumulative histograms in Table~\ref{tab:cumhists}, thus confirming the proposed loss function's robustness to difficult test scenarios, as the frames of greatest error consistently have less than or comparable errors to that of the default PoseNet. 

\begin{table}[h]
    \centering
    \begin{tabular}{ *{3}{P{4.5cm}} }
    
        
        \includegraphics[width=4.5cm]{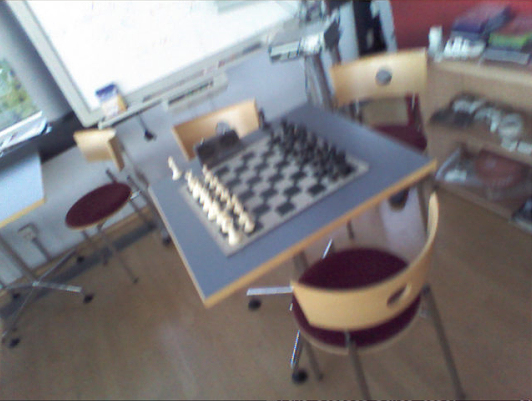} & 
        \includegraphics[width=4.5cm]{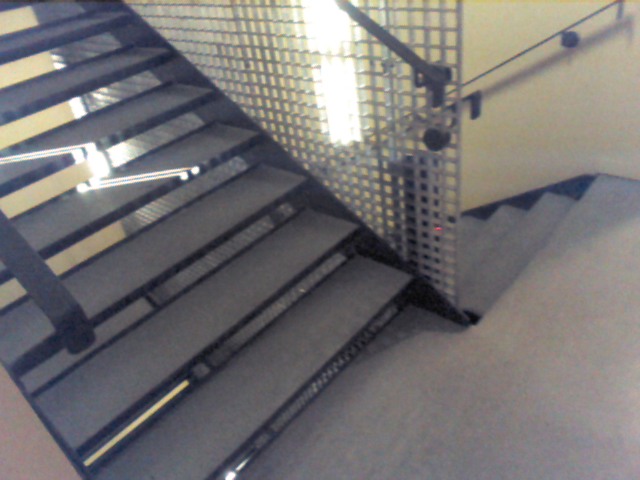} & 
        \includegraphics[width=4.5cm]{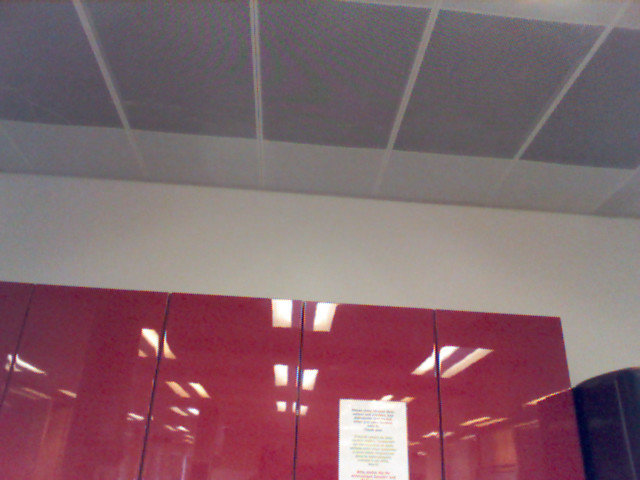} \\
        
        (a) Motion blur & (b) Repeated structures & (c) Textureless \& reflective surfaces\\
        
    \end{tabular}

    \caption{ (a) --- (c) Images from the \textit{7Scenes} dataset where accurately regressing pose is challenging. }
    \label{tab:hardframes}
    
\end{table}

\textbf{Performance criteria --- Resource intensiveness \& ease-of-training}

\textbf{Training time}. The duration of the training stage compared between the proposed model and default PoseNet is by design, very similar, and highly competitive when compared to the other systems analyzed in Table~\ref{tab:benchresults1}. This is due to the relatively inexpensive computing cost of introducing a \textit{simple} geometric loss term into the network's overall loss function. The average training time for default PoseNet and for the proposedmodel over the \textit{University} dataset is $10:21:31$ and $10:23:33$ respectively (HH:MM:SS), where both procedures are performed on the same hardware. 

\textbf{Testing time}. The network operation during the test time is obviously not affected by the loss function augmentation, as the loss function is only calculated during training time. The time performance when testing is similar to that of the default PoseNet and in general is competitive amongst camera relocalization pipelines. A total elapsed time of $16.04$ seconds is observed when evaluating the entire \textit{Coffee Room} scene testing set, whereas it takes $16.03$ seconds using the default PoseNet. In other words, both systems take $\sim16.8$ ms to complete a single inference.

\textbf{Memory cost}. The memory cost in general for CNNs is low --- only the weights for the trained layers and the input image need to be loaded into memory. When compared to feature matching techniques, which need to store feature vectors for all instances in the test set, or SIFT-based matching methods with large memory and computational overheads, CNN approaches in general appear desirable --- especially in resource constrained environments. Both the proposed model and the default PoseNet take $8015$MiB and $10947$MiB to train and test respectively (as reported by \textit{nvidia-smi}), and this is with maximum resolution images. For interest, the network weights for the proposed model's TensorFlow implementation total only $200$MB. 

\section{Experiment 3.1: Simple paths}
\label{ch:resdis:s:simplepaths}

As per Section~\ref{ch:methods:s:exp:ss:rnav:sss:simplepaths}, the TurtleBot is tele-operated to gather $4$ simple paths testing rotation regression (rotating in place) and $4$ simple paths testing position regression (moving in a straight line). Visualizations for each of these pathing experiments are provided in Table~\ref{tab:turtlebotspin} and Table~\ref{tab:turtlebotline} respectively, with numerical metrics made available in Table~\ref{tab:simplepathstatsspin} and Table~\ref{tab:simplepathstatsline}. 

\begin{table*}[h]
    \vspace{-3cm}
    \centering
    
    \begin{tabularx}{\linewidth}    
        { *{4}{P{6cm}} }
        
        \includegraphics[width=6cm]{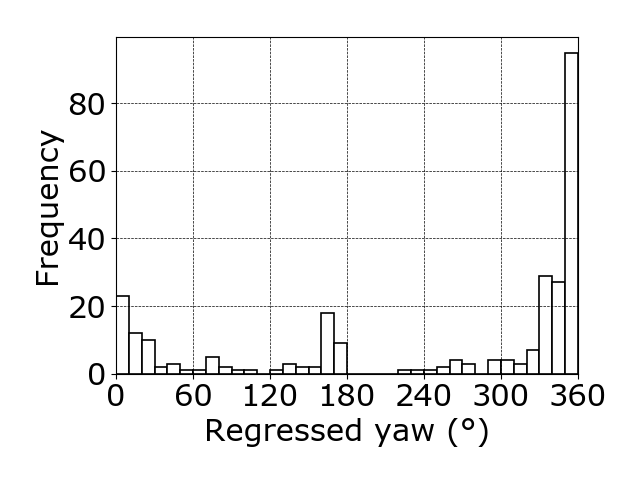} &
        \includegraphics[width=6cm]{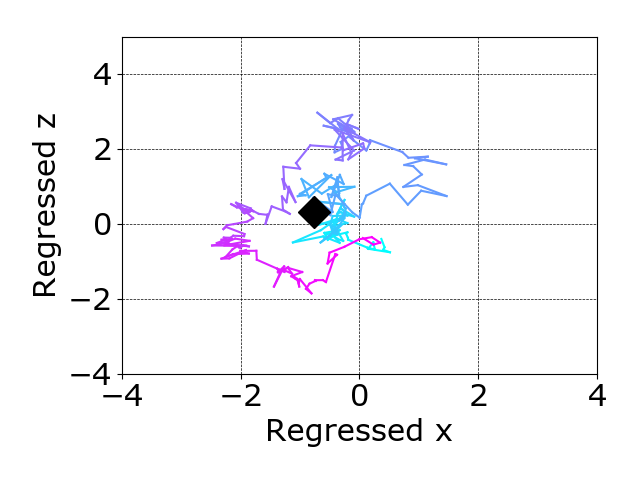} \\
        
        \multicolumn{2}{c}{Rotation (a)} \\
        
        \includegraphics[width=6cm]{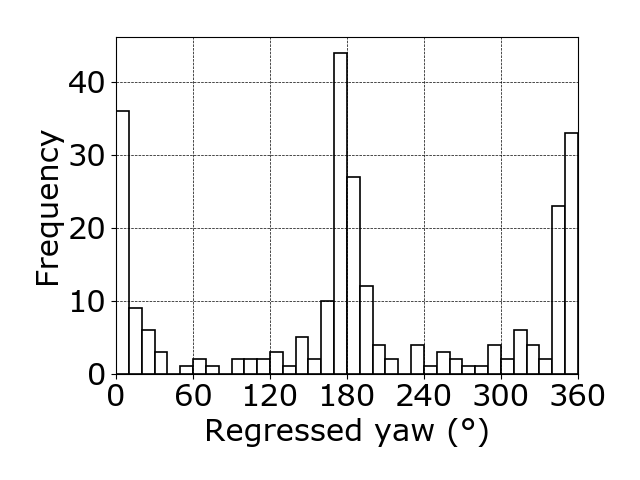} &
        \includegraphics[width=6cm]{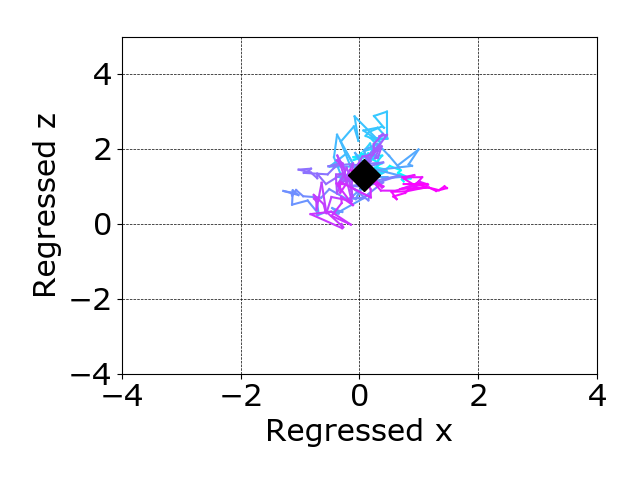} \\
        
        \multicolumn{2}{c}{Rotation (b)} \\
        
        \includegraphics[width=6cm]{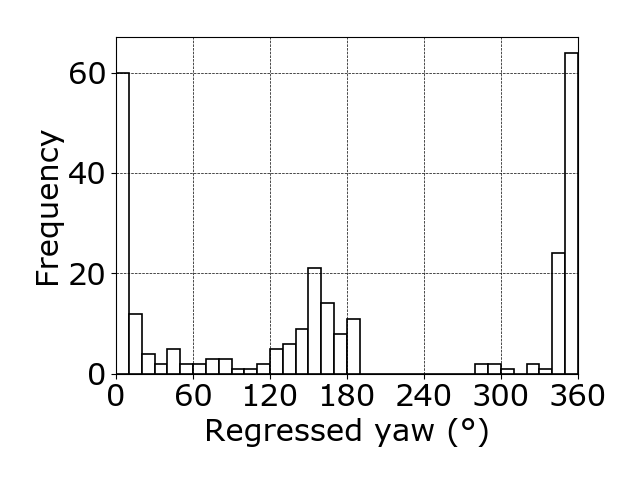} &
        \includegraphics[width=6cm]{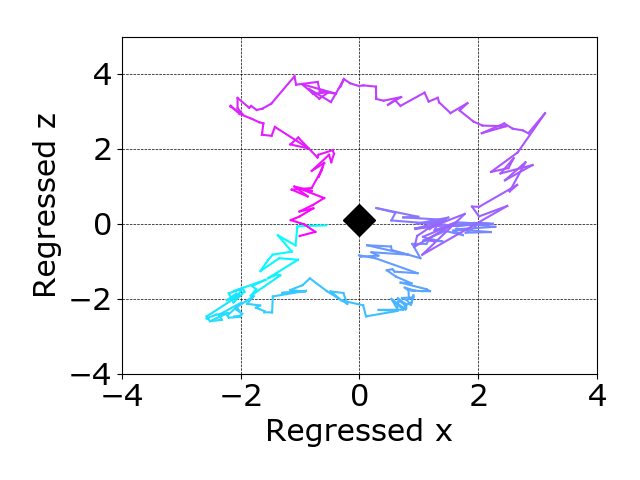} \\
        
        \multicolumn{2}{c}{Rotation (c)} \\
        
        \includegraphics[width=6cm]{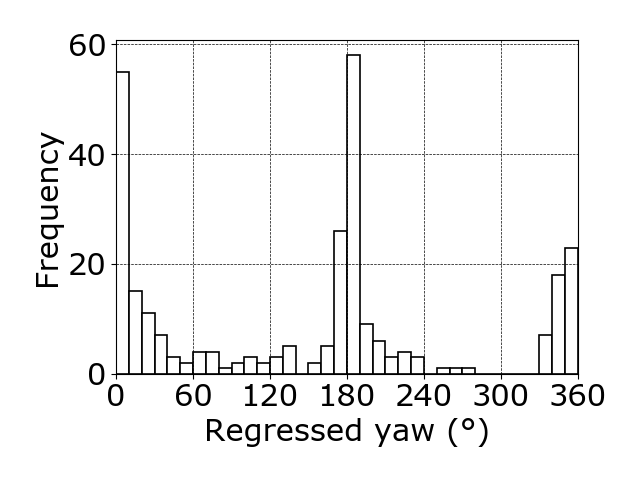} &
        \includegraphics[width=6cm]{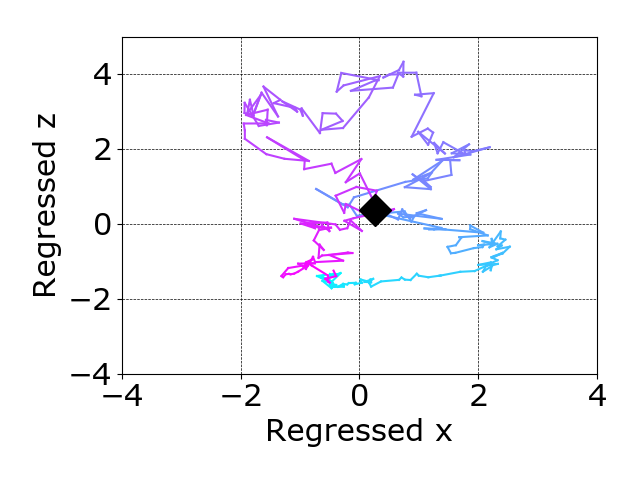} \\
        
        \multicolumn{2}{c}{Rotation (d)} \\
        
    \end{tabularx}
    \caption{ The regressed $(x,z)$ positions and regressed yaw yielded from the proposed network over the course of $4$ \textbf{in place rotations}. These paths are referred to as `rotations' (a) through to (d) from top to bottom, where each path's regressed yaw and position graphs are displayed on the left and right respectively. Mean positions are marked with a black diamond. }
    \label{tab:turtlebotspin}
\end{table*}

\begin{table*}[h]
    \vspace{-3cm}
    \centering
    
    \begin{tabularx}{\linewidth}    
        { *{4}{P{6cm}} }
        
        \includegraphics[width=6cm]{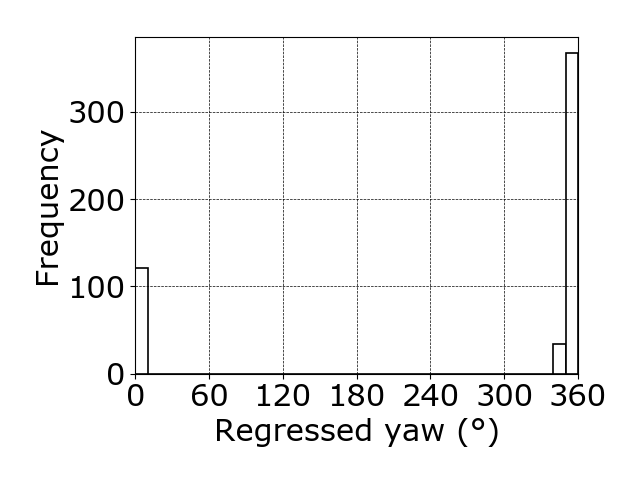} &
        \includegraphics[width=6cm]{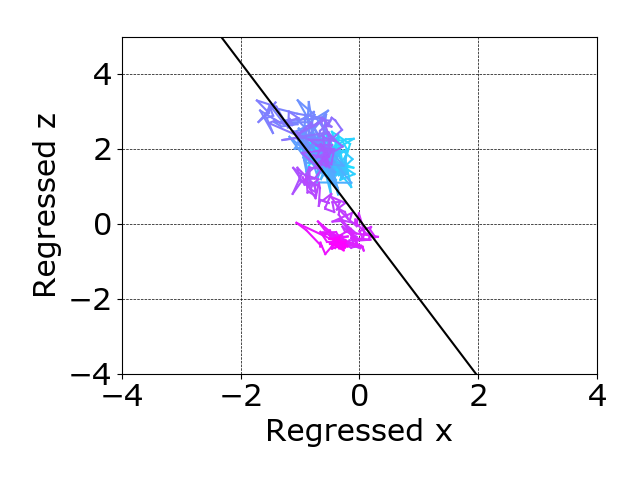} \\
        
        \multicolumn{2}{c}{Line (e)} \\
        
        \includegraphics[width=6cm]{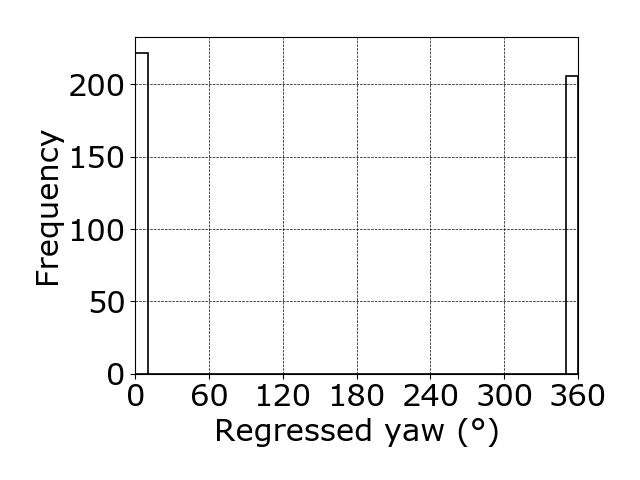} &
        \includegraphics[width=6cm]{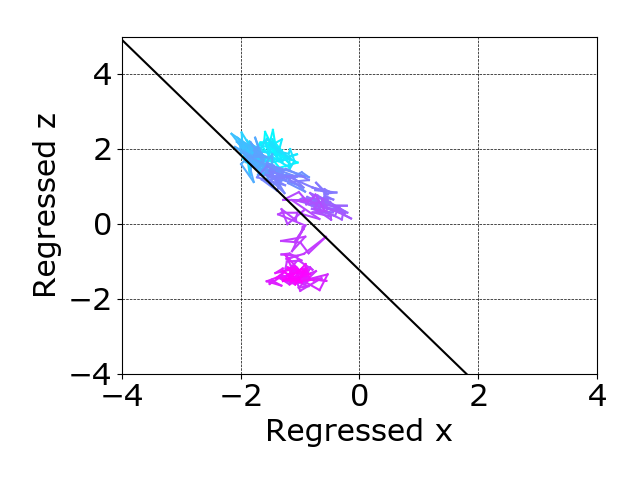} \\
        
        \multicolumn{2}{c}{Line (f)} \\
        
        \includegraphics[width=6cm]{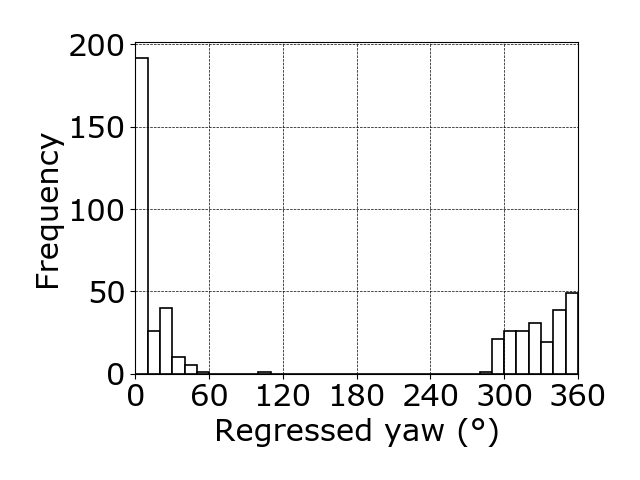} &
        \includegraphics[width=6cm]{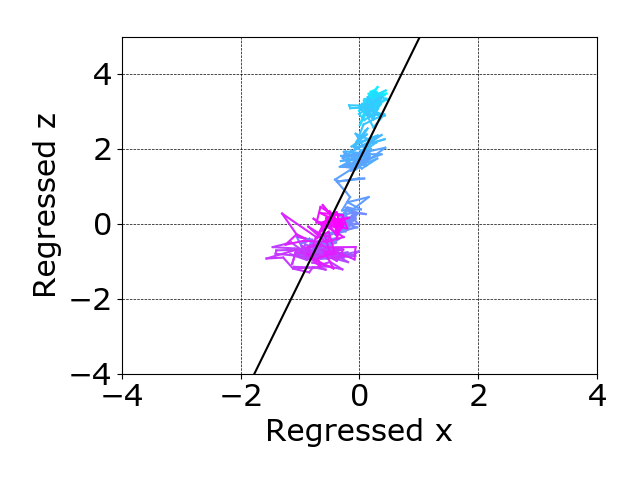} \\
        
        \multicolumn{2}{c}{Line (g)} \\
        
        \includegraphics[width=6cm]{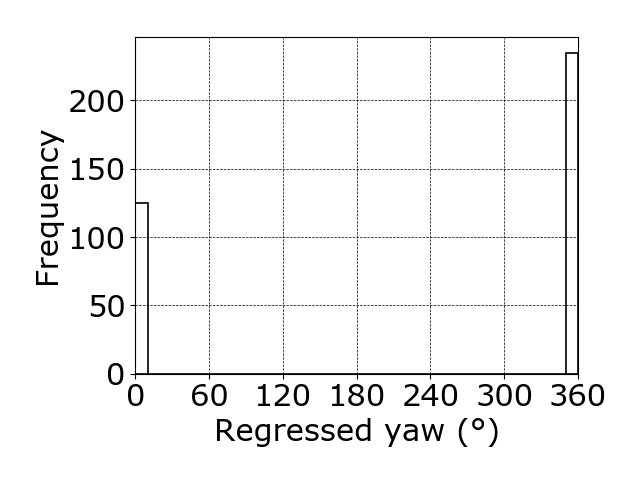} &
        \includegraphics[width=6cm]{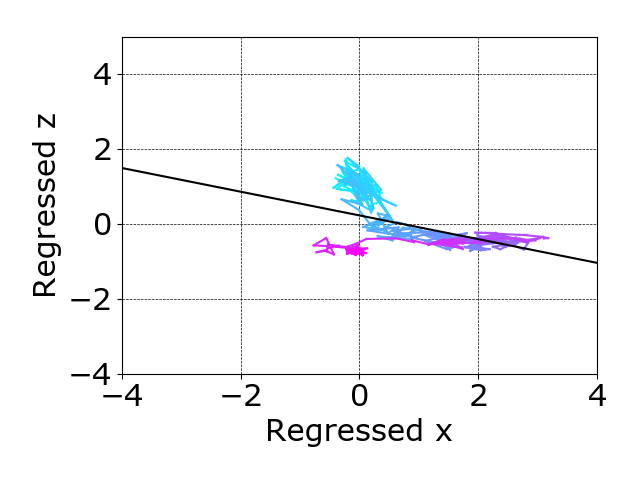} \\
        
        \multicolumn{2}{c}{Line (h)} \\
        
    \end{tabularx}
    \caption{ The regressed $(x,z)$ positions and regressed yaw yielded from the proposed network over the course of $4$ \textbf{straight line movements}. These paths are referred to as `lines' (e) through to (h) from top to bottom, where each path's regressed yaw and position graphs are displayed on the left and right respectively. The line of least residual squared error is plotted to ensure that the regressed positions follow a straight line as expected. }
    \label{tab:turtlebotline}
\end{table*}

Equation~\eqref{eq:aggstd} is used in order to produce values for standard deviation in the $(x,z)$ plane, which gives an idea of the spread in positions during in place rotations. Ideally this standard deviation here would be low --- suggesting that the positions predicted over the course of the in place rotation are relatively constant.

\begin{equation}
    \label{eq:aggstd}
    \sigma_{x, z} = \frac{1}{N - 1} \sum_{i=0}^{N} \sqrt{(x_i - \bar{x})^2 + (z_i - \bar{z})^2}
\end{equation}

\begin{table}[h]
    \begin{adjustwidth}{-0.8cm}{} 
        \centering
        \begin{tabular}{ c | c | c | c | c | c }
        
            \hline\hline
        
                 & Num. of & $(x, z)$ mean & $\sigma_{x, z}$              & Yaw s/b mean      & Yaw s/b mean \&         \\
            ID   & samples & position      & Equation~\eqref{eq:aggstd}   & \& std. ($^{o}$)  & std. (filtered) ($^{o}$)     \\
        
            \hline\hline
            
            (a) & $277$ & $(-0.76, 0.31)$ & $8.58\times10^{-2}$ & $7.69, 16.5$ & $2.74, 2.99$ \\ 
            (b) & $260$ & $(0.07, 1.31)$ & $4.88\times10^{-2}$ & $7.22, 10.9$ & $3.13, 2.85$ \\ 
            (c) & $267$ & $(-0.01, 0.11)$ & $1.53\times10^{-1}$ & $7.42, 14.4$ & $3.06, 3.78$ \\ 
            (d) & $284$ & $(0.25, 0.37)$ & $1.27\times10^{-1}$ & $7.89, 13.3$ & $3.35, 3.48$ \\ 
        
            \hline\hline
            
        \end{tabular}
        
        \caption{ An explicit readout of the calculated path-related metrics (described in full in Section~\ref{ch:methods:s:exp:ss:rnav:sss:simplepaths}, for the \textbf{in place rotation} graphs displayed in Table~\ref{tab:turtlebotspin}. Note that the yaw values are filtered in the last column via outlier removal with an inter-quartile range constant of $1.5$. Abbreviations: num., number; s/b, samples per bin; std., standard deviation. 
        \label{tab:simplepathstatsspin}}
    
        \vspace{1cm}
        
        \centering
        
        \begin{tabular}{ c | c | c | c | c }
        
            \hline\hline
        
                  & Num. of & Residuals  & Residuals squared & Yaw's mean  \\
            ID & samples & squared    & per sample        & \& std. ($^{o}$)  \\
        
            \hline\hline
            
            (e) & $523$ & $2395.52$ & $4.58$  & $356.77, 3.97$ \\ 
            (f) & $428$ & $2042.28$ & $4.77$  & $359.73, 1.90$ \\ 
            (g) & $487$ & $976.93$  & $2.01$  & $352.62, 25.8$ \\ 
            (h) & $360$ & $1040.76$ & $2.89$  & $359.39, 1.27$ \\ 
        
            \hline\hline
            
        \end{tabular}
        
        \caption{ An explicit readout of the calculated path-related metrics (described in full in Section~\ref{ch:methods:s:exp:ss:rnav:sss:simplepaths}, for the \textbf{straight line movement} graphs displayed in Table~\ref{tab:turtlebotline}. Abbreviations: num., number; std., standard deviation. 
        \label{tab:simplepathstatsline}}
    \end{adjustwidth}
    
\end{table}

\textbf{Firstly}, the results of the rotations are considered. Of note in Table~\ref{tab:simplepathstatsspin} is the standard deviation in positions $\sigma_{x, z}$. See that it is considerably low, especially considering that the spatial extents of the scene are of the order of $\sim 8$ (the units here \textit{are not metres}, but simply the length of the scene in the reconstructions). This confirms that during the TurtleBot's rotations, the regressed position remained approximately constant (as expected). Of note is that the mean position (plotted in black in Table~\ref{tab:turtlebotspin}) aligns with the relative position of the TurtleBot in the scene during the experiment. 

The regressed yaw should be evenly distributed in the range $0$ --- $360$\degree; hence when this range is divided into $36$ bins and plotted as a histogram, the number of samples in each bin should be roughly constant. However, the histograms in Table~\ref{tab:turtlebotspin} indicate a preference in regressing a yaw of $\sim 0$\degree and $\sim 180$\degree, these values align with the most highly sampled directions in the training dataset (due to the elongated geometry of the scene). Numerically, Table~\ref{tab:simplepathstatsspin} shows that standard deviation in the number of samples per bin is often greater than the mean, which in itself is skewed from the outliers. If this experiment were to be repeated, care should be taken to ensure that the training set is balanced not only in its positional distribution, but its rotational distribution, in order to prevent such regression disparities.

Nonetheless, the distributions are much more uniform than the straight line paths --- indicating that the pose regression network is still adequately sensitive to uniquely rotated images in the scene. When the outliers are removed from number of samples per bin, the standard deviation is considerably less, and the distribution is much more uniform (as expected).

\textbf{Secondly}, the results of the straight line movements are considered. In Table~\ref{tab:simplepathstatsline}, the value of the squared residuals per sample is low (considering the spatial scale of the scene displayed in Table~\ref{tab:turtlebotline}), meaning that the regressed positions fit to a straight line (as expected, since the data was gathered by moving the camera in a straight line). This fit can be confirmed visually in the plots provided in Table~\ref{tab:simplepathstatsline}.  

The rotations are expected to be constant, since the camera doesn't rotate during the straight line movement. This is reflected in the distribution of the regressed yaw values --- as observed in the standard deviations in Table~\ref{tab:simplepathstatsline}. Indeed, the regressed yaws are distributed tightly (low standard deviation) about the angle that the straight line path made with the scene during the collection of data.

In general the paths deviate slightly from the shapes/distributions expected but on average, tend towards the expected.

\section{Experiment 3.2: Compound paths}
\label{ch:resdis:s:advancedpaths}

As per Section~\ref{ch:methods:s:exp:ss:rnav:sss:advancedpaths}, the TurtleBot is tele-operated to gather $2$ compound paths which adequately survey the scene. Due to the more complex distribution of poses in the compound paths, it is not possible to generate metrics as in Section~\ref{ch:resdis:s:simplepaths} with which to compare the results to the expected distributions. As such, the visualizations provided in Table~\ref{tab:turtlebotadv} provide more general conclusions about the robot's navigation.

They demonstrate that the correct pose can be predicted \textbf{on average} from a number of image samples within a scene, and that these predictions do not group too tightly in visually similar areas, nor are they incorrectly outside the spatial extents of the scene --- meaning that extreme failure cases are limited. The `jumps' between positions are also sufficiently small: if they were too large it would suggest that the device has moved from one position to a significantly farther away position in the space of approximately $\frac{1}{30}^{th}$ of a second. In other words, the regressed poses are consistent with the TurtleBot's maximum operational velocity and angular velocity.

As an aside: the distinct paths in each of the figures in Table~\ref{tab:turtlebotadv} qualitatively reflect the paths taken by the TurtleBot when the image samples were collected. In short, this demonstrates that the TurtleBot is able to maintain an accurate estimation of where it is (relative to a scene). This allowed for the creation of a simple point-to-point navigation algorithm, based on shortest path algorithms and a floormap automatically generated from the photgrammetry results. This was successfully ran on the robotic platform in real time to demonstrate how the model proposed in this work can be used for navigation.

\begin{table*}[h]
    \vspace{-3cm}
    \begin{adjustwidth}{-1.3cm}{} 
        \centering
        
        \begin{tabularx}{\linewidth}    
            { *{1}{P{\linewidth}} }
            
            \includegraphics[width=16cm]{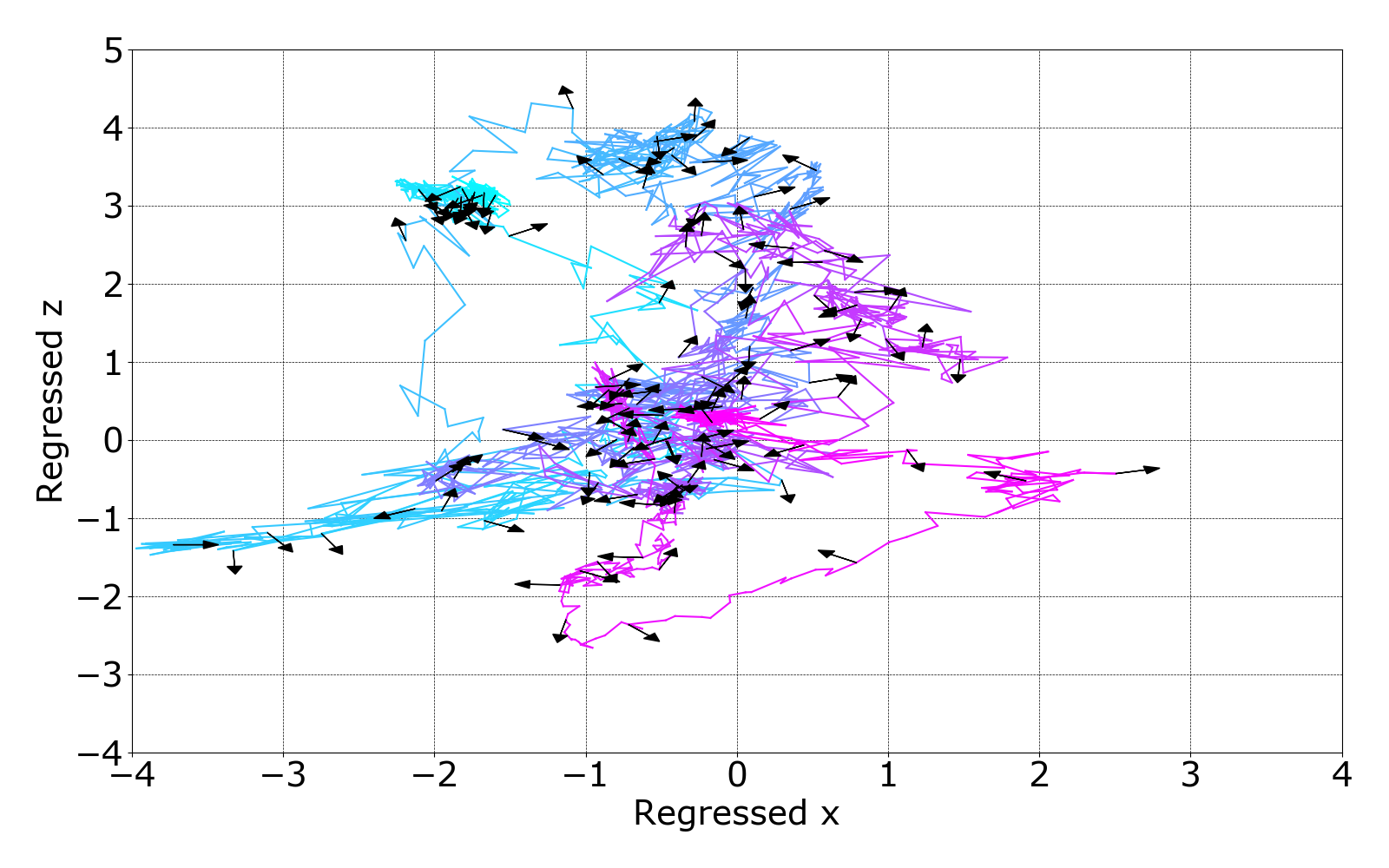} \\
            \includegraphics[width=16cm]{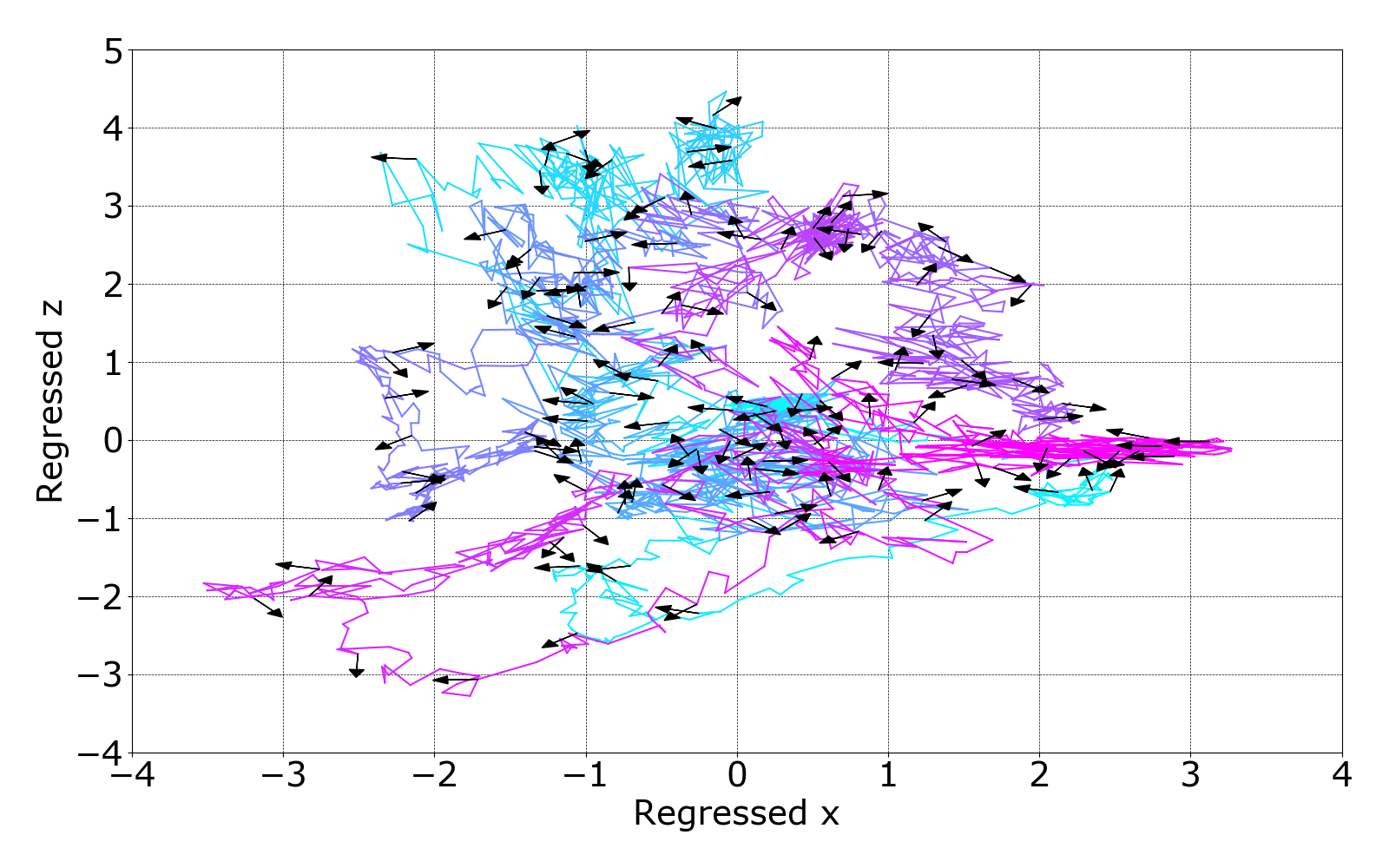} 
            
        \end{tabularx}
    \end{adjustwidth}
    \caption{ The regressed $(x,z)$ positions and regressed yaw yielded from the proposed network over the course of $2$ compound paths (\textbf{involving in place rotations and straight line movements}). For every $16$ regressed poses, a black arrow is rendered which represents the average yaw regressed over the said poses, for the purpose of visualization. }
    \label{tab:turtlebotadv}
\end{table*}

\chapter{Conclusion}
\label{ch:conc}

In conclusion, the effects of adding a geometric loss term to an existing pose regression network are investigated, and the performance of such a system is compared to other similar systems across common image localization benchmarks. Improvements to performance in the image localization task are observed, without any drastic increase in evaluation speed or training time. Particularly, the median positional accuracy is --- on average --- increased for indoor datasets when compared to a version of the model without the suggested loss term. 

This work suggests that the simple Euclidean distance between the ground truth and the regressed vectors, although often used as a measure of loss for many neural networks, can be improved upon. Specifically, loss functions designed with the primary goal of the network in mind may yield better performing models. For pose regression networks, the distinct nature of positional and rotational quantities indicates that non-naive methods must be used to combine them.

Furthermore, the robustness, accuracy and generalization of a pose regression network is inherited by a robotic navigation algorithm (which has the pose regression network at is core). Proof-of-concept investigations demonstrate that a navigation pipeline can be created end-to-end for an unknown scene with the only input being $\sim 320$ seconds of video which adequately surveys the scene. Such a pipeline has applications in domestic robotics, living assistance and unmanned ground vehicle technology, and is a stepping stone in the voyage towards more intelligent, robust and performant indoor robots.

\section{Future work}

The results obtained during this research project suggest a number of interesting avenues for future work, especially in relation to the pose regression neural networks explored throughout this work. 

It could be said that this work's primary concern was with the development of an augmented pose regression network, with particular attention being paid to the loss function. However, this is obviously not the only way to alter the architecture of a neural network. In fact, for pose regression networks, a post-processing module that operates completely independently of the network could be utilized to improve performance. The module would operate as such:

\begin{enumerate}
    \item Use a pose regression network to predict a pose estimate for a given image in a scene.
    \item Synthetically render a population of images around the estimated pose using a dense, photogrammetric reconstruction of the scene. These images will thus be labelled with a known pose (to explictly define the camera matrix, the pose must be defined in full).
    \item Compute the similarity of each synthetic image to the original input image, and return the pose of the most similar synthetic image, thus refining the estimated pose.
\end{enumerate}

The pipeline described throughout this work could also be used to inform flying drones, in order to fully take advantage of the 6 degrees of freedom regressed by the network. Data augmentation could be used in this case to artificially generate frames which demonstrate different rotation values in the roll axis, as such images would be expected during drone flight.

Finally, an understated benefit of the proposed pipeline which could be explored is how the pose regression network cooperates with other aspects of an `intelligent' robot's operation. Such a device is likely to leverage other advancements in computer vision in order to inform it's operation regarding object detection, grasp estimation, scene understanding and so forth. These visual tasks are often approached using deep neural networks or other deep learning architectures. Such architectures typically, as an initial step, use CNNs to extract feature sets of varying dimensionality from input images, which can then be used for regression or classification. 

Since the pose regression network defined in this work already extracts said features, the computational overhead incurred when adding object detection networks (or networks which solve other computer vision tasks) is considerably decreased. Investigating how the models proposed in this work could be used in an \textit{ensemble} of task specific networks could also form the basis for future work in the field of deep learning and robotics.

\appendix

\chapter{Project proposal}
The project proposal is rendered within this section to provide the reader with an idea of what the initial focus of the project was.



# Revised Research Proposal

**Title:**       Indoor Robotic Navigation

**Author:**      Isaac Ronald Ward

**Supervisor:**  W/Prof. Mohammed Bennamoun

**Degree:**      B.Sc.Hons. (12 point project)

# Background

With the advent of cheap RGB-D sensors, unmanned ground vehicles (UGVs) that can intelligently navigate environments have become increasingly common. Domestic robots such as the iRobot Roomba 980 have begun to move away from naive navigational techniques and now incorporate vision-based navigational techniques in their operation [1].

These sophisticated techniques include visual-based localization [2] and navigation [3], simultaneous localization and mapping (SLAM) [4], 3D mapping [5], visual-inertial odometry [6] and path planning [7]. Tractable, real-time and integrated implementations of these techniques are thus required for intelligent robotic navigation in UGVs and in other devices (drones, rovers, etc.).

State of the art implementations have been demonstrated to generalize to robotic devices [1], [6] & [7] and have shown improvements in transport costs [3], memory usage [8] and reinforcement learning convergence rates [2], [9].

# Aim

The aim of this project is to design a navigation pipeline that is to robust to variations in lighting and occlusion. A robotic implementation is to be created using a TurtleBot for the mobile base.

It is feasible to generate maps of the environment using only RGB inputs; the monocular SLAM technique implemented in [4] is able to run in real-time on a CPU *without* using scaled sensors (RGB-D) and is able to handle inputs at varying scales with a direct (featureless) approach. Furthermore, indoor SLAM has been achieved for small autonomous devices by using a combination of odometry, laser scans and real time RGB-D input [6], but this approach *does not* optimize navigation or implement vision-based global localization.

RGB-D sensors have been leveraged to create 3D voxel-based occupancy maps for the purpose of navigation [10], but this project aims to construct semi-dense point clouds (from laser scans and real-time RGB-D inputs) and proceed to use these as inputs for localization and optimized path planning algorithms rather than use large occupancy maps [8].

Machine learning methods have been used to create solutions to the most challenging problem in the navigation pipeline: localization. Methods that have already been investigated include the work of Brachmann et al. [12] which provides localization comparable to GPS localization using only RGB inputs. Such publications ensure that such demanding localization results are reasonable (see literature review for a more detailed discussion of existing results).

These achievements, alongside the autonomous devices constructed in [1], [2], [3], [6] & [9] ensure that the aforementioned aim is reasonable and that the project is feasible.

This project will demonstrate that state of the art techniques can be employed to

create devices that can autonomously navigate indoor environments in real-time, resource constrained contexts using consumer-level hardware. This research applies to domestic/service robots (vacuum cleaners) and other lightweight self-driving vehicles (UGVs, drones) and will provide a platform for the investigation of more complex technologies (such as environment interaction and recognition).

## Method

The project first requires developing/investigating a model that can construct 3D point cloud maps in real time from RGB-D inputs for the purpose of comparison to laser scans gathered prior. Point clouds — which can be created from laser scans and RGB-D inputs and photogrammetry methods — are to be used in place of voxel-based occupancy maps (which typically have much higher memory demands) [8]. The mapping techniques outlined in [5], [8] are to be investigated.

Using the a map of the computer science building's interior as a reference, the TurtleBot will locate its position within the building based on the current RGB-D sensor data. (Adaptive) Monte Carlo Localization (MCL) [6], [8] is to be investigated but machine learning techniques will be created to ensure peak performance.

Path planning and navigation algorithms (which will take into account object detection and avoidance) are to be created. Algorithms for path planning, path execution and object avoidance do exists but they will have to be tailored for the proposed device. Any developed methods should be compared to heuristically aided methods and other techniques as in [7]. Commonly considered evaluation metrics include trajectory length averages [2], time penalised success rates [2], average deviation from predicted position [3], dimensionless energetic cost of transport (CoT) [3] and comparisons to simple heuristics [7].

Real-world (rather than synthetic) datasets — such as Microsoft's '7 Scenes' — are to be used for training and evaluating any machine learning based subsystems.

## Software and Hardware Requirements

| Requirement | Number | Total cost |
|---|---:|---:|
| Computer science building map | x1 | already owned |
| Microsoft Kinect v1 RGB-D sensor | x1 | already owned |
| TurtleBot | x1 | already owned |
| 13.3", 8GB RAM, Ubuntu laptop | x1 | $1206.97 (ex GST) |

All other required software is open source (computer vision libraries etc.). If computational time is required cloud computing services and accelerated hardware solutions will be investigated. If image surveys of the computer science building are required (for the purpose of photogrammetry) then DSLR cameras will be investigated.






\chapter{Code repositories}
\label{ap:repos} 

Links to the codebases which were written, adapted and extended in this work are provided here. For completeness, repositories holding early versions of code are also provided. Since some of these repositories are under private access (due to concerns regarding sensitive data, intellectual property \etc), please request access to them as required. 

An great number of scripts were also written during the course of this project for the purpose of generating visualizations, robotic navigation and reading photogrammetry statistics. Copies of these can also be provided upon request.

Codebases:
\begin{itemize}
    \item Pose visualization system for point clouds and meshes. \\Available at \url{https://github.com/Isaac-Ronald-Ward/robotnav}.
    \item Pose visualization system and synthetic image rendering pipeline for photogrammetric outputs. \\Available at \url{https://github.com/Isaac-Ronald-Ward/poser}.
    \item Adapted TensorFlow implementation of PoseNet \cite{kendall2015posenet} for designing and testing loss functions. \\Available at \url{https://github.com/Isaac-Ronald-Ward/geoloss}.
\end{itemize}

\chapter{Requirements}
\label{ap:reqs}

An explicit list of the software and and hardware required for this project is provided in Table~\ref{tab:software} and Table~\ref{tab:hardware}.

\begin{table}[H]
    \centering
    \begin{tabular}{ | p{3cm} | p{11cm} | }
        
        \hline
        
        Item & Rationale \\
        
        \hline
    
        COLMAP  & 
        To complete the SfM and MVS processes, this open-source, GPU-accelerated software is used. \\ 
        
        \hline 
    
        NVIDIA CUDA Toolkit  & 
        To allow the machine learning code to interface to the GPU outlined in Table~\ref{tab:hardware}, thus heavily speeding up network training operations. \\ 
        
        \hline 
        
        TensorFlow & 
        To allow for high level development and modification of existing neural network architectures. This machine learning library obfuscates many of the minute details of neural networks and provides a set of building blocks for developing such architectures. The modified PoseNet is designed chiefly using this library. \\
        
        \hline 
        
        The Robotic Operating System (ROS) & 
        To facilitate the driving and operation of the TurtleBot. ROS is a \textit{set} of software libraries which provides drivers, common algorithms (pathfinding and navigation), visualization tools (Rviz) and so forth. \\ 
        
        \hline
        
        CloudCompare V2 &
        To visualize large point clouds and meshes ($> 45000000$ points). This open source software uses an OpenGL backend to efficiently visualize large 3D data on consumer grade computers and laptops.  \\ 
        
        \hline
        
    \end{tabular}
    
    \caption{ Descriptions of the software required by this project and the rationale behind their requirement. }
    \label{tab:software}
    
\end{table}

\begin{table}[H]
    \centering
    \begin{tabular}{ | p{3cm} | p{11cm} | }
        
        \hline
        
        Item & Rationale \\
        
        \hline
    
        NVIDIA Tesla K40c enthusiast GPU  & 
        To facilitate the photogrammetry processes, and to considerably accelerate the network training times. In general, powerful GPUs are useful in heavy image processing and vision tasks, and many such tasks need to be completed in this project. \\ 
        
        \hline
        
        FujiFilm X-T20 camera body                  & 
        To allow for high definition, $1980\times1080$px images to be taken in a video sequence, such that data collection times are not excessive. High resolution and frame rate are required to avoid blur and allow for the maximum number of SIFT features to be computed (to encourage successful reconstructions). \\ 
        
        \hline
        
        FujiFilm $23$mm f$1.4$---f$16$ prime lens   & 
        Used in coalition with the camera body, this lens provides autofocus, low distortion and a wide lens, hence the images it produces survey the scene well. \\
        
        \hline
        
        TurtleBot v2                                & 
        To locally test the navigation algorithms, and to ensure that the entire pipeline described in this work (see Figure~\ref{fig:fullpipeline}) operates as expected. The Kobuki base provides a number of utilities and can be controlled with ROS --- allowing for teleoperation, AMCL navigation, \etc \\
        
        \hline
        
        Microsoft Kinect RGB-D sensor               & 
        To capture an image feed from on-board the TurtleBot and feed said images to the trained model, thus determining the pose. Using a different camera to the FujiFilm camera ensures that the generalization qualities of the network are tested: does the network generalize to noisier, lower resolution, wider angle, less exposed (darker) images? \\ 
        
        \hline
        
        Wireless router                             & 
        To allow for teleoperation of the TurtleBot (useful for data collection), a wireless connection must be made to an on-board router. \\
        
        \hline
        
        13.3”, 8GB RAM, HP Spectre with Ubuntu      & 
        To SSH into the TurtleBot, run the ROS nodes and the pose regression network. \\
        
        \hline
        
        Cables various                              & 
        To connect the router, Kinect and Kobuki base to the HP Spectre, and to charge the above devices. \\
        
        \hline
        
    \end{tabular}
    
    \caption{ Descriptions of the hardware required by this project and the rationale behind their requirement. }
    \label{tab:hardware}
    
\end{table}

\bibliography{cshonours}

\end{document}